\documentclass[review, nopreprintline, authoryear]{elsarticle}

\usepackage[a4paper, left=2.5cm, right=2.5cm, top=2.5cm, bottom=2.5cm]{geometry}

\usepackage[english]{babel}
\usepackage[utf8]{inputenc}
\usepackage[english]{babel}

\usepackage{amsmath}
\usepackage{amstext}
\usepackage{amssymb}
\usepackage{latexsym}
\usepackage{dsfont}
\usepackage{mathrsfs}
\usepackage{bbold}
\usepackage[toc,page]{appendix}
\usepackage{tikz}
\usetikzlibrary{fit,arrows.meta}
\definecolor{dblue}{RGB}{31,77,122}
\usetikzlibrary{calc}
\usetikzlibrary{arrows.meta}
\usetikzlibrary{decorations.pathreplacing}
\usepackage{hyperref}
\hypersetup{
  colorlinks=true,
  unicode=true
}

  \usepackage{graphicx}
\usepackage{wrapfig}
\usepackage{float}
\usepackage{caption, booktabs}
\usepackage{graphics}
\usepackage{subcaption}
\usepackage{lipsum}
\usepackage{lscape}
\usepackage{amsmath,amssymb,amsfonts,amsthm}
\usepackage[table, dvipsnames]{xcolor}
\usepackage[referable, flushleft]{threeparttablex} 
\usepackage{tabularx}
\usepackage{tikz}
\usepackage{xcolor}
\definecolor{mypurple}{RGB}{128,0,128}  
\usepackage{float}
\usepackage{mathrsfs}
\usepackage{soul}

\usepackage{url} 

\usepackage{bm}
\usepackage{bbm}
\usepackage{array}
\usepackage{cals}

\usepackage[noend, linesnumbered,ruled,vlined]{algorithm2e}

\usepackage{tikz}

\usepackage{attachfile}
\usepackage[dvipsnames,x11names]{xcolor}
\definecolor{darkgreen}{rgb}{0.0,0.4,0.0} 
\definecolor{darkpurple}{rgb}{0.4,0.0,0.4} 
\definecolor{darkviolet}{rgb}{0.58,0.0,0.83} 

\usepackage{siunitx} 
\usepackage{colortbl, xcolor, booktabs}
\usepackage{siunitx}
\usepackage{pgfplotstable, colortbl, booktabs}
\pgfplotsset{compat=1.18}

\usepackage{etoolbox}
\newrobustcmd\Bold{\DeclareFontSeriesDefault[rm]{bf}{b}\bfseries}

\definecolor{b}{rgb}{0,0,.8}  
\definecolor{g}{rgb}{0,.6,0}  
\definecolor{n}{rgb}{0,0,0}  
\definecolor{h}{rgb}{0.4,0.2,0.2}  
\definecolor{v}{rgb}{0.2,0.6,0}

\definecolor{lightgray}{rgb}{0.9,0.9,0.9}
\definecolor{darkgreen}{rgb}{0,0.5,0}

\newcommand{\bsb}{\boldsymbol b}

\newcommand{\bsW}{\boldsymbol W}

\newcommand{\ov}\overline

\newcommand{\rig}\right
\newcommand{\lef}\left
\newcommand{\nf}\normalfont

\definecolor{darkgray}{HTML}{6e6e6e}

\definecolor{dcyan}{rgb}{0,0.5,.5}

\definecolor{dgreen}{rgb}{0,0.35,0}

\usepackage{tikz}
\usepackage{lineno}

\usepackage{svg}
\usepackage[final]{changes}
\usepackage{booktabs}
\usepackage{multirow}
\usepackage[table]{xcolor}
\usepackage{adjustbox}

\modulolinenumbers[0]

\makeatletter
\newread\pin@file
\newcounter{pinlineno}
\newcommand\pin@accu{}
\newcommand\pin@ext{pintmp}
\newcommand*\partialinput [3] {%
  \IfFileExists{#3}{%
    \openin\pin@file #3
    \setcounter{pinlineno}{1}
    \@whilenum\value{pinlineno}<#1 \do{%
      \read\pin@file to\pin@line
      \stepcounter{pinlineno}%
    }
    \addtocounter{pinlineno}{-1}
    \let\pin@accu\empty
    \begingroup
    \endlinechar\newlinechar
    \@whilenum\value{pinlineno}<#2 \do{%
      \readline\pin@file to\pin@line
      \edef\pin@accu{\pin@accu\pin@line}%
      \stepcounter{pinlineno}%
    }
    \closein\pin@file
    \expandafter\endgroup
    \scantokens\expandafter{\pin@accu}%
  }{%
    \errmessage{File `#3' doesn't exist!}%
  }%
}
\makeatother

\makeatletter
\def\namedlabel#1#2{\begingroup
  #2%
  \def\@currentlabel{#2}%
  \phantomsection\label{#1}\endgroup
}
\makeatother

\makeatletter
\newcommand{\newtag}[2]{%
  \phantomsection
  \def\@currentlabel{#2}%
  \def\@currentlabelname{#2}%
  \label{#1}%
}
\makeatother

\makeatletter
\newcommand{\newtagbf}[2]{%
  \phantomsection
  \def\@currentlabel{\textbf{#2}}%
  \def\@currentlabelname{\textbf{#2}}
  \label{#1}%
}
\makeatother

\graphicspath{{./anc/plots/}}
\svgpath{{./anc/plots/}}

\tikzset{state/.style={circle, draw, minimum size=0.8cm}}
\definecolor{dblue}{RGB}{31,119,180}
\definecolor{dcyan}{RGB}{0,188,212}
\definecolor{dred}{RGB}{214,39,40}

\begin{document}

\begin{frontmatter}

  \journal{JournalName}

  \title{Recurrent Neural Networks with Linear Structures for Electricity Price Forecasting.}

  \author[1]{Souhir Ben Amor\corref{cor1}}
  \ead{souhir.benamor@uni-due.de}
  \cortext[cor1]{Corresponding author}

  \author[1]{Florian Ziel}
  \ead{florian.ziel@uni-due.de}

  \address[1]{Chair of Data Science in Energy and Environment \\ University of Duisburg-Essen \\
  Germany}
  \newpage
  \begin{abstract}
    We present a novel recurrent neural network architecture specifically designed for day-ahead electricity price forecasting, aimed at improving short-term decision-making and operational management in energy systems. Our combined forecasting model embeds linear structures, such as expert models and Kalman filters, into recurrent networks, enabling efficient computation and enhanced interpretability. The design leverages the strengths of both linear and non-linear model structures, allowing it to capture all relevant stylized price characteristics in power markets, including calendar and autoregressive effects, as well as influences from load, renewable energy, and related fuel and carbon markets. For empirical testing, we use hourly data from the largest European electricity market spanning 2018 to 2025 in a comprehensive forecasting study, comparing our model against state-of-the-art approaches, particularly high-dimensional linear and neural network models. \added{In terms of RMSE,} the proposed model achieves approximately \deleted{12\%} \added{11\%} higher accuracy than \deleted{leading benchmarks} \added{the best-performing benchmark}. We evaluate the contributions of the interpretable model components and conclude on the impact of combining linear and non-linear structures.\added{ We further evaluate the temporal robustness of the model by examining the stability of hyperparameters and the economic significance of key features. Additionally, we introduce a probabilistic extension to quantify forecast uncertainty.}
  \end{abstract}

  \begin{keyword}
    Recurrent Neural Networks, Electricity Price Forecasting, linear expert model, Kalman Filter, Parallel Branches, Skip Connections.
  \end{keyword}
  \begin{highlights}
  \item A Parallel-Branch Recurrent Neural Network model with Skip Connections is introduced.
  \item We combine linear expert models with two branches of recurrent neural networks.
  \item Our methodology is tested on the largest European electricity market data (2018-2025).
  \item The hybrid network architecture provides superior electricity price forecasts.
  \item In terms of the RMSE, it outperforms \added{the best} state-of-the-art benchmark by \deleted{over 12\%} \added{$\approx 11\%$}.
  \end{highlights}
\end{frontmatter}

\section{Introduction, Motivation}
\deleted{Since the 1990s, the deregulation of the electricity market, as well as structural reforms, has resulted in radical changes in global markets. More precisely, a competitive market replaced the traditional vertically integrated electricity market. Therefore, like other commodities, electricity is today traded under competitive instructions using spot and derivative contracts . For that purpose, companies that trade in the electricity market need robust price prediction methods to remain competitive.}
\added{Following the deregulation of the electricity market, vertically integrated systems have been restructured into competitive markets in which electricity is traded through spot and derivative contracts (\cite{shahidehpour2003market}). Consequently, market participants require robust price forecasting methods to maintain competitiveness.}\\
Forecasting electricity prices is an essential \deleted{and} \added{yet} challenging \deleted{issue} \added{task} for all market players \deleted{due to its} \added{given their} inherent \deleted{features} \added{volatility} and the wide consumption of electricity in \deleted{our} modern society. Indeed, accurate electricity price forecasting \added{(EPF)} is crucial for \deleted{building} \added{effective} risk management and formulating reasonable competition strategies. Specifically, producers and traders can develop bidding strategies to minimize risks, maximize profits, and allocate purchases between spot prices and long-term bilateral contracts. In addition, electricity producers can rely on the forecast results to optimize unit production, while electricity consumers can adopt the forecast results to optimize their purchase portfolio (\cite{abedinia2015electricity}).\\
Therefore, price forecasting becomes an essential field of management, as future price changes are a challenging issue for all market players. In fact, an accurate price forecasting represents the basis for the decision makers to propose an optimal decision and formulate a reasonable plan to reduce market risk and consequently maximize the social and economic management benefits (\cite{hao2020modelling}; \cite{zhao2014combining}). Moreover, \deleted{as electric power is a clean and promising energy source, it plays an essential role in our daily lives, as it is considered environmentally friendly compared to other traditional energy sources and is therefore indispensable to the economy} \added{accurate forecasting is vital in strategically managing energy production, optimizing investment plans, and policy decisions. In this regard, precise forecasting supports sustainability overall financial stability of the organization. Therefore, besides achieving financial security, precise EPFs are instrumental in achieving climate goals by reducing emissions and aiding the incorporation of renewable energy into grids} ((\cite{o2025conformal}), \cite{o2025conformal}, \cite{jiang2019coal};\cite{wang2017multi}). \\
However, \deleted{regarding the specific characteristics of electricity and the uncertainty of market and bidding strategies, electricity price forecasts become more complex. Indeed, in contrast to other commodities, the impossibility of storing the electricity in order to use it in the future makes its prices more difficult to forecast. In other words, the electricity production and consumption should be taken simultaneously, which caused a high level of complexity and ambiguity in electricity markets.\\
This specific behaviour exhibits specific characteristics of the electricity price time series; such as high volatility, high frequency, non-constant mean and variance, a high percentage of unusual prices, unexpected price jumps or spikes quickly decay (associated with shock price elasticity demand and supply)} \added{electricity price forecasting presents significant challenges due to the inability to store electricity, which necessitates real-time balancing of supply and demand. This dynamic leads to high volatility, non-stationarity, and frequent price spikes.} (\cite{nogales2002forecasting}, \cite{kuo2018electricity}).\\
Moreover, \deleted{the seasonal behaviour of the prices is a direct result of demand fluctuations, which commonly arise from deterministic conditions (such as the weekly business hours and the number of daylight hours) or climate conditions (like precipitation levels and temperature).\\
Furthermore, there is an increasing incorporation of renewable energy sources and the development of smart grids, which is making the Electricity Price Forecasting (EPF) more challenging than it already is. Electricity grids are kept stable thanks to spot markets, which include the Day-Ahead Market, Intra-Day Market, and Balancing Market. It is noteworthy to mention that often, the high price volatility created by renewable generation forecast errors presents significant challenges as these sources are often intermittent and subject to variability in production} \added{electricity prices exhibit pronounced seasonality, primarily influenced by demand patterns and weather conditions. The growing integration of renewable energy sources and smart grids further increases price volatility because of the intermittent nature and forecast uncertainty associated with renewable generation.} (\cite{ortner2019future}).\\
\deleted{Accurate forecasting is vital in strategically managing energy production, optimizing investment plans, and policy decisions aimed at risk reduction. In this regard, precise forecasting supports sustainable overall financial stability of the organization. In addition to achieving financial security, precise EPFs are instrumental in achieving climate goals by reducing emissions and aiding the incorporation of renewable energy into grids.}
All these \deleted{various reasons} \added{factors} make \deleted{the prediction} \added{predicting} of electricity prices very difficult. This, in turn, has significantly enhanced research efforts toward modeling and forecasting spot electricity prices.

Over the past few decades, various forecasting techniques have been developed. Statistical models (\cite{muniain2020probabilistic}), and Artificial Intelligence methods (\cite{lago2021forecasting}) representing two major paradigms.
Traditional statistical models, such as time series models, multivariate regression, and econometric approaches (\cite{misiorek2006point}, \cite{ziel2018day}), have been extensively applied to EPF due to their ability to capture linear relationships and underlying seasonality in electricity markets (\cite{ben2018forecasting}). \cite{misiorek2006point} proposed the linear expert model \added{(LEM)} for \deleted{spot price electricity forecasting} \added{EPF}, which is considered a linear state-of-the-art model for its interpretability and robustness, particularly because they enhance forecasting efficiency by explicitly incorporating fundamental factors like load demand, fuel prices, and renewable energy generation. These models are widely adopted later in various EPF research (\cite{uniejewski2016automated}, and \cite{ziel2016forecasting}).\\
Although the electricity market is characterized by complex price formation processes and nonlinear interactions \deleted{between} \added{among} fundamental factors, purely linear or parametric statistical models are often \deleted{limited in effectiveness}\added{ineffective}, particularly when market conditions are volatile or extreme.
To address these limitations, Artificial Intelligence (AI) methods, particularly Machine Learning (ML) and Deep Learning (DL) techniques, have gained prominence in EPF. These methods excel at capturing nonlinear relationships and complex patterns in large datasets, making them suitable for modeling the intricate dynamics of electricity prices. Several deep learning models are considered for the Day Ahead EPF(\cite{pesenti2025explaining}, \cite{lopez2025forecasting}, and \cite{lago2018forecasting}), and they are considered as state-of-the-art models used as benchmarks in the field (\cite{lago2021forecasting}).
In particular, \added{recurrent neural networks} (RNNs) have shown promise for modeling temporal dependencies and nonlinear dynamics present in electricity markets(\cite{castello2025univariate}, \cite{yan2025novel}, and \cite{yang2024attnet}).

EPF has evolved in recent years, but there is still debate over which approach is best, as linear and nonlinear models each have distinct advantages. There have been studies that combine both approaches into a pure econometric hybrid model (\cite{ben2018forecasting}, and \cite{zhang2020adaptive}).
In this context, our paper integrates an expert model (\cite{misiorek2006point}, \cite{uniejewski2016automated}, and \cite{ziel2016forecasting}) into the architecture of the \deleted{recurrent neural network} RNN to leverage both models' strengths in forecasting day-ahead electricity prices, we also implement a Kalman filter branch.
\deleted{There are many directions in which this work contributes to the existing literature, which can be summarized in the following way:} \added{This research contributes to the existing literature in several key areas, which are summarized as follows:}

\begin{itemize}
  \item We propose a novel parallel-branch RNN architecture with skip connections (linear expert model), which allows the model to capture both linear and nonlinear relationships in the data. To the best of our knowledge, this is the first time such an architecture with such model components has been applied to day-ahead EPF.
    \deleted{On one hand we use the Kalman filter \added{(KF)} to replace the RNN in the architecture to highlight the importance of non-linearity in the RNN component by comparing it with a Kalman filter configuration results. On the other hand comparing the Kalman filter configuration with the linear expert model results to highlight the importance of temporal dynamics in the linear expert model.}
  \item \added{On one hand, when the Kalman filter is employed instead of the RNN in the architecture, the significance of non-linearity in the RNN component is highlighted when its results are compared with those of a Kalman filter configuration. On the other hand, a comparison between the Kalman filter configuration and the
    linear expert model emphasizes the importance of temporal dynamics in the latter.}
    \deleted{We demonstrate the effectiveness of our combined forecasting model on real-world data from the largest Europen electricity market between 2018 and 2025, which is chracterised by the presence of diffrent data regimes mainly due to the energy crisis in 2022.}
  \item \added{We demonstrate the effectiveness of our combined forecasting model on real-world data from the largest European electricity market between 2018 and 2025, which is characterised by multiple data regimes, mainly due to the 2022 energy crisis.}
  \item \added{We conduct a segment analysis of the test period to evaluate the model's performance across different market conditions, including periods of high volatility and stability, providing insights into its robustness and adaptability.}
  \item We implement our model using PyTorch. The code is available on GitHub \footnote{https://github.com/souhirbenamor/RNN-for-EPF}, making it accessible for further research and development in EPF.
  \item \added{We conduct an in-depth interpretability analysis of the hybrid architecture by examining the stability of economically meaningful linear feature coefficients before and after joint optimization.}
  \item We provide a deep analysis of the model hyperparameters selected during the optimization process, and explain how they impact forecasting accuracy. \added{We further evaluate the temporal robustness of the model by examining the stability of hyperparameters across different market regimes.}
  \item We also provide a forecast decomposition analysis to understand the contributions of different components of our model to the overall forecasting performance.
  \item  We provide a comprehensive evaluation of our model's performance, including comparisons with existing state-of-the-art models, and discuss the implications of our findings for future research in electricity price forecasting.
  \item \added{ We extend our model to a probabilistic forecasting framework that enables the quantification of forecasts uncertainty, a critical component for effective risk management in electricity markets.}
\end{itemize}

The rest of the paper is organized as follows: Section \ref{sec:lit_review} provides a literature review of existing methods for electricity price forecasting. Section \ref{sec:data} describes the data and study design. Section \ref{sec:methodology} presents the methodology, including the architecture of our proposed models. Section \ref{training_section} discusses the training procedure and hyperparameter optimization. Section \ref{results_section} discusses the results and performance evaluation of our model. Finally, Section \ref{conclusion_section} concludes the paper and outlines future research directions.

\section{Literature Review}
\label{sec:lit_review}
Different forecasting techniques have been proposed over the past few decades to solve the aforementioned management challenge.
EPF literature can be categorized into five main categories: (1) Fundamental methods (\cite{kanamura2022market}, \cite{paschalidou2025risk}, and \cite{tselika2024quantifying}) that model electricity price dynamics based on fundamental factors (e.g., demand, loads, and weather conditions). (2) Production cost models for long-term forecasting. (3) Reduced-form models, such as regime-switching models (\cite{xiong2019higher}, \cite{de2025comparison}) which explain price dynamics through seasonal volatility, spikes, mean reversion, and correlations between commodity prices. In addition, (4) A substantial amount of literature focuses on quantitative or statistical analysis (\cite{chen2025outlier}) to develop modeling frameworks based on the values of historical prices and/or exogenous variables or regressors, besides probabilistic models (\cite{marcjasz2023distributional}). (5) Artificial intelligence-based (or non-parametric) methods (\cite{ko2020deep}). (6) Hybrid approaches, which combine techniques from two or more groups listed above (\cite{zhang2020adaptive}, \cite{zhang2019hybrid}, and \cite{amor2024bridging}).
Due to the fact that statistical and machine learning methods and their combined models have shown to yield the best results (\cite{lago2021forecasting} and \cite{weron2014electricity}), this review pays particular attention to them and the related benchmarking model will also be centered around them.

To describe electricity prices' linear characteristics, statistical models use mathematical techniques combined with historical prices or price-related information to predict current prices. Among these models are the autoregressive moving average (ARMA) (\cite{chu2009forecasting}), autoregressive integrated moving average (ARIMA) (\cite{ramos2015performance} and \cite{zhao2017improving}), seasonal autoregressive integrated moving average (SARIMA) (\cite{camara2016energy}; \cite{mohamed2010short} and \cite{soares2008modeling}), and the generalized autoregressive conditional heteroscedasticity (GARCH) (\cite{garcia2005garch}). The ARIMA (or SARIMA) models have been coupled with GARCH models in some forecasting studies to simultaneously model the conditional mean and the conditional variance (volatility) (\cite{tan2010day}, \cite{kumar2017short}).\\
In addition to univariate baseline models, multivariate approaches incorporating fundamental market data are widely used in EPF, such as ARX and SARIMAX ( X refers to regressors) (\cite{mulla2024times} and \cite{misiorek2006point}).
In numerous studies, researchers have compared univariate and multivariate models. They show that the inclusion of exogenous variables enhances forecasting accuracy (\cite{gianfreda2020comparing}, \cite{ziel2018day}).
Building on these findings, so-called "expert" models have emerged. In this model, domain knowledge and market fundamentals guide the selection of regressors. This model class is based on a parsimonious autoregressive structure, originally proposed by \cite{misiorek2006point}.
Since their introduction, expert-type models have become a central approach to electricity price forecasting (EPF). The literature widely documents their adoption (\cite{uniejewski2016automated}, \cite{maciejowska2015short}, and \cite{weron2008forecasting}).
For example, \cite{nowotarski2018recent} benchmarks a classical ARX expert model and its multi-day extension (mARX) as primary forecasting tools for probabilistic day-ahead price prediction. They demonstrated competitive results for both point and interval forecasts using the GEFCom2014 dataset. The models incorporate expert-chosen lag prices, minimum price statistics, and load forecasts, sometimes outperforming neural networks and data-driven approaches.
\cite{ziel2016forecasting} develops a LASSO regularisation expert-type autoregressive model that captures cross-hour and intraday variations in electricity prices. Applied to multiple European markets, the model achieves robust forecasting accuracy, outperforming traditional expert and benchmark models.
An examination of 12 power markets was conducted in \cite{ziel2018day} by comparing both univariate and multivariate model frameworks for estimating day-ahead electricity prices. Various expert-type and high-dimensional autoregressive models were assessed, and they found that performance varied across markets and contexts rather than being uniformly superior. In addition, the authors demonstrate that combining forecasts from both approaches can improve predictive accuracy and provide useful advice on selecting variables for lasso-type models.
A recent study in \cite{ghelasi2025day} developed expert-type models for forecasting electricity prices in the German market over short, mid-, and long-term horizons. A comprehensive overview of the literature concludes that despite their state-of-the-art status for day-ahead forecasting, expert models based on autoregressive terms, calendar variables, and market fundamentals deteriorate over longer forecasting horizons. According to the authors, these challenges can be addressed by constraining model coefficients based on energy-economic theory and incorporating seasonal forecasts of key regressors. Their findings show that regularised linear models with carefully engineered constraints and inputs perform transparently and robustly for both operational and scenario-based long-term forecasting compared to machine learning approaches.\\
Given that these approaches are built on prior knowledge determined by experts, following \cite{uniejewski2016automated} and \cite{ziel2018day}, we refer to them as expert models.
Although these models can be interpreted, they often fail to capture the nonlinear dynamics in electricity markets (\cite {chan2012load}) as well as the high-dimensional interactions (\cite{aggarwal2009electricity} and \cite{weron2014electricity}), which can lead to inaccurate electricity price forecasting.

Due to the limitations outlined above and to consider the features characterising the electricity price dynamics, ML models have gained increasing attention in recent decades. Most publications in the EPF field adopt these methods because they have shown excellent modelling performances due to their ability to handle unstable nonlinear dynamics in high-dimensional data, making them suitable for EPF (\cite{hornik1989multilayer}).\\
As a result, a wide range of approaches has been developed that can be applied to complex dynamic systems and achieve accurate power price predictions (\cite{castello2025univariate}). This includes Artificial neural network (ANN) (\cite{lin2010enhanced}, \cite{panapakidis2016day}; \cite{sandhu2016forecasting}, \cite{ortiz2016price}, and \cite{keles2016extended}), fuzzy neural network (FNN) (\cite{hung2014wavelet}), weighted nearest neighbors (WNN) (\cite{lora2007electricity}), and adaptive wavelet neural network (AWNN) (\cite{pindoriya2008adaptive}, \cite{abedinia2015electricity}), Feed-forward neural network (FFNN) (\cite{anbazhagan2014day}), support vector machine (SVM) (\cite{yan2014mid}), and Random forest regressor (\cite{tschora2022electricity}).\\
\added{In addition to their application in the electricity market, ML models have demonstrated strong performance in predicting highly nonlinear, volatile price series across economic and commodity markets. To exemplify, \cite{xu2022thermal} and \cite{xu2023regional} proposed a non-linear autoregressive neural network (NAR-NN) with strong empirical forecasting performance, for thermal coal prices and regional steel prices, respectively. They demonstrated that carefully tuned non-linear autoregressive neural networks can achieve high predictive accuracy and stability over long time horizons and under significant market volatility. The suggested model enhances the forecasting accuracy and also the interpretability of these economic time series by explicitly separating trend, seasonal, and residual components and adaptively selecting pertinent features.}\\
As opposed to statistical approaches, ML does not make any assumptions about the functional form or statistical properties of the data set under consideration. Rather, nonlinear optimisation techniques are employed. Due to these features, they are more effective, accurate, and therefore popular for forecasting.

It has been demonstrated in recent EPF works that DL methods perform better than ML or statistical models (\cite{lago2018forecasting}. This is attributed to the efficiency with which DL models map complex nonlinear relationships among price time series and external regressors (\cite{li2022dense}).\\
  To illustrate, \cite{lago2021forecasting} introduced state-of-the-art deep neural networks and statistical models evaluated on five European markets for day-ahead electricity price forecasting. They demonstrated that well-optimised DNNs often attain competitive or high accuracy. To promote comparable research on the subject, they provided both methodological guidance and a replicable benchmark framework.
  Recently, \cite{yan2025novel} introduced a time-series decomposition-forecasting framework based on recurrent neural networks (RNN) for electricity prices. To break down time series data into trend, seasonal, and residual components, a hierarchical RNN-based time series decomposition model is presented. The time-series forecast is created by combining the predictions made by the RNN-based forecasting models for each component series. The suggested model is a promising and successful forecasting model, as shown by experimental findings.\\
  The use of advanced RNN structures, such as LSTMs and gated recurrent units (GRUs) (\cite{meng2022electricity}) networks, has proven much more accurate when analysing complex nonlinear time sequences (\cite{graves2013generating} and \cite{chung2014empirical}), such as electricity prices.
  In \cite{peng2018effective}, an LSTM with a differential evolution algorithm has been used to predict electricity prices. To verify the performance of the proposed model, experiments are conducted under electricity price scenarios \& Austria, New South Wales, and France.
  One example is \cite{lago2018forecasting}, which combines GRU and a DNN network as well as LSTM and DNN for day-ahead electricity price forecasting in Belgium. The DL algorithms were shown to be more accurate.
  In \cite{lehna2022forecasting}, a Convolutional Neural Network (CNN) combined with an LSTM model (resulting in a CNN-LSTM model) is used as a hybrid model for electricity price forecasting. To performance, they incorporated common external factors, including wind speed, average solar radiation, fuel and emission prices, and load. They suggested that the combined model leads to even better predictions of electricity spot prices.
  It is important to emphasise that linear problems are difficult to handle (\cite{tang2014novel}). Therefore, we cannot effectively extract all the features present in pricing and achieve reliable forecasting results if we only model linearity.
  In this vein, theoretical and empirical research has shown that combining several models can be a useful strategy to increase forecast accuracy and mitigate the risk associated with the inadequacy of individual models. Integrating the advantages of each method to improve is the primary objective of combining different methods.

  The existing hybrid models can be classified into \deleted{three} \added{four} categories;
  For the first category, the original electricity price is decomposed into linear and non-linear components, and a single appropriate model predicts each component. Thereafter, the final predicted price is the sum of the two predicted components. In this category, the most adopted approach is to combine the ARIMA and the ANN models (\cite{khashei2010artificial}; \cite{tseng2002combining}; \cite{valenzuela2008hybridization}; \cite{valenzuela2008hybridization}).
  Despite the fact that these hybrid models have achieved good forecasting results, they suppose that the nonlinear relations exist only in the residuals, and the linear and non-linear components can be modelled separately (\cite{ben2018forecasting}). Hence, the electricity price features cannot be well estimated.\\
  To avoid this inefficiency, a second hybrid model is proposed. Firstly, different models are applied separately to forecast the electricity prices; accordingly, different existing features can be captured by the suitable model. Secondly, an optimisation algorithm is adopted to compute each model’s weight coefficient. The final forecasting results are the product of each model’s prediction results and its weight coefficient (\cite{zhang2022short}, \cite{hussain2024enhancing}, \cite{ben2024meta} and \cite{lago2021forecasting}). To exemplify, \cite{nowotarski2014empirical} used twelve different individual models to predict electricity price separately, then identified weight coefficients of each model. \cite{niu2019combined} and \cite{darudi2015electricity} applied auto-regressive moving average, artificial neural network to predict electricity price; the weight coefficients of each model are identified. Although the accuracy of the second category of the hybrid model is enhanced compared to the first one, computing each model’s weight coefficient is challenging, which can affect the forecasting accuracy. \added{Moreover, estimating the model's components separately increases computational time and can hinder day-ahead EPF applications.}\\
  To overcome the above-mentioned limitation, some researchers have proposed a third type of hybrid model.. For this category, the electricity price is first decomposed into several components using a signal decomposition algorithm. Thereafter, an appropriate model is attributed to each component. Then, the final forecasting results are the sum of each component’s forecasting results. Among the existing signal decomposition approaches, discrete wavelet transform, empirical mode decomposition, Variational mode decomposition are the popular methods for signal decomposition (\cite{krishna2023electricity}, \cite{zhang2020adaptive}, \cite{wang2021new}, and \cite{paraschiv2023hybridising}).
  To exemplify, \cite{xiong2023hybrid} proposed a new hybrid forecasting framework to improve the forecasting accuracy of day-ahead electricity prices. The proposed model consists of three strategies. First, an adaptive copula-based feature selection algorithm is proposed for selecting model input features. Second, a new method of signal decomposition technique for the EPF field is proposed based on a decomposition denoising strategy. Third, a Bayesian optimisation and hyperband optimised LSTM model is used to improve the effect of hyperparameter settings on the prediction results. The effectiveness of the different methods was validated using five datasets set up for the PJM electricity market, and the results indicated that the proposed hybrid algorithm is more effective and practical for day-ahead EPF.\\
  Despite the third kind of hybrid model having demonstrated good performance, the choice of the decomposition approach and the appropriate single model is not suitable for the electricity price, which makes it difficult to further improve the accuracy and stability of the hybrid model. Moreover, decomposition techniques are dependent on parameters, e.g., mode numbers, and are thereby vulnerable to configuration choices and less adaptable to changes in market conditions. This results in suboptimal decomposition, particularly when price spikes are expected (which is very common in electricity markets). By ignoring these factors, the decomposition fails to achieve its intended purpose,  resulting in inaccurate as well as unstable forecasts (\cite{luo2024advanced}, and \cite{chai2024forecasting}).\\
  \added{To address the mentioned challenges, a fourth hybrid architecture has been developed across multiple disciplines in machine learning research, including image denoising, finance, environmental modelling, and CO2 emission prediction. Specifically, skip-connection architectures create implicit linear-nonlinear models by embedding linear structures through identity mappings or residual connections. This skip path directly transmits linear transformations to deeper network layers. It enables nonlinear components to learn how to correct the residuals. Notable examples include ResNet (\cite{he2016deep}) in Image recognition that incorporates skip connections to facilitate the training of deep networks. In the field of energy–economy–environment modeling, \cite{han2023novel} propose a novel economy and CO$_2$ emissions prediction model based on a residual neural network (ResNet) to optimize and analyze energy structures across different countries and regions worldwide. Skip links are used in the inner residual block of the ResNet to alleviate vanishing gradients as deep neural networks increase in depth. Consequently, the proposed ResNet can optimize this problem and protect the integrity of information by bypassing the input directly to the output, thereby increasing the precision of the prediction model. The experimental results show that the ResNet has higher correctness and functionality than traditional neural networks. For example, in finance, \cite{liu2025supply} suggest using enhanced graph attention neural networks combined with contrastive learning to assess financial risk in complex, interconnected systems. More precisely, they developed an enhanced graph attention neural network and contrast learning model (EGATCL). The model utilizes multi-layer graph attention mechanisms to capture higher-order dependencies while adaptively weighting adjacent nodes, and employs residual attention mechanisms with skip connections to mitigate over-smoothing problems. In addition, contrastive learning is applied to enhance the discriminative capability and robustness of node representations. Experimental results on real-world supply chain finance datasets demonstrate that EGATCL outperforms baseline models.
    In addition, \cite{wu2025image} developped an image denoising estimation network, a convolutional neural network with skip connections and two branches. The model combines a nonlinear branch for learning residual patterns with a noise-estimation branch for capturing structured signal components. As a result of this approach, baseline structure and nonlinear corrections are explicitly separated within a jointly trained architecture, improving both performance and interpretability.
    In addition, \cite{kittelsen2024physics} proposes an improved physics-informed neural network with skip connections and parallel components to model and control highly nonlinear and non-smooth dynamical systems in oil-well operations. This architecture combined linear dynamics with flexible neural components, using skip connections to stabilize training and improve gradient flow while preserving interpretability and robustness. The results indicate improved predictive ability and better control performance in gas-lifted oil wells.
    Concerning the field of building energy modeling, \cite{abbass2023comprehensive} proposes a comprehensive machine-learning framework that combines skip-connection neural network architectures with Bayesian hyperparameter optimization to address nonlinear, high-dimensional prediction problems. The study demonstrates that incorporating skip connections substantially improves gradient propagation, mitigates vanishing gradient issues, and enhances predictive accuracy compared to conventional fully connected networks. Similarly, Gaussian process-based Bayesian optimization is applied in parallel to efficiently navigate large hyperparameter spaces while balancing prediction performance and computational cost. Moreover, for the long sequence time series forecasting literature, \cite{ silva2023descinet} proposed DESCINet, a deep convolutional architecture that enhances the SCINet (Sample Convolution and Interaction Network) framework by introducing dense skip connections across hierarchical levels.
    These connections allow linear and nonlinear temporal features, such as trends, seasonality, and long-range dependencies, to be shared and combined more effectively across the network depth. In various areas, including environmental data, electricity, and exchange rates, their results show that facilitating information flow among model components improves stability, convergence, and predictive accuracy.
    To assess the role of skip connections in deep learning architectures, \cite{ oyedotun2021training} developed a skip-connection deep neural network architecture, which has been analysed both theoretically and empirically as a principled mechanism to improve both optimisation and generalisation in highly nonlinear models.
    To assess the role of skip connections in deep learning architectures, \cite{oyedotun2021training} developed a skip-connection deep neural network architecture, which has been analysed both theoretically and empirically as a principled mechanism to improve both optimisation and generalisation in highly nonlinear models. Results show that using identity skip connections effectively demonstrates that deep networks can avoid problems in latent representations that disrupt training in standard feed-forward architectures. A skip connection stabilizes gradient propagation, improves parameter conditioning, and promotes robust convergence by constraining hidden representations to remain close to the input space. Standard image classification benchmark datasets were used for the experiments.\\
    Across all of these applications, skip connections serve as architectural mechanisms to improve training robustness and expressiveness, but linear paths themselves are not intended to represent explicit economic or physical relationships (\cite{liu2025supply}, and \cite{kittelsen2024physics}). Further, despite the growing body of literature showing that jointly trained hybrid models with explicit linear components and nonlinear refinement are well-suited for complex, non-stationary time series problems, this architecture, to our knowledge, it has not yet been applied to day-ahead electricity price forecasting.
  }

  To develop a robust model for EPF, it is essential to first review the existing literature:\\
  Previous studies revealed several gaps. First, expert-based statistical models are rarely compared directly with advanced ML approaches (\cite{nowotarski2018recent} and \cite{ziel2016forecasting}). In addition, to the best of our knowledge, no prior studies have explicitly integrated linear expert models and ML-based models together to exploit their complementary linear and nonlinear capabilities.

  Second, most ML and DL research in EPF has concentrated on two main directions: (1) benchmarking the performance of various DL architectures (\cite{mujeeb2019deep}), and (2) designing hybrid DL models that combine multiple architectures, such as convolutional neural networks (CNNs) and LSTM networks (\cite{ahmad2019electricity,kuo2018electricity,xie2018day}). However, these studies typically omit comparisons with established state-of-the-art statistical models. For instance, (\cite{ugurlu2018electricity}) investigated RNNs for electricity price prediction, but restricted the analysis to a single market and benchmarked only against relatively simple statistical models such as seasonal ARIMA and Markov regime-switching. Although other DL methods were considered, the study did not evaluate their performance against simpler ML models.\\
  While advanced RNN variants such as LSTMs are commonly used in EPF (\cite{meng2022electricity}), there is no conclusive evidence that these increasingly complex architectures consistently produce better forecasts. Particularly, comparisons with the basic Elman RNN are often absent, even though standard DNNs remain widely used and are considered state-of-the-art in EPF (\cite{chinnathambi2018deep} and \cite{lago2021forecasting}). Similar findings have emerged in financial forecasting, where \cite{dautel2020forex} compared LSTM, GRU, simple RNN, and FNN for exchange rate forecasting and found that while deep models can perform well, they are also hard to train and tune. In some cases, simpler architectures matched or even outperformed more complex ones, especially in terms of trading profitability. In the same vein, the Elman RNN is preferable to more complex LSTM-based models in short-term predictions of stochastic processes because the additional complexity of sophisticated architectures often yields marginal gains that cannot be statistically distinguishable \cite{hewamalage2023forecast}.\\
  Motivated by these observations, we propose an EPF framework that combines an expert model with a standard Elman RNN and Kalman filter model. Using this design, we can assess each component's contribution as a stand-alone and exploit the strengths of both approaches.

  Thirdly, NNs have one major disadvantage despite their advantages. It involves optimising a nonlinear, nonconvex loss function over a large parameter space, which can lead to local minima, slow convergence, or reduced robustness. With many weights and layers, overfitting becomes a concern since noise is incorporated into the learned patterns. Despite numerous mitigation methods, they remain challenging (\cite{bejani2021systematic} and \cite{baldassi2023typical}).\\
  This is addressed in our approach through adaptive hyperparameter optimisation via Optuna, which allows the model to explore and select optimal configurations, learning rates, architectural settings, and regularisation parameters without manual trial and error. This, along with rolling-window evaluation, ensures that the model remains adaptable, retraining on the most recent market conditions and maintaining strong performance over time.

  Finally, in the broader literature, many hybrid EPF models adopt highly complex designs, incorporating multiple algorithms for data decomposition, feature selection, clustering, and single forecast models whose forecasts are combined (\cite{singh2018pso}, \cite{bento2018bat}, and \cite{xiao2017research}). A further problem is that these studies rarely assess the impact of substituting different variants of hybrid components, leaving it unclear whether each module is relevant. \added{Therefore, building on the success of skip-connection-based hybrid models in machine learning across different disciplines, we adopt a hybrid architecture that uses a similar principle to stabilise training and improve learning under changing conditions (especially given that our data includes the 2022 energy crisis period). In the majority of the literature mentioned above, skip connections create an implicit linear pathway that helps optimise DNNs. Our approach builds on this idea but makes the linear component clear and meaningful. In our model, the linear branch will be configured as a linear model with an interpretable econometric interpretation, and it will be trained alongside nonlinear components that provide the architecture with the stability and training reliability of modern hybrid neural networks while maintaining the clarity of traditional economic models. }

  \deleted{This is addressed in our method by integrating the expert model with a skip connection within the RNN architecture, with the final output being the sum of both components' forecasts. In spite of the simplicity of this structure, it retains the ability to capture both linear and nonlinear dynamics, making it easier to train and interpret. The network is extended with an additional branch to leverage the time-dependence captured by the RNN under a linear activation function, resulting in a Kalman filter configuration. }
  \added{Our research addresses these limitations by developing a hybrid architecture tailored to non-stationary electricity price dynamics based on jointly trained, structurally interpretable models. Specifically, four key aspects distinguish the proposed LEM-KF-RNN framework from existing hybrid models.
    First, the linear branch (LEM) is explicitly specified and economically interpretable, thereby preserving transparent price formation mechanisms for fuel prices, demand, and renewable generation. It is important to emphasise that the proposed framework allows us to investigate the economic meaning of the used set of features both before and after joint optimisation. In fact, the LEM coefficients are first estimated via ordinary least squares, which provides a clear econometric benchmark, and are then fine-tuned through gradient-based training along with KF and RNN parameters. Consequently, we can determine whether joint training may distort linear effects by refining rather than obscuring nonlinear and dynamic components.\\
    Second, the inclusion of a Kalman filter, as a trainable linear dynamical system, introduces state-space structures and temporal smoothing, which are rarely incorporated directly into deep learning architectures for day-ahead forecasting. As a result, the model captures evolving latent states and regime shifts more effectively than static linear terms or purely data-driven nonlinear dynamics.\\
    Third, the RNN branch provides an adaptable nonlinear modeling component. This can capture nonlinear relationships that are difficult to model with linear or state-space linear components. Training the RNN branch alongside the LEM and KF branches can serve as an adaptive correction component, selectively increasing and decreasing the linear signals based on market conditions. This enables an optimal balance among interpretability, dynamic structure, and nonlinear representation.\\
    Fourth, by simultaneously training all model components rather than ensemble forecasting (ex post), we benefit from coordinated learning between model branches. Further stabilisation is achieved by initializing the linear component using ordinary least squares (OLS) to speed up convergence and enhance training stability. In addition, simpler ensemble approaches require independent estimation of each component and additional decisions for optimising ensemble weights. They also incur high computational costs, as multiple models must be trained and maintained separately. In contrast, the proposed framework integrates all components within a single end-to-end model.\\
  Although the specific LEM–KF–RNN combination is, to the best of our knowledge, adopted for the first time in this study, the proposed model does not claim novelty in the general idea of hybrid linear–nonlinear modeling, but rather advances the literature by demonstrating how interpretable linear economics, state-space dynamics, and nonlinear neural representations can be coherently integrated within a single trainable architecture. Our research addresses this gap by presenting a framework for electricity price forecasting that aligns with recent advancements in ML for handling non-stationarity and structural breaks. The framework also incorporates domain-specific requirements to ensure interpretability, robustness, and feasibility.}\\
  Our dataset spans 2018 to early 2025, covering both stable periods with low price volatility and highly volatile phases, including the COVID-19 pandemic, the global energy crisis of 2021–2023, and the post-crisis years. Through this dual-period coverage, our framework is tested under diverse and challenging market conditions.

  \section{Data and Study Design}
  \label{sec:data}

  The purpose of this paper is to forecast next-day electricity prices in the German electricity market using a multivariate deep learning model that predicts all 24-hourly prices simultaneously (as indicated in the network architecture in Figure \ref{fig:combined_architecture}). This setup mirrors the European day-ahead market, which determines hourly prices in a single auction each day. Specifically, in the day-ahead electricity market, market participants submit binding bids for each of the 24 hours of the following day ($T+1$) by noon on day $T$. Afterwards, the market-clearing price (MCP) is determined at the point of intersection of the aggregated supply and demand curves, forming a uniform-priced auction. As the same information set is available for all 24 hours, we use a uniform feature set, incorporating multi-output neural architectures into our joint forecasting strategy.\\
  The data is sourced from the ENTSO-E Transparency Platform (ENTSO-E) \footnote{https://www.entsoe.eu/} and contains hourly observations of electricity prices, day-ahead forecasts of load, wind (on and offshore), and solar generation. Daily fuel prices (coal, gas, oil) and carbon emission allowances (EUA) are also included and they are sourced from Datastream platform \footnote{https://www.lseg.com/en/data-analytics/products/datastream-macroeconomic-analysis}. This dataset covers the period between 10 October 2018 and 13 January 2025, and has been preprocessed to ensure time consistency and proper alignment.

\begin{figure}[H]
    \centering
    \includegraphics[width=0.99\linewidth]{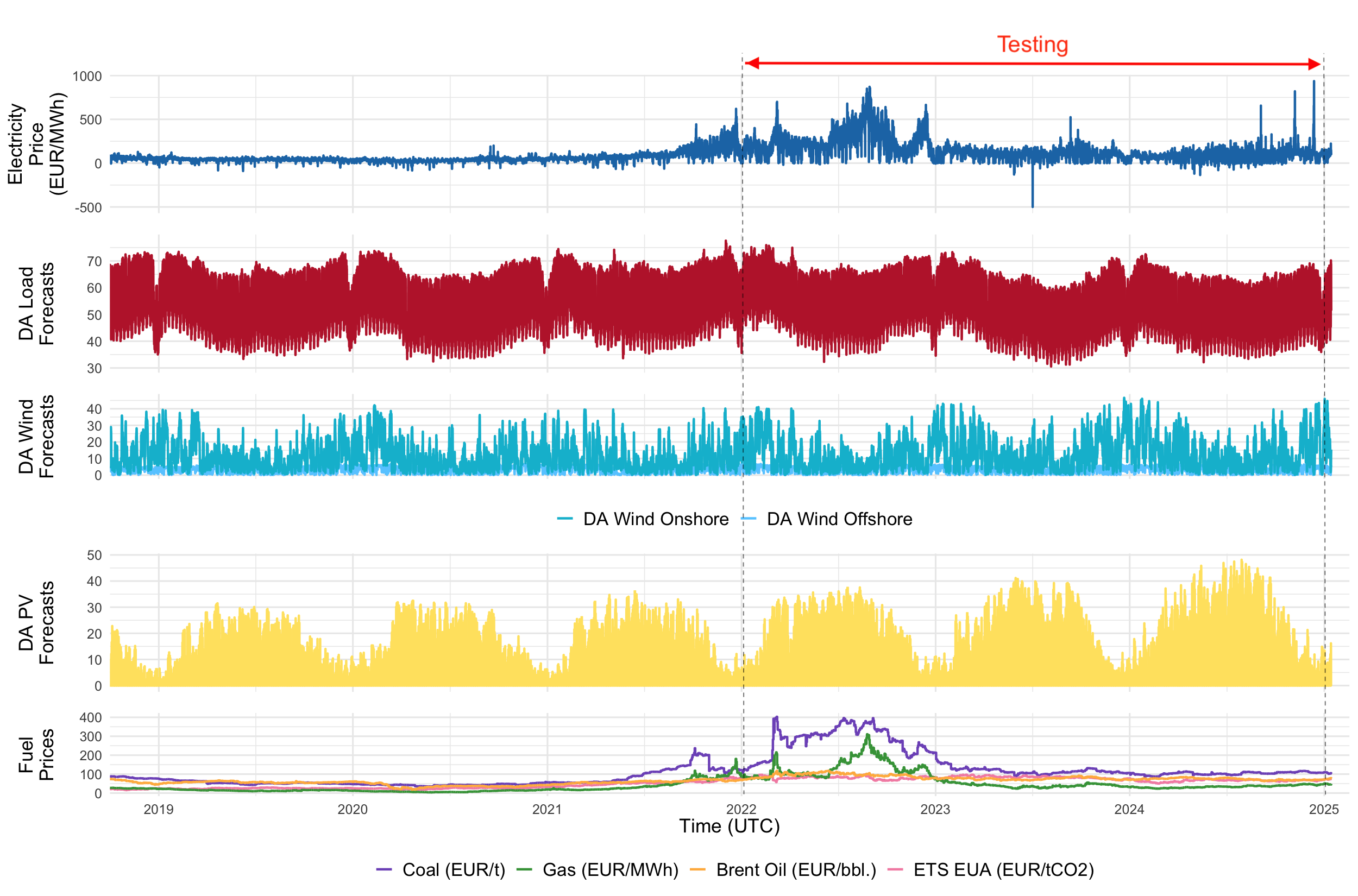}
    \caption{
      Multi-panel time series for Germany's day-ahead electricity price and related feautures (2018–2025).
      The panels show:
      \textcolor[HTML]{1f77b4}{Electricity Price (EUR/MWh)},
      \textcolor[HTML]{BE2637}{Day-Ahead Load Forecasts},
      \textcolor[HTML]{00BCD4}{Wind Onshore Forecasts},
      \textcolor[HTML]{66CCFF}{Wind Offshore Forecasts},
      \textcolor[HTML]{FFE36E}{PV (Solar) Forecasts},
      and fuel prices comprising
      \textcolor[HTML]{7E57C2}{Coal},
      \textcolor[HTML]{43A047}{Natural Gas},
      \textcolor[HTML]{FFB74D}{Brent Oil},
    and \textcolor[HTML]{F48FB1}{EU ETS EUA Allowances}, and \textcolor{red}{Test sample (2022–2025)} that includes the energy crisis period and the subsequent recovery period.}
    \label{fig:data}
  \end{figure}

  Figure \ref{fig:data} plots the electricity price and the other above mentioned features used to predict it. According to the plot, the energy crisis led to record-high German power prices in 2022. This could be assumed to be derived from prices for fuels (particularly gas and coal), which are used to generate power \cite{ghelasi2024day}. In this way, when gas and coal prices increase, electricity prices also rise.

  Summary statistics are presented in Table \ref{tab:descriptive_stats} (in the Appendix) for the variables during the initial calibration period and the subsequent out-of-sample test period. As we mentioned, electricity prices have become volatile in recent years. In our data, the price range is between a minimum of -500 EUR/MWh and a maximum exceeding 930 EUR/MWh in the test window reflects heightened volatility. In addition, the price standard deviation \deleted{decreased from 112.4 to 50.8} \added{increased from 53 to 115} EUR/MWh, indicating more outliers, particularly negative prices.

\begin{figure}[H]
    \centering
    \resizebox{\textwidth}{!}{
      \begin{tikzpicture}[>=Stealth, font=\sffamily, every node/.style={align=center}]
        \fill[blue!15]   (0,7.8) rectangle (3,8.2);
        \fill[orange!15] (3,7.8) rectangle (5.5,8.2);
        \fill[red!15]    (5.5,7.8) rectangle (8,8.2);

        \draw[thick, gray!70] (0,8) -- (8,8);
        \foreach \x in {0,3,5.5,8} {
          \draw[thick, gray!70] (\x,7.9) -- (\x,8.1);
        }

        \node[blue!80!black,   font=\small] at (1.5,8.4)  {Training sample};
        \node[orange!80!black, font=\small] at (4.25,8.4) {Validation sample};
        \node[red!80!black,    font=\small] at (6.75,8.4) {Test sample};

        \node[anchor=north, font=\scriptsize] at (0,7.8)   {Oct 2018};
        \node[anchor=north, font=\scriptsize] at (3,7.8)   {Jan 2020};
        \node[anchor=north, font=\scriptsize] at (5.5,7.8) {Jan 2022};
        \node[anchor=north, font=\scriptsize] at (8,7.8)   {Jan 2025};

        \draw[decorate,decoration={brace,amplitude=3pt,mirror},thick]
        (0,7.5) -- (3,7.5) node[midway,below=5pt,font=\scriptsize] {};
        \draw[decorate,decoration={brace,amplitude=3pt,mirror},thick]
        (3,7.5) -- (5.5,7.5) node[midway,below=5pt,font=\scriptsize] {2 years};
        \draw[decorate,decoration={brace,amplitude=3pt,mirror},thick]
        (5.5,7.5) -- (8,7.5) node[midway,below=5pt,font=\scriptsize] {3 years};

        \node[anchor=west, font=\bfseries] at (0,6.5) {Phase 1:};
        \node[blue!80!black, font=\scriptsize\itshape] at (1.0,6.8) {Initialize weights with OLS};
        \fill[blue!40] (0,6.33) rectangle (2.0,6.67);
        \draw[thick, blue!80!black] (0,6.33) rectangle (2.0,6.67);
        \node[white, font=\tiny\bfseries] at (1.0,6.50) {Train};

        \fill[blue!30]   (2.10,6.33) rectangle (2.58,6.67);
        \draw[thick, blue!80!black]   (2.10,6.33) rectangle (2.58,6.67);
        \fill[blue!30]   (2.70,6.33) rectangle (3.18,6.67);
        \draw[thick, blue!80!black]   (2.70,6.33) rectangle (3.18,6.67);
        \fill[orange!60] (3.30,6.33) rectangle (3.78,6.67);
        \draw[thick, orange!80!black] (3.30,6.33) rectangle (3.78,6.67);
        \node[white, font=\tiny\bfseries] at (3.54,6.50) {Val};

        \fill[blue!30]   (2.70,5.88) rectangle (3.18,6.22);
        \draw[thick, blue!80!black]   (2.70,5.88) rectangle (3.18,6.22);
        \fill[orange!60] (3.30,5.88) rectangle (3.78,6.22);
        \draw[thick, orange!80!black] (3.30,5.88) rectangle (3.78,6.22);
        \fill[orange!60] (3.90,5.88) rectangle (4.38,6.22);
        \draw[thick, orange!80!black] (3.90,5.88) rectangle (4.38,6.22);

        \fill[orange!60] (3.30,5.43) rectangle (3.78,5.77);
        \draw[thick, orange!80!black] (3.30,5.43) rectangle (3.78,5.77);
        \fill[orange!60] (3.90,5.43) rectangle (4.38,5.77);
        \draw[thick, orange!80!black] (3.90,5.43) rectangle (4.38,5.77);
        \fill[orange!60] (4.50,5.43) rectangle (4.98,5.77);
        \draw[thick, orange!80!black] (4.50,5.43) rectangle (4.98,5.77);

        \node[orange!80!black, font=\large]    at (4.74,5.12) {$\cdots$};
        \node[orange!80!black, font=\scriptsize] at (4.74,4.94) {Rolling validation};

        \draw[->, green!70!black, thick] (3.78,6.67) to[out=30, in=150, looseness=1.6] (2.70,6.22);
        \draw[->, green!70!black, thick] (4.38,6.22) to[out=30, in=150, looseness=1.6] (3.30,5.77);

        \fill[orange!60] (4.90,4.33) rectangle (5.38,4.67);
        \draw[thick, orange!80!black] (4.90,4.33) rectangle (5.38,4.67);
        \fill[orange!60] (5.50,4.33) rectangle (5.98,4.67);
        \draw[thick, orange!80!black] (5.50,4.33) rectangle (5.98,4.67);
        \fill[red!60]    (6.10,4.33) rectangle (6.58,4.67);
        \draw[thick, red!80!black]    (6.10,4.33) rectangle (6.58,4.67);
        \node[white, font=\tiny\bfseries] at (6.34,4.50) {Test};

        \fill[orange!60] (5.50,3.88) rectangle (5.98,4.22);
        \draw[thick, orange!80!black] (5.50,3.88) rectangle (5.98,4.22);
        \fill[red!60]    (6.10,3.88) rectangle (6.58,4.22);
        \draw[thick, red!80!black]    (6.10,3.88) rectangle (6.58,4.22);
        \fill[red!60]    (6.70,3.88) rectangle (7.18,4.22);
        \draw[thick, red!80!black]    (6.70,3.88) rectangle (7.18,4.22);

        \fill[red!60]    (6.10,3.43) rectangle (6.58,3.77);
        \draw[thick, red!80!black]    (6.10,3.43) rectangle (6.58,3.77);
        \fill[red!60]    (6.70,3.43) rectangle (7.18,3.77);
        \draw[thick, red!80!black]    (6.70,3.43) rectangle (7.18,3.77);
        \fill[red!60]    (7.30,3.43) rectangle (7.78,3.77);
        \draw[thick, red!80!black]    (7.30,3.43) rectangle (7.78,3.77);

        \node[red!80!black, font=\large]     at (7.54,3.05) {$\cdots$};
        \node[red!80!black, font=\scriptsize] at (7.54,2.82) {Rolling testing};

        \draw[->, green!70!black, thick] (6.58,4.67) to[out=30, in=150, looseness=1.6] (5.50,4.22);
        \draw[->, green!70!black, thick] (7.18,4.22) to[out=30, in=150, looseness=1.6] (6.10,3.77);

        \draw[thin, black] (-0.7,2.6) rectangle (8.9,8.9);

      \end{tikzpicture}
    }
    \caption{Rolling window and training strategy for electricity price forecasting. The approach consists of three phases: (1) \textcolor{blue!80!black}{Initial training on 3 years of historical data (from 10 October 2018 to 13 January 2020)}, (2) \textcolor{orange!80!black}{including Rolling validation with weight transfer between windows }, and (3) \textcolor{red!80!black}{Rolling testing (from 15 January 2022 to 13 January 2025)} with continuous model updates.}
    \label{fig:rolling_window}
  \end{figure}
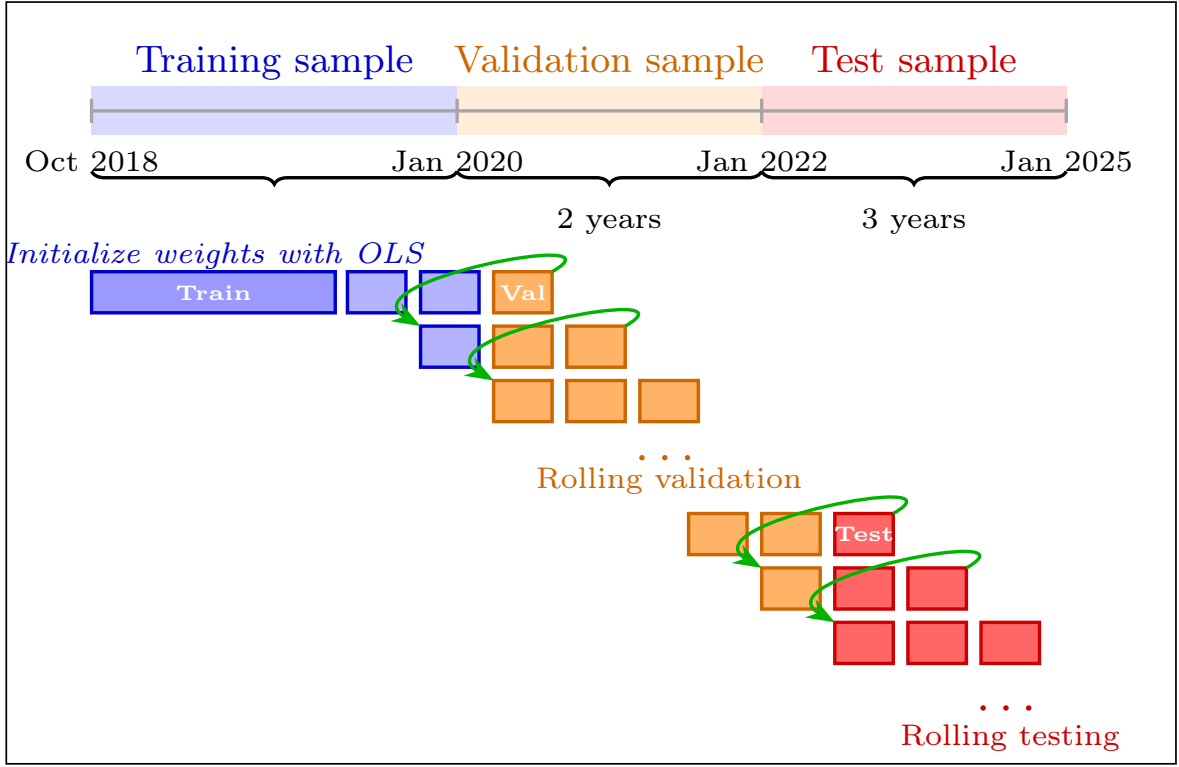

  Figure \ref{fig:rolling_window} shows the data split. The initial training window is further subdivided into two subsets: a training set, which consists of the first $D_{train}$ days, and a validation set, which comprises the $D_{val}$ days at the end of the window, two years before the test sample. The split is used exclusively during Optuna's hyperparameter tuning. As soon as the optimal configuration is identified, it is fixed and reused across all rolling forecasts.\\
  We also adopt a rolling forecasting scheme that mimics daily forecasting operations to ensure a realistic evaluation. As illustrated in Figure \ref{fig:rolling_window}, for each day of the out-of-sample period, the model is retrained using the most recent T=1091 days. Each iteration advances the rolling window by one day, generating a 24-hour forecast for the following day. The window then shifts forward by 1 day, and the process repeats. This continues sequentially until all test days (\deleted{730 days}\added{1095 days}) are forecasted, which runs from 15 January \deleted{2023} \added{2022} to 13 January 2025.\\
  Moreover, parameter warm-starting is one of the key features of the forecasting pipeline: the model parameters obtained at day $T$ are used to initialise the model at day $T+1$. By recursively reusing weights, training can be significantly stabilised, and convergence times are reduced. During the first step of the rolling loop, this warm-starting is combined with OLS-based initialisation of the linear component for models that contain skip-connections, i.e., linear expert model as one of the components or when it used as stand-alone model.
  As a result of this rolling and retraining strategy, along with the separation of training, validation, and test data, a robust assessment of generalisation performance can be achieved, mimicking the operational conditions of market forecasters.

  \section{Methodology}
  \label{sec:methodology}

  The objective of this study is to develop and evaluate a novel neural network architecture for day-ahead electricity price forecasting. The proposed architecture integrates RNN, Kalman filter and skip connections, aiming to improve predictive accuracy by leveraging both linear and nonlinear modeling capabilities.
  The proposed approach is used to model the vector of daily electricity prices $\boldsymbol{Y}_{t} = (Y_{t,0}, \ldots, Y_{t,S-1} )$ for $S=24$ as a dimension. Table \ref{tab:model_architectures} summarises the model architectures used in this study, including the linear expert model, the single RNN branch types models, and the different combined architectures.

  \subsection{Models' Input Features}
  Our input features are selected to reflect both key economic drivers and electricity market dynamics in order to accurately forecast day-ahead electricity prices.\\
  Most predictors are shared between linear and RNN-based forecasting models. These predictors include historical price data, market expectations, fuel fundamentals, and calendar effects.

  \begin{enumerate}
    \item Lagged electricity prices \\
      Autoregressive behaviour is accounted for by including historical day-ahead prices. \\
      For the linear expert model, we use lags of one, two, and seven days: $Y_{t-1}$, $Y_{t-2}$, and $Y_{t-7}$. \\
      The RNN models, however, includes only $Y_{t-1}$, since the sequential structure of the network takes into account additional temporal dependencies.

    \item Day-ahead fundamental indicators, including:
      \begin{itemize}
        \item Load forecast: Expected demand for the target day, denoted $\text{Load}_{t+1}$.

        \item Renewable energy forecasts:
          Onshore and offshore wind generation forecast, denoted $\text{Wind}_{t+1}$.
          Solar PV forecast, denoted $\text{Solar}_{t+1}$.
          These reflect the anticipated supply from variable renewable sources.

      \end{itemize}
    \item Fuel prices: These serve as proxies for marginal generation costs.
      Brent oil, denoted $\text{Oil}_{t-2}$.
      Natural gas, denoted $\text{NGas}_{t-2}$.
      Hard coal, denoted $\text{Coal}_{t-2}$.

    \item Carbon pricing: EU Emission Allowances (EUA) are included to account for CO$_2$ costs borne by fossil fuel generators, denoted $\text{EUA}_{t-1}$.

    \item Calendar dummies: Indicators for specific days of the week (especially Monday, Saturday, and Sunday) are included to capture weekly seasonality, usually weekday dummies, resp. one-hot encoded information and denoted collectively as $\text{cal}_t$.

  \end{enumerate}

  All features are available before the market closes at noon and reflect the information set accessible to market participants when submitting bids.
  For linear models, 24 separate outputs are generated for each hour of the next day, based on a fixed input vector. As a result, each output is modelled independently, and the structure makes no allowance for sequential time dynamics beyond the explicit inclusion of lag variables.\\
  RNN inputs consist of multivariate sequences over the past $T$ days, where the sequence length is considered a hyperparameter tunable within the interval $[1, 7]$. This temporal structure enables the network to learn dynamic patterns such as seasonality or delayed effects in the data. Consequently, lag features beyond $Y_{t}$ are omitted, as the architecture captures these dependencies implicitly.\\
  As a result, we have two sets of input features:
  \begin{itemize}
    \item Input features for linear expert models
      \begin{equation}
        \boldsymbol{X}_{t,s} = ( \text{\textbf{Ylag}}_{t,s}, \text{\textbf{cal}}_{t}, \text{\textbf{Fund}}_{t,s}, \text{\textbf{price}}_{t} )
        \label{eq:input_features_linear}
      \end{equation}

    \item Input features for RNN models
      \begin{equation}
        \boldsymbol{X}^{\text{RNN}}_{t} = (\boldsymbol{Y}_{t}, \text{\textbf{cal}}_{t}, \text{\textbf{Fund}}_{t}, \text{\textbf{price}}_{t} )
        \label{eq:input_features_rnn}
      \end{equation}
  \end{itemize}

  \subsection{State-of-the-art linear structures}

  The linear model for all $s=0,\ldots, S-1$ is given by the following equation:
  \begin{equation}
    \begin{aligned}
      Y_{t+1,s} = \beta_{s,0} + \text{LEM}(\boldsymbol{X}_{t,s};\boldsymbol{\beta}_s) + \varepsilon_{t+1,s}
    \end{aligned}
    \label{linearreduced}
  \end{equation}
  with $\text{LEM}(\boldsymbol{X};\boldsymbol{\beta}) = \boldsymbol{X} \boldsymbol{\beta}$ representing a linear expert model (\cite{bille2023forecasting, maciejowska2024probabilistic, uniejewski2025probabilistic}), where the input features are defined as
  $\boldsymbol{X}_{t,s} = ( \text{\textbf{Ylag}}_{t,s}, \text{\textbf{cal}}_{t}, \text{\textbf{Fund}}_{t,s}, \text{\textbf{price}}_{t} )$.\\
  Note that, usually, as the notation suggests,
  $\text{\textbf{cal}}_{t}$ and $\text{\textbf{price}}_{t}$ is daily information, while $\text{\textbf{Ylag}}_{t,s}$ and $\text{\textbf{Fund}}_{t,s}$ are high-frequent (hourly) data.
  Moreover, $\text{\textbf{cal}}_{t}$ is deterministic information whereas $\text{\textbf{price}}_{t}$ and $\text{\textbf{Fund}}_{t,s}$ are stochastics.

  Here, we utilize the expert model specification from \cite{ziel2016forecasting}:
  \begin{align}
    \text{\textbf{Ylag}}_{t,s} &=  (Y_{t-1,s}, Y_{t-2,s} ,Y_{t-7,s})  \\
    \text{\textbf{cal}}_{t} &= (\text{Mon}_{t+1},\text{Sat}_{t+1},\text{Sun}_{t+1}) \in \mathbb{R}^{D_{\text{cal}}} \\
    \text{\textbf{Fund}}_{t,s} &= (\text{Load}_{t+1,s} ,\text{Wind}_{t+1,s} , \text{Solar}_{t+1,s} ) \in \mathbb{R}^{D_{\text{fund}}} \\
    \text{\textbf{price}}_{t} &= (\text{EUA}_{t-1,s} ,  \text{NGas}_{t-1,s} , \text{Oil}_{t-1,s} , \text{Coal}_{t-1,s} ) \in \mathbb{R}^{D_{\text{price}}}
    \label{detailedlm}
  \end{align}
  The calendar information $\textbf{cal}_t$ can be regarded as a one-hot encoded information of the univariate day-of-week time series, taking values from 1 to 7 characterising the corresponding weekday of time $t$.

  This may be written in reduced multivariate form as
  \begin{equation}
    \boldsymbol{Y}_{t} = \boldsymbol{\beta}_0 +
    \text{LEM}( \boldsymbol{X}_t, \textbf{B})
    +\boldsymbol{\varepsilon}_t
    \label{reducedmultivariate}
  \end{equation}
  \deleted{with $\textbf{B}$ and $\boldsymbol{X}_t $ as corresponding regression matrix which leads to equivalance to \eqref{linearreduced}.}
  \added{Equation \eqref{linearreduced} specifies the LEM model separately for each hour $s$.
    For notational convenience, the $S$ hourly equations are stacked into a reduced multivariate form in Equation \eqref{reducedmultivariate}.
    Let $\boldsymbol{Y}_t = (Y_{t,1}, \ldots, Y_{t,S})^\top$ denote the vector of electricity prices across all
    hours, $\boldsymbol{X}_t \in \mathbb{R}^{S \times K}$ the corresponding regression matrix collecting the
    $K$ explanatory variables for each hour, and $\textbf{B} \in \mathbb{R}^{K \times S}$ the matrix of
    hour-specific coefficient vectors. With these definitions, the multivariate formulation below is
  algebraically equivalent to the set of univariate hourly specifications in Equation~\eqref{linearreduced}.}

  \subsection{The Recurrent Neural Network RNN Model}
  \label{rnnmodel}

  The RNN (\cite{elman1990finding}) is a Feed-forward Neural Network in which the information is transferred from the input layer to the output layer. However, the RNN saves the output of a specific layer and connects back to the input to predict the output. More precisely, the RNN uses their internal state (memory) to process sequences of inputs with variable length. It consist of three layers: input, hidden, and output layer. Here we consider Elman regression Network given by the following equation:

  \begin{equation}
    \begin{aligned}
      \boldsymbol{H}_{t} &= \text{RNN}_{\text{cell}}\left( \boldsymbol{H}_{t-1},  \boldsymbol{X}^{\text{RNN}}_t  \right) \\
      \boldsymbol{Y}_{t} &= \mathbf{W}_{\text{out}}
      \boldsymbol{H}_{t} + \mathbf{b}_{\text{out}}
    \end{aligned}
    \label{rnneq}
  \end{equation}
  where
  $ \text{RNN}_{\text{cell}}(\boldsymbol{H}, \boldsymbol{X}^{\text{RNN}}; \mathbf{W}_{\text{hid}}, \mathbf{W}_{\text{ext}},
  \mathbf{b}_{\text{hid}}  )
  =
  \boldsymbol{g}\left( \mathbf{W}_{\text{hid}} \boldsymbol{H} + \mathbf{W}_{\text{ext}} \boldsymbol{X}^{\text{RNN}} + \mathbf{b}_{\text{hid}} \right)
  $
  and
  $\boldsymbol{X}^{\text{RNN}}_{t \in \mathbb{R}^{D}}$
  where $D=S+D_{\text{cal}}+S D_{\text{fund}}+D_{\text{price}}$, and
  $\text{\textbf{Fund}}_{t} = (\text{\textbf{fund}}_{t,0}, \ldots, \text{\textbf{fund}}_{t,S-1} )$

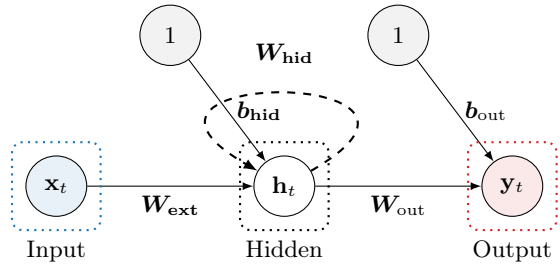
\begin{figure}[htb!]
    \centering
    \begin{tikzpicture}[>=latex, auto, font=\small,
      every state/.style={draw, circle, minimum size=1cm}]

      \node[state, fill=dblue!10] (x1) at (0,0) {$\mathbf{x}_{t}$};

      \node[state, fill=gray!10] (b0) at (1.5,2) {$1$};

      \node[state] (h1) at (3,0) {$\mathbf{h}_{t}$};

      \node[state, fill=gray!10] (b1) at (4.5,2) {$1$};

      \node[state, fill=dred!10] (y1) at (6,0) {$\mathbf{y}_{t}$};

      \draw[->] (x1) edge node[midway, below] {$\bsW_{\mathbf{ext}}$} (h1);

      \draw[->, dashed, thick] (h1) to[out=30, in=150, loop, looseness=10] (h1) node[above=1.5cm] {$\bsW_{\mathbf{hid}}$};

      \draw[->] (b0) edge node[midway, right] {$\bsb_{\mathbf{hid}}$} (h1);

      \draw[->] (h1) edge node[midway, below] {$\bsW_{\text{out}}$} (y1);

      \draw[->] (b1) edge node[midway, right] {$\bsb_{\text{out}}$} (y1);

      \node[draw=dblue,dotted, rectangle, thick, rounded corners, fit=(x1), inner sep=5pt, label=below:{Input}] (InputBox) {};
      \node[draw, dotted, rectangle, thick, rounded corners, fit=(h1), inner sep=5pt, label=below:{Hidden}] (HiddenBox) {};
      \node[draw=dred,dotted, rectangle, thick, rounded corners, fit=(y1), inner sep=5pt, label=below:{Output}] (OutputBox) {};

    \end{tikzpicture}
    \caption{Graph of a simple Elman network with one input, one hidden layer, and one output.
      The hidden state \(\mathbf{h}_{t}\) is computed from the previous state \(\mathbf{h}_{t-1}\) (dashed loop arrow) and the current input \(\mathbf{x}_{t}\).
      The output \(\mathbf{y}_{t} \in \mathbb{R}^{S}\) is then obtained through a linear transformation of the hidden state via the output weight matrix
    \(\mathbf{W}_{\text{out}} \in \mathbb{R}^{S \times h}\) and bias term \(\mathbf{b}_{\text{out}} \in \mathbb{R}^{S}\).}
    \label{fig:hidden_state}
  \end{figure}

  The hidden state at each time step \( t \) is denoted as \( \boldsymbol{H}_t \in \mathbb{R}^{H} \), where \( H \) represents the number of hidden units in the network. The hidden state is updated recursively using the previous hidden state \( \boldsymbol{H}_{t-1} \in \mathbb{R}^{H} \), the current input vector \( \boldsymbol{X}^{\text{RNN}}_{t,s} \), and the associated weight matrices.\\
  As shown in Figure \ref{fig:hidden_state}, the transition from the previous hidden state to the current state is governed by the recurrent weight matrix \( \mathbf{W}_{\text{hid}} \in \mathbb{R}^{H \times H} \), which captures temporal dependencies, and the input weight matrix \( \mathbf{W}_{\text{ext}} \in \mathbb{R}^{h \times D} \), which maps the input features into the hidden state space. A bias term \( \mathbf{b}_{\text{hid}} \in \mathbb{R}^{h} \) is also included to adjust the transformation.

  The object $ \boldsymbol{g}= (g_1,\ldots, g_H) $ represents the vector of activation function. We utilize here the ReLU (Rectified Linear Unit) activation function for $g_h$ to update the hidden state in each time step of the RNN. Where: $
  \boldsymbol{g}_h=\boldsymbol{g}_{\text{ReLU}}(z) =\text{ReLU}(z) = \max(0,z) = z\mathbbm{1}_{\{z>0\}} , \quad z \in \mathbb{R}, \quad \boldsymbol{g}_{\text{ReLU}}(z) \in [0,\infty).
  $\\
  The output of the RNN at each time step, \( \mathbf{Y}_t \in \mathbb{R}^{S} \), is computed using the hidden state transformation through an output weight matrix \( \mathbf{W}_{\text{out}} \in \mathbb{R}^{S \times h} \) and a bias term \( \mathbf{b}_{\text{out}} \in \mathbb{R}^{S} \) (see Figure \ref{fig:hidden_state}).

  \subsection{The Kalman Filter as a Special Case of RNN}
  The Kalman filter \cite{kalman1960new} is a recursive algorithm that estimates the state of a dynamic system from a series of noisy measurements. It operates in two steps: prediction and update. In the context of RNNs, the Kalman filter can be viewed as a special case where the hidden state is updated based on the previous state and the current observation, similar to the recurrent connections in RNNs.

  The Kalman filter can be expressed as follows:
  \begin{equation}
    \begin{aligned}
      \boldsymbol{H}_{t} &= \mathbf{A}_{\text{hid}} \boldsymbol{H}^{KF}_{t-1} + \mathbf{A}_{\text{ext}} \boldsymbol{Y}_{t} + \mathbf{b}_{\text{hid}} \\
      \boldsymbol{Y}_{t} &= \mathbf{A}_{\text{out}} \boldsymbol{H}^{KF}_{t} + \mathbf{b}_{\text{out}}
    \end{aligned}
    \label{kalman}
  \end{equation}
  where \( \boldsymbol{H}^{KF}_{t} \) is the hidden state at time \( t \), \( \boldsymbol{Y}_{t} \) is the observation, and \( \mathbf{A}_{\text{hid}} \), \( \mathbf{A}_{\text{ext}} \), and \( \mathbf{A}_{\text{out}} \) are weight matrices. The Kalman filter assumes a linear relationship between the hidden state and the observation, making it suitable for linear systems. Considering identity function $ \boldsymbol{g}_h= \boldsymbol{g}_{\text{id}}(z) = \text{id}(z  )= z, \quad z \in \mathbb{R}, \quad \boldsymbol{g}_{\text{id}}(z) \in \mathbb{R}$ as an activation function, instead of ReLU in the case of RNN model, the Kalman filter can be considered as a linear RNN with a specific structure.

  \added{Our implementation differs from classical Kalman filtering in three important ways: First, we use a deterministic state-space model without explicit covariance propagation. Process and measurement noise are absorbed into residuals rather than maintained as distributions. Second, parameters are learned through gradient-based optimisation rather than maximisation under distributional assumptions. Finally, the KF component is considered as a linear dynamical layer that captures smooth temporal dependencies, supporting the nonlinear RNN branch while keeping computation tractable.
    Using the identity activation, the KF branch can be viewed as a linear RNN sharing the recursive structure but operating under strict linearity constraints.
  In summary, our KF component should be interpreted as a trainable linear state-space layer; t's a Kalman "structure" rather than a fully probabilistic Kalman filter.}

  \deleted{Recently, a new sequence modeling architecture called \textit{Mamba} has been introduced as a general form of the state-space model 
    As an extension of the classical Kalman filter, the Mamba architecture provides flexible sequence modeling capabilities. In a similar way to the Kalman filter, Mamba models time-based dependencies through a hidden state that changes over time. In contrast to the Kalman filter which uses fixed matrices to define state transitions and observations, \textit{Mamba} replaces them with learnable operators that change with input data.\\
  The \textit{Mamba} model is expressed as:}
  \deleted{
    where \(\mathbf{D}_t\), \(\mathbf{E}_t\), and \(\mathbf{C}_t\) are dynamically computed from the input sequence data. When these matrices are fixed across time, such that: \(\mathbf{D}_t = \mathbf{A}_{\text{hid}}\), \(\mathbf{E}_t = \mathbf{A}_{\text{ext}}\), \(\mathbf{C}_t = \mathbf{A}_{\text{out}}\), the model reduces to the classical linear state-space equations of the Kalman filter as expressed in Equation \ref{kalman}.\\
  Based on this relationship, the Kalman filter can be viewed as a specific case within the larger Mamba architecture, which uses fixed, linear, and time-invariant matrices. Mamba extends this by allowing adaptive, input-dependent dynamics. }

  \subsection{The Parallel-Branch RNNs with Skip Connections}

\begin{figure}[H]

    \centering
    \includegraphics[width=0.9\linewidth]{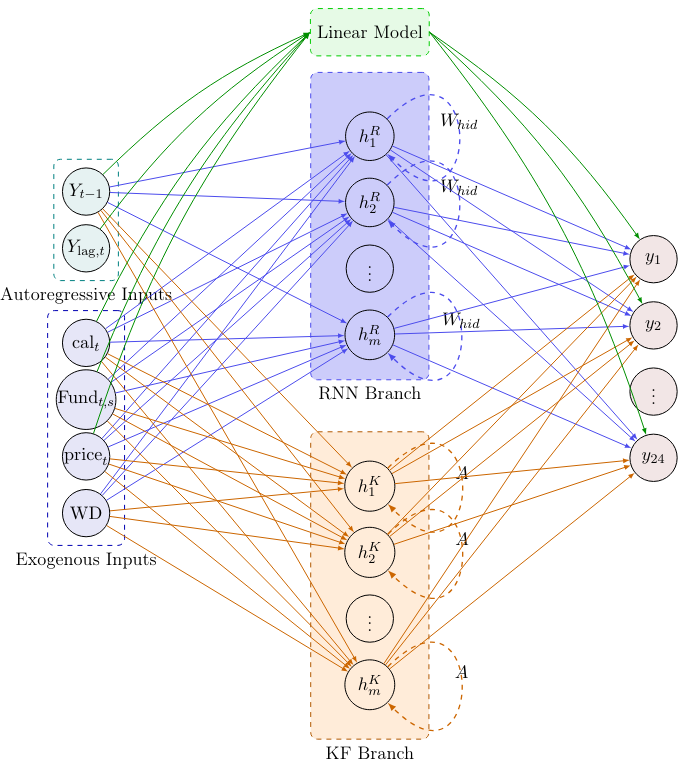}
    \caption{A Parallel-Branch Recurrent Neural Network architecture with skip connections for day-ahead electricity price forecasting. \textcolor{green}{The linear model for direct effects}, an \textcolor{blue!80!black}{RNN branch for nonlinear temporal patterns}, and \textcolor{orange}{the KF branch represents the Kalman Filter, which is for state-space dynamics}. \deleted{The outputs from the three branches are summed to produce 24-hour-ahead forecasts $(y_1, \ldots, y_{24})$}. \added{The three branches are trained jointly end-to-end; each produces its own 24-hour forecast vector, and these three vectors are then summed to yield the final 24-hour-ahead forecast $(y_1,\ldots,y_{24})$. Arrows indicate forward information flow; colored boxes correspond to the labeled branches; shared inputs are depicted with common incoming links.}}
    \label{fig:combined_architecture}
  \end{figure}

  The parallel-branch RNN architecture with skip connections is designed to enhance the model's ability to capture complex temporal dependencies and non-linear relationships in the data. As shown in Figure \ref{fig:combined_architecture}, this architecture consists of multiple parallel branches, each processing the input sequence independently and then merging their outputs. The skip connections allow for direct information flow between layers, facilitating the learning of long-range dependencies.

  The combined model will include the non-linear layer by the RNN model, linear layer with time dependencies consideration (Kalman filter, KF) and the linear model (LEM), skip connection. Therfore, the new model will combine Equation \ref{rnneq}, Equation \ref{kalman}, and \ref{linearreduced}.
  The architecture can be represented as follows:

  \begin{equation}
    \begin{aligned}
      \boldsymbol{Y}_{t+1} &= \boldsymbol{\beta}_{0} + \text{LEM}(\boldsymbol{X}_{t}, \boldsymbol{\beta})
      + \mathbf{W}_{\text{out}} \boldsymbol{H}_t^{\text{RNN}}
      + \mathbf{A}_{\text{out}} \boldsymbol{H}_t^{\text{KF}}
      + \boldsymbol{\varepsilon}_{t+1} \\
      \boldsymbol{H}_t^{\text{RNN}} &= \text{RNN}_{\text{cell}}(\boldsymbol{H}_{t-1}^{\text{RNN}}, {\boldsymbol{X}}^{\text{RNN}}_{t,s}; \mathbf{W}_{\text{hid}}, \mathbf{W}_{\text{ext}}, \mathbf{b}_{\text{hid}}^{\text{RNN}}) \\
      \boldsymbol{H}_t^{\text{KF}} &= \mathbf{A}_{\text{hid}} \boldsymbol{H}_{t-1}^{\text{KF}} + \mathbf{A}_{\text{ext}} {\boldsymbol{X}}^{\text{RNN}}_{t,s} + \mathbf{b}_{\text{hid}}^{\text{KF}}
    \end{aligned}
    \label{linearrnn_kalman}
  \end{equation}

  LEM input is represented by the vector of exogenous variables ${X}_{t,s}$ defined as indicated in Equation \ref{eq:input_features_linear}, and the RNN input and the Kalman filter input are represented by the vector of exogenous variables ${X}^{\text{RNN}}_{t,s}$ defined as indicated in Equation \ref{eq:input_features_rnn}.

  \subsection{Different Model Architectures}
  \label{sec:architectures}

  We implement and compare the following model architectures, ranging from single models to all possible hybrid and parallel combinations. Table~\ref{tab:model_architectures} provides a brief explanation of each model architecture, explaining its structure, input features, and how linear and nonlinear components are integrated to capture both the temporal and fundamental dynamics of the market.

  \begin{itemize}
    \item \textbf{Stand-alone Models:}
      \begin{itemize}
        \item Elman RNN (with ReLU): Recurrent neural network models of the Elman type, with ReLU activation as non-linear deep neural networks model (see Equation \ref{rnneq}).
        \item Elman RNN (with Identity) or Kalman filter: Recurrent neural network models of the Elman type, with identity activation , with is considerred as a linear Kalman-type recurrence model ( see Equation \ref{kalman}).
        \item Linear expert model with OLS-initialized: Baseline linear regression model using all input features and lags, with optional initialization from the ordinary least squares (OLS) solution (see , Equation \ref{linearreduced}).
      \end{itemize}

    \item \textbf{Hybrid Architectures:}
      \begin{itemize}
        \item Models with skip connection that combine LEM and RNN branches (Equation \ref{rnneq} and Equation \ref{linearreduced}).
        \item Models with parallel RNN branches (RNN and Kalman) without skip connections (Equation \ref{rnneq} and Equation \ref{kalman}).
        \item Models with parallel LEM and RNN branches (RNN and Kalman) with skip connections (Equation \ref{rnneq}, Equation \ref{kalman}, and Equation \ref{linearreduced}).
      \end{itemize}
  \end{itemize}

  \begin{table}[H]
    \centering
    \renewcommand{\arraystretch}{1.4}
    \footnotesize
    \resizebox{\textwidth}{!}{
      \begin{tabular}{>{\centering\arraybackslash}p{0.8cm}|
          >{\centering\arraybackslash}p{1.8cm}|
          >{\raggedright\arraybackslash}p{3cm}|
          >{\raggedright\arraybackslash}p{5.4cm}|
        >{\raggedright\arraybackslash}p{3cm}}
        \toprule
        \rowcolor{lightgray}
        \textbf{Type} & \textbf{Abbrev.} & \textbf{Architecture} & \textbf{Description} & \textbf{Input Features} \\
        \midrule

        \textbf{1} &
        \textbf{RNN} &
        Elman RNN \newline (ReLU) &
        No skip connections; purely nonlinear processing path. &
        \textbf{RNN Input:} \newline
        $\boldsymbol{X}^{\text{RNN}}_t$ \\
        \midrule

        \textbf{2} &
        \textbf{KF} &
        Elman RNN \newline (Identity) &
        RNN with identity activation representing a Kalman filter model. &
        \textbf{RNN Input:} \newline
        $\boldsymbol{X}^{\text{RNN}}_t$ \\
        \midrule

        \textbf{3} &
        \textbf{LEM} &
        Linear expert model \newline (OLS init) &
        Purely linear model; can be initialized via ordinary least squares (OLS). &
        \textbf{Linear Input:} \newline
        $\boldsymbol{X}_t$ \\
        \midrule

        \textbf{4} &
        \textbf{LEM-RNN} &
        Hybrid: Linear + RNN \newline (ReLU, Skip, OLS) &
        Combined model with a ReLU RNN branch and a linear skip connection. The final output is the sum of both branches. &
        \textbf{RNN Branch:} $\boldsymbol{X}^{\text{RNN}}_t$ \newline
        \textbf{Linear Skip:} $\boldsymbol{X}_t$ \\
        \midrule

        \textbf{5} &
        \textbf{KF-RNN} &
        Dual Parallel RNNs \newline (ReLU + Identity) &
        Two parallel RNNs: RNN \& KF. \deleted{Their outputs are concatenated and passed to a final linear layer.} \added{The final output is the sum of both branches.} &
        \textbf{RNN Inputs:} $\boldsymbol{X}^{\text{RNN}}_t$ \\
        \midrule

        \textbf{6} &
        \textbf{LEM-KF-RNN} &
        ReLU + Identity + Linear, OLS &
        Allowing both nonlinearand linear components to contribute to the forecast. &
        \textbf{Dual RNN:} $\boldsymbol{X}^{\text{RNN}}_t$ \newline
        \textbf{Linear Skip:} $\boldsymbol{X}_t$ \\

        \bottomrule
      \end{tabular}
    }
    \caption{Overview of implemented model architectures, their structures, and input features.}
    \label{tab:model_architectures}
  \end{table}

  \section{Training and Forecasting Study, and Evaluation Metrics}
  \label{training_section}
  \subsection{Data preprocessing}
  Data preprocessing ensures that the forecasting framework has appropriate inputs, including temporal consistency and feature completeness. A dataset of hourly day-ahead power prices along with relevant explanatory variables is available for Germany. First, to accommodate daylight saving time (DST) shifts, a proprietary transformation aligned all hourly data to a consistent temporal grid, ensuring uniform spacing and comparability between seasons.
  Second, the dataset was restructured into a daily-by-hour matrix representation, with each row corresponding to a calendar day and each column to a specific hour of the day.
  As a result, a regressor matrix was built by aligning the endogenous variable with its associated regressors. We removed the rows with missing values that resulted from lag construction.\\
  The matrix formed here will serve as the basis for training the model. Finally, we standardise all variables using the mean/standard deviation to ensure comparable scales of contribution from all variables. During each iteration of the rolling-window evaluation, train-only standardisation is applied to both input features and target variables, i.e., the mean ($\mu_{\text{train}}$) and ($\sigma_{\text{train}}$) are calculated based only on the available data on the rolling window, which is important to ensure no data leakage. Through the standardisation process, each variable is transformed so that it has a zero mean and unit variance as follows:
  \begin{equation}
    \label{eq:standardization}
    \boldsymbol{x}_i^{\text{std}} = \frac{\boldsymbol{x}_i - \mu_{\text{train}}}{\sigma_{\text{train}}}, \quad
    \boldsymbol{y}_i^{\text{std}} = \frac{\boldsymbol{y}_i - \mu_{\text{train}}^{(y)}}{\sigma_{\text{train}}^{(y)}}
  \end{equation}
  Where $\boldsymbol{x}^{\text{std}}_i$ represents the standardised feature matrix and $\boldsymbol{y}^{\text{std}}$ denotes the target value vector. It is necessary to perform this standardisation step to optimise model stability and convergence during training.
  In order to return predictions to their original units (e.g., €/MWh), the standardised forecasts of the target value, associated with the model outputs, are de-standardised. The inverse transformation is presented as follows:
  \begin{equation}
    \label{eq:destandardization}
    \boldsymbol{\hat{y}_i} = \boldsymbol{\hat{y}_i^{\text{std}}} \cdot \sigma_{\text{train}}^{(y)} + \mu_{\text{train}}^{(y)}
  \end{equation}
  This method guarantees that evaluation metrics like RMSE, MAE, etc., are calculated in their original scale, facilitating a significant and authentic evaluation of forecasting precision in genuine out-of-sample scenarios.
  \subsection{Optimization of the objective functions}
  In our forecasting methodology, we use a dual-level optimization approach with an inner learning process for estimating model parameters and an outer optimisation process for determining the optimal configuration of hyperparameters. In the following subsections, each objective function and its associated estimation process are described in detail.
  \subsubsection{Estimation and Update of the models parameters}

  As shown in Equation \ref{linearrnn_kalman}, a number of parameters will be optimised. Depending on the model type, including recurrent weights that connect hidden states over time, such as hidden to hidden connections ($ \mathbf{W}_{\text{hid}}$ and $ \mathbf{A}_{\text{hid}}$ for RNN and KF, respectively), hidden to output weights ($ \mathbf{W}_{\text{out}}$ and $ \mathbf{A}_{\text{out}}$), as well as bias terms. Weights from our linear expert model, initialised using OLS estimates, will be also optimised.\\
  In order to estimate the neural component parameters, supervised gradient-based learning is employed within a rolling window approach, while LEM coefficients are initialised with OLS estimates rather than learned directly from gradient updates.
  The model generates forecasts $\boldsymbol{\hat{Y}}^{std}_t=f(\boldsymbol{X}^{std}_t, \theta)$ for a given set of standardised inputs $\boldsymbol{X}^{std}_t$ and target outputs $\boldsymbol{Y}^{std}_t$, where $\boldsymbol{\theta}$ denotes the set of all trainable parameters (weights and biases) in the network.

  Over the training window, the parameters are estimated by minimising the following loss function:
  \begin{equation*}
    \boldsymbol{\mathcal{L}}_{\text{train}}(\theta)
    =
    \underbrace{
      \frac{1}{N} \sum_{t=1}^{N}
      \| \hat{y}_t - y_t \|_2^2
    }_{\text{Mean Squared Error (MSE)}}
    +
    \underbrace{
      \lambda_{1} \sum_{i,j} |w_{ij}|
    }_{\text{L1 Regularization (Sparsity)}}
    +
    \underbrace{
      \lambda_{2} \sum_{i,j} w_{ij}^2
    }_{\text{L2 Regularization (Weight Decay)}}.
  \end{equation*}
  where L1 regularisation (LASSO) penalises only output-side linear mappings, promoting sparsity and interoperability, while L2 regularisation (Ridge) penalises overfitting by implementing weight decay in the optimiser.\\
  The parameters are updated using the Adam optimisation algorithm (\cite{adam2014method}), which is an adaptive variation of stochastic gradient descent that estimates the gradients' first and second moments to dynamically update learning rates.\\
  At each iteration $t$, the parameters are updated as follows:
  \[
    m_t = \gamma_1 m_{t-1} + (1 - \gamma_1) g_t, \quad
    v_t = \gamma_2 v_{t-1} + (1 - \gamma_2) g_t^2, \quad
    \theta_{t+1} = \theta_t - \eta \frac{m_t}{\sqrt{v_t} + \epsilon}.
  \]
  Where $\boldsymbol{g}_t = \nabla_{\theta_t} \boldsymbol{\mathcal{L}}_{\text{train}}$ denotes the gradient of the loss, and ${\eta}$ is the learning rate.
  We implement ReduceLROnPlateau scheduling, which reduces the learning rate automatically (by a half) when the validation loss stagnates for five epochs. In addition, we adopt a gradient clipping of $\|\boldsymbol{g}_t\|_2 \leq 5$ which minimises the risk of gradients exploding in recurrent components.\\
  As the training proceeds, rolling windows are simulated, mimicking real-time learning.
  Models are retrained each forecast day based on fixed-length windows of recent observations. Model types differ in terms of how the weights corresponding to the initial window ($D_{init}$) are initialised, with RNN models starting with small uniform random weights and LEM models starting with OLS-based warm-starts.\\
  Once convergence has been achieved for the current window, the optimised weights are saved as initialisation weights for the next window ($D_{all}$).
  While maintaining learned temporal dependencies, this sequential re-estimation enables the parameters to change smoothly across time.
  We thus implement a continuously adaptable model in the estimation process by combining serial parameter transfer between windows with stochastic gradient-based learning within each window.

  \subsubsection{Optimization and Hyperparameter Training}

  The previous section explains how the model's (internal) weights ($\theta$) are estimated and updated on a rolling basis. The following section will discuss hyperparameter optimisation, which is an outer optimisation procedure that optimises a set of hyperparameters ($\phi$), responsible for the model's architecture and learning dynamics.

  \begin{table*}[ht]
    \centering
    \caption{Overview of tuned hyperparameters and their search spaces used in Optuna (TPE) optimization.}
    \label{tab:hyperparams}
    \resizebox{\textwidth}{!}{%
      \begin{tabular}{llcl}
        \toprule
        \textbf{Name} & \textbf{Symbol} & \textbf{Type / Interval} & \textbf{Description} \\
        \midrule
        Hidden layer size & $H$ & $\{1, 2, \dots, 128\}$ & Number of neurons in each RNN branch. \\[4pt]
        Sequence length & $L$ & $\{1, 2, \dots, 7\}$ & Number of past days fed into the RNN. \\[4pt]
        Initial training window size & $D_{\mathrm{init}}$ & $[30, 730]$ & Number of days used for the first rolling forecast window. \\[4pt]
        Updated training window size & $D_{\mathrm{all}}$ & $[2, 365]$ & Number of days used for subsequent rolling windows. \\[4pt]
        Initial epochs & $E_{\mathrm{init}}$ & $\{10, 20, 50, 100\}$ & Number of training epochs for the initial window. \\[4pt]
        Updated epochs & $E_{\mathrm{all}}$ & $\{5, 10, 20, 50\}$ & Number of training epochs for each updated rolling step. \\[4pt]
        Initial learning rate & $\eta_{\mathrm{init}}$ & $[10^{-5}, 10^{-2}]$ & Adam learning rate for the initial forecast. \\[4pt]
        Updated learning rate & $\eta_{\mathrm{all}}$ & $[10^{-4}, 10^{-2}]$ & Adam learning rate for rolling updates. \\[4pt]
        Initial weight decay ($L_2$) & $\lambda_{w,\mathrm{init}}$ & $[10^{-8}, 10^{-2}]$ & $L_2$ regularization for the initial window. \\[4pt]
        Updated weight decay ($L_2$) & $\lambda_{w,\mathrm{all}}$ & $[10^{-8}, 10^{-2}]$ & $L_2$ regularization for subsequent windows. \\[4pt]
        Initial $L_1$ regularization & $\lambda_{1,\mathrm{init}}$ & $[10^{-6}, 10^{-1}]$ & $L_1$ penalty applied in the initial window. \\[4pt]
        Updated $L_1$ regularization & $\lambda_{1,\mathrm{all}}$ & $[10^{-6}, 10^{-1}]$ & $L_1$ penalty applied in rolling updates. \\[4pt]
        OLS initialization weight & $\alpha$ & $[0, 2]$ & Scaling factor for OLS-based weight initialization. \\[4pt]
        OLS initialization flag & \texttt{use\_ols\_weights} & $\{\text{True}, \text{False}\}$ & Enables warm-start from OLS weights (skip/linear models). \\[4pt]
        Batch size & $B$ & $\{8, 16, 32, 64\}$ & Mini-batch size used during optimization. \\[4pt]
        Gradient clip norm & $c_{\mathrm{grad}}$ & $[0.1, 10]$ & Maximum gradient norm for clipping (stability). \\[4pt]
        Dropout rate & $p_{\mathrm{drop}}$ & $[0.0, 0.5]$ & Dropout applied to hidden layers for regularization. \\[4pt]

        \bottomrule
      \end{tabular}%
    }
  \end{table*}
  An overview of the adopted key hyperparameters is explained in Table \ref{tab:hyperparams}. These hyperparameters are tuned using \texttt{optuna} (\cite{akiba2019optuna}) with the Tree-structured Parzen Estimator (TPE) sampler for Bayesian optimization, the models are implemented in PyTorch (\cite{paszke2019pytorch}), which efficiently explores the search space. We also adopt a rolling window forecasting approach (as described in Section \ref{sec:data}) where a search space is designed to strike a balance between computing efficiency, convergence stability, and model adaptability.\\
  For a given \texttt{optuna} trial $k$, the entire model training and validation cycle is evaluated based on a specific hyperparameter configuration $\phi_k$.
  A trial's performance is determined by its Root Mean Square Error (RMSE), which is calculated over all forecast hours and evaluation days as follows:
  \[
    \boldsymbol{\mathcal{L}}_{\text{train}}(\phi_k)
    =
    \text{RMSE}_{\text{overall}}
    =
    \sqrt{
      \frac{1}{S T}
      \sum_{s=1}^{S} \sum_{t=1}^{T}
      \left( \hat{y}_{s,t} - y_{s,t} \right)^2
    }.
  \]
  where $S$ denotes the number of hours of the day $S=0,1,...,23$, and $T$ represents the number of days in the evaluation period (rolling window).\\
  The set of tuned hyperparameters $\phi$ consist of:\\
  \textit{The hyperparameters of the RNN architecture} are hidden layer size ($H$) and sequence length ($L$). RNNs with larger hidden dimensions can capture complex nonlinear dynamics. Sequence length determines how many past observations are available to learn temporal dependencies (ranging here from one day to one week).\\
  \textit{The size of the training dataset} is determined by the data-window parameters $D_{\mathrm{init}}$ and $D_{\mathrm{all}}$ in initial and subsequent rolling estimation. This allows our models to take into consideration the structural changes in the underlying data-generating process as additional information is made available, ensuring that it evolves adaptively as new information becomes available.\\
  We consider two \textit{training schedule parameters}: $E_{\mathrm{init}}$ and $E_{\mathrm{all}}$, which specify how many optimisation epochs are allocated to each window. The initial window has a greater number of epochs to ensure stable convergence, while successive windows have fewer training to accelerate updates as previously learned weights are refined.\\
  Concerning \textit{the optimisation hyperparameters}, we adopt the learning rate of the Adam optimiser $\eta_{\mathrm{init}}$ and $\eta_{\mathrm{all}}$, which scale the gradient $ \nabla_{\theta} \boldsymbol{\mathcal{L}}$ during parameter updates, controlling the step size in the optimisation space. The optimiser can progress more quickly towards a minimum with a high learning rate ($\eta$) value, but it risks skipping it, which may result in unstable training or even divergence. A slower training process may arise from a lower learning rate, even while it guarantees smoother, more stable convergence.\\
  In addition, to improve generalisation and control model complexity, our loss function integrates both Ridge ($L2$) and Lasso ($L1$) regularisation terms. The result is an Elastic Net penalty that balances shrinkage and sparsity as follows:
  $
  \mathcal{L}_{\text{total}} = \mathcal{L}_{\text{MSE}} + \lambda_w \|\mathbf{W}\|_2^2 + \lambda_1 \|\mathbf{W}\|_1,
  $
  Where MSE loss function is defined as follows:
  $
  \mathcal{L}_{\text{MSE}} = \frac{1}{N} \sum_{i=1}^{N} (y_i - \hat{y}_i)^2,
  $
  $\|\mathbf{W}\|_2^2 = \sum_j w_j^2$ denotes the weight decay (ridge) penalty, and $\|\mathbf{W}\|_1 = \sum_j |w_j|$ represents the sparsity of the L1 penalty. The coefficients $\lambda_w$ and $\lambda_1$ control the strength of each term separately for initial and rolling windows.\\
  For models with linear or skip connections (model types 3, 4, and 6), the linear branch is warm started using coefficients estimated from OLS. OLS based initialisations are controlled by the scaling factor $\alpha$ and the Boolean flag \texttt{use\_ols\_weights}. Note that:
  $
  \mathbf{W}_{\text{OLS}} = (\mathbf{X}^{\top} \mathbf{X})^{-1} \mathbf{X}^{\top} \mathbf{y},
  $ and
  $
  \mathbf{W}_{\text{init}} = \alpha \cdot \mathbf{W}_{\text{OLS}},
  $
  where $\alpha \in [0,2]$ adjusts the relative contributions from the OLS solution to the initial weight configuration, which facilitates faster convergence and ensures stable results.\\
  There are \textit{three training control parameters}: batch size $B$, gradient clipping norm $c_{\mathrm{grad}}$, dropout probability $p_{\text{drop}}$, and activation type $\sigma$. Euclidean gradients are constrained by gradient clipping, satisfying:
  $
  \|\nabla_{\theta} \boldsymbol{\mathcal{L}}\|_2 \leq c_{\mathrm{grad}},
  $
  which improves recurrent architectures by mitigating exploding gradients. Dropout deactivates hidden neurons randomly with probability $p_{\text{drop}}$, resulting in enhanced generalisation due to a reduction in co-adaptation.\\
  The objective is to minimize the RMSE. Each \texttt{optuna} trial runs for 500 iterations, evaluating different hyperparameter configurations. The best-performing set is then selected.
  In the rolling window training approach, the best hyperparameters found in the first window using \texttt{optuna} are stored and reused as starting values for subsequent windows instead of being randomly selected (See Figure \ref{fig:rolling_window}).
  Firstly, the model is initialized with random weights (or with OLS weights), but after each iteration, the best hyperparameters are saved and loaded into the next estimation window. In addition to ensuring continuity in learning, this approach accelerates training and improves stability by retaining previously learned information. As the model evolves rather than starting from scratch, it effectively captures evolving patterns in our data.

  \subsubsection{Out-of-sample rolling window forecasting}
  After optimising the weights of the models ($\theta^*$) and the hyperparameter configuration ($\phi^*$), the next step consists of computing out-of-sample forecasts using $\phi^*$ and $\theta^*$. Our test sample consists of the last two years of the dataset (730 days).
  During testing, the rolling-window procedure is followed, ensuring that each forecast is produced using only the most recent information available. Based on the most recent historical data, the model receives standardised input features $\boldsymbol{X}^{std}_t$, and produces standardised predictions $\boldsymbol{\hat{Y}}^{std}_t=f(\boldsymbol{X}^{std}_t, \theta^*)$. These forecasts are then de-standardised (as shown in Equation \ref{eq:destandardization}) to be transformed back to their original scale and compared to the corresponding real values in the test sample.\\
  Hybrid architectures using LEM and/or multiple RNN components, such as model types 4–6, compute the final forecast as the sum of several standardised sub-forecasts associated with the model's internal components. Let $\boldsymbol{\hat{Y}}^{std}_t(c)$ be the standardised forecast produced by component $c$ (i.e, LEM, ReLU-RNN, or identity-RNN branches, i.e., KF branch). After combining these components, we obtain the total standardised forecast:
  $
  \hat{Y}_{t}^{\text{std}} = \sum_{c} \hat{Y}_{t}^{\text{std}}(c)
  $
  To restore forecasts at their original scale, the aggregated standardised output is de-standardised based on the mean and standard deviation determined from the corresponding training window:
  \begin{equation}
    \hat{Y}_{t} =
    \left( \sum_{c} \hat{Y}_{t}^{\text{std}}(c) \right)
    \sigma_{y}^{(\text{train})}
    + \mu_{y}^{(\text{train})}.
    \label{unstadard_comb}
  \end{equation}
  By analysing the forecasts of stand-alone models, we can quantify the contributions of linear and nonlinear components to the overall forecast, which enables a transparent evaluation of the contributions of each sub-model to the final forecast.

  \subsubsection{Forecasting evaluation}
  \label{sec:forecasting_evaluation}
  To evaluate the forecasting performance of the proposed models, a variety of evaluation metrics are used, in addition, we assess their statistical significance.

  \paragraph{Evaluation metrics:}\

  We utilise two widely accepted metrics: Root Mean Squared Error (RMSE) and Mean Absolute Error (MAE). This metric quantifies the average magnitude of errors between predicted and actual electricity prices, providing insight into the accuracy of the forecast. Furthermore, the Naive weekly model is used to calculate the relative Mean Absolute Error (rMAE).

  The model predicts electricity prices for the next 24 hours, generating an output vector: $\widehat{\boldsymbol{Y}}_t=(\widehat{Y}_{t,0}, \widehat{Y}_{t,1}, \ldots, \widehat{Y}_{t,S-1})$.\\
  To evaluate the accuracy of the model in the test sample, forecasted prices $\widehat{\boldsymbol{Y}}_t$ are compared with the actual prices $\boldsymbol{Y}_t$ from the test set. Then RMSE and MAE over the test period are calculated as fellow:
  \begin{align}
    \text{RMSE} = \sqrt{ \frac{1}{ST} \sum_{t=1}^{T} \sum_{s=0}^{S-1}  \left( {Y}_{t,s} - \hat{Y}_{t,s} \right)^2 } \ , \
    \text{MAE} = \frac{1}{ST} \sum_{t=1}^{T} \sum_{s=0}^{S-1}  | {Y}_{t,s} - \hat{Y}_{t,s} |
    \label{mae}
  \end{align}
  Where $T$ is the number of days in the test sample, \( \hat{Y}_{t,s} \) is the forecasted price for day $t$ at hour $s$ and ${Y}_{t,s}$ is the actual price (real price value).
  As well as the RMSE and the MAE for every individual hour:
  \begin{align}
    \text{RMSE}_s = \sqrt{ \frac{1}{T} \sum_{t=1}^{T}   \left( {Y}_{t,s} - \hat{Y}_{t,s} \right)^2 } \ , \
    \text{MAE}_s = \frac{1}{T} \sum_{n=1}^{T}   | \hat{Y}_{t,s} - Y_{t,s} |
    \label{mae_s}
  \end{align}

  The RMSE and MAE provide insights into the model's performance, with lower values indicating better accuracy.\\
  The relative MAE (rMAE) (\cite{hyndman2006another}) normalises the error by the MAE obtained from a naive forecasting model. Likewise, naive forecasts are constructed using out-of-sample data. In this study, we applied the weekly Naive model, defined as follows:
  \begin{equation}
    \hat{Y}_{t,s}^{\text{naive}} = Y_{t-7,s}
    \label{naive}
  \end{equation}
  This choice is motivated by the fact that it is easy to compute, and unlike the rMAE based on random walk model ($\hat{Y}_{t,s}^{\text{naive,1}} = Y_{t-1,s}$), it captures weekly effects.

  Hence, the rMAE is described as:
  \[
    \text{rMAE} = \frac{\text{MAE}}{\text{MAE}_{\text{naive}}}
  \]
  We also consider the relative RMSE ($\text{rRMSE}_{\text{naive}}$), defined similarly as fellows:
  \[
    \text{rRMSE}_{\text{naive}} = \frac{\text{RMSE}}{\text{RMSE}_{\text{naive}}}
  \]
  where $\text{RMSE}_{\text{naive}}$ is the RMSE of the naive weekly model defined in Equation \ref{naive}.

  \added{We additionally report the relative root mean square error with respect to the mean of the actual values ($\text{rRMSE}_\text{mean} $), defined as:}
  \[
    \text{rRMSE}_\text{mean} = \frac{\text{RMSE}}{\bar{Y}}
  \]

  \added{where $\bar{Y}$ represents the arithmetic mean of electricity prices during the test period.\\
  The $\text{rRMSE}_\text{mean} $rRMSE provides a scale-free, normalized measure of performance relative to average price levels. In particular, this normalization allows for direct comparison of forecasting accuracy across models and market conditions, especially when electricity prices show strong level shifts or regime changes. For example, an $\text{rRMSE}_\text{mean} $ of 0.20 means the typical forecasting error is 20\% of the mean price. Moreover, reporting $\text{rRMSE}_\text{mean} $ makes it easier to compare results with existing studies and interpret them effectively. In the literature on energy and environmental forecasting, such relative error measures are widely recommended to improve the robustness and interpretability of forecasts.}

  \added{Besides the above error metrics, we also consider the Directional Accuracy (DA) metric, which evaluates the model's ability to predict the direction of price movements correctly between two successive time points. The DA is defined as follows:}

  \begin{equation}
    \text{DA}_h = \frac{1}{T-1} \sum_{t=2}^{T} \mathbb{1}\left[\text{sign}(Y_{t,h} - Y_{t-1,h}) = \text{sign}(\hat{Y}_{t,h} - Y_{t-1,h})\right]
  \end{equation}

  \added{where $\text{sign}(\cdot)$ returns the sign of the argument ($+1$, $0$, or $-1$). $\mathbb{1}[\cdot]$ is the indicator function that equals 1 if the condition is true, 0 otherwise.\\
    The overall directional accuracy is computed as the average across all hours:
    $ \text{DA} = \frac{1}{S} \sum_{h=1}^{S} \text{DA}_h$,
    where $S = 24$ is the number of hours per day.\\
    A value of DA $= 1.0$ (100\%) indicates perfect directional accuracy, which means the model always correctly predicts whether prices will increase or decrease.\\
    For electricity price forecasting, this metric is particularly useful because correctly predicting the direction of price movements is often as important as predicting their precise magnitude, especially for trading
  strategies. It is important for market participants to decide if prices are likely to go up or down before making bidding, hedging, and dispatch decisions. In addition to point-error measures, DA provides a more practical evaluation of forecast performance and demonstrates the model’s ability to predict market trends.}

  \paragraph{Evaluation Based on Statistical Testing}\
  \label{stat-tests}

  The analysis of evaluation metrics must also be accompanied by consideration of whether any difference in accuracy is statistically significant. In other words, to determine whether forecast accuracy differences are real or just random differences between forecasts, we need to perform statistical tests. Diebold-Mariano (DM) test (\cite{diebold2002comparing}) and Giacomini-White (GW) test (\cite{giacomini2006tests}) have been widely used to test the statistical significance of electricity price forecasting. However, the GW test is considered preferable as it can be viewed as a generalisation of the DM test (\cite{lago2021forecasting}, \cite{lopez2025forecasting}, and \cite{marcjasz2020neural}).\\
  In our research, a multivariate GW test is performed jointly for all hours using the multivariate loss differential series or the daily loss differential series defined as:
  \begin{equation}
    \Delta_d^{\mathrm{A}, \mathrm{B}}=\left\|\varepsilon_d^{\mathrm{A}}\right\|_p-\left\|\varepsilon_d^{\mathrm{B}}\right\|_p,
    \label{eqmultivariateDM}
  \end{equation}
  where $\|\cdot\|_1$ denotes the $L_1$ norm.
  For each day, the error vector $\boldsymbol{\varepsilon}_d$ is 24-dimensional, with:
  $
  \boldsymbol{\varepsilon}_{d} = (\varepsilon_{d,1}, \varepsilon_{d,2}, \dots, \varepsilon_{d,24})
  $
  and $\varepsilon_{d,h} = P_{d,h} - \hat{P}_{d,h}$, where $P_{d,h}$ is the observed value and $\hat{P}_{d,h}$ the forecasted value.
  Based on Equation \ref{eqmultivariateDM}, the GW test yields a single $p$-value assessing the null hypothesis:
  \begin{equation}
    H_{0}: \mathbb{E}(\Delta_{A,B}) \leq 0
  \end{equation}
  which indicates that model A has lower forecasting error than model B. The alternative hypothesis is:
  \begin{equation}
    H_{A}: \mathbb{E}(\Delta_{A,B}) \geq 0
  \end{equation}

  \subsubsection{Reproducibility and Code Availability}
  \added{The code is available on GitHub} \footnote{https://github.com/souhirbenamor/RNN-for-EPF} \added{under an open-source license and runs in Visual Studio Code using Python 3.12.8. A number of Python libraries are used, including NumPy, pandas, PyTorch, Optuna, and scikit-learn. Furthermore, a custom utility file (\texttt{my\_function.py}) is required to handle time-related preprocessing, including daylight saving time (DST) adjustments, and must be located in the same repository directory as the project. To facilitate full reproducibility, the cleaned data for the Germany-Luxembourg electricity market has also been included.\\
  At the top of the main script, the model architecture (single or hybrid) can be selected by modifying the \texttt{Model\_type} variable, and all results are automatically saved in the output directory with the corresponding model identifier.}

  \section{Empirical Results and discussion}
  \label{results_section}
  Empirical results, including hyperparameter optimisation outcomes and forecasting performance evaluation, are presented and discussed in this section.

  \subsection{Model Explainability: LEM component and feature importance}

\begin{figure}[h!]
    \centering
    \includegraphics[width=0.9\textwidth]{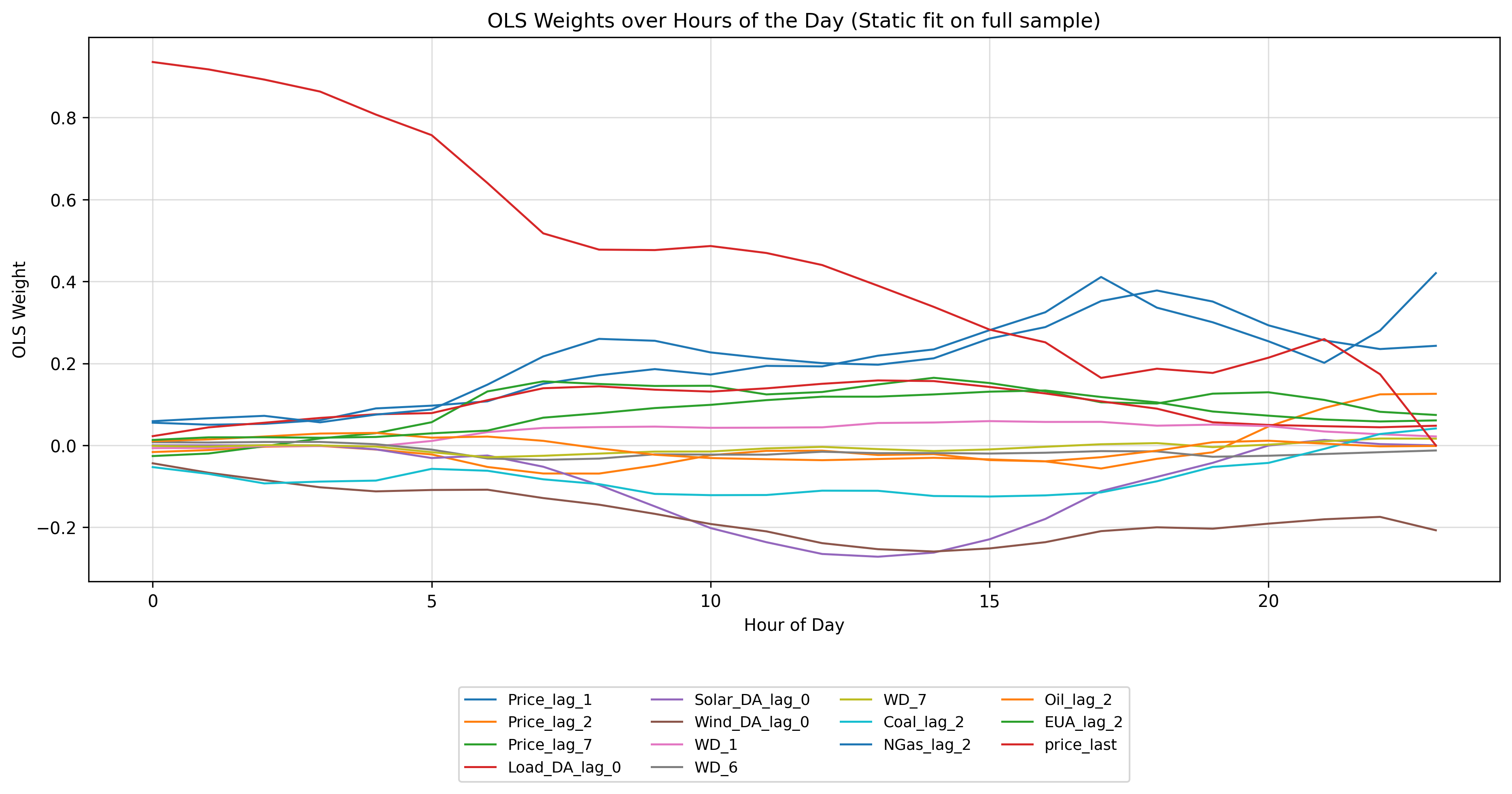}
    \caption{\added{Features importance from the LEM component of the LEM-KF-RNN model across all hours.}}
    \label{fig:ols_weights_hour}
  \end{figure}

  Figure \ref{fig:ols_weights_hour} captures the variation in the influence of relevant variables on electricity prices throughout the day based on the linear estimation component (LEM) estimations from Model LEM-RNN-KF. The highest impact is attributed to the last price of the previous day. It shows a high influence weight in the early hours and decreases in the evening hours. The lagged prices maintain a moderate positive weight, while the installed loads have a steady positive influence all day. Wind and solar generation show negative influence weights, thus suggesting a price-reducing impact. Similarly, the variable solar generation maintains a position of consistently negative weight over the day, especially between 10 am and 15 am. This aligns with standard expectations of the price-suppressing impact of solar generation during daylight hours. The fuel and carbon price variables, namely oil, gas, coal, and EUA, have influence weights that are close to zero, suggesting a weak influence in the short run.\\

  \added{To further investigate the stability of feature in the hybrid model, Table \ref{tab:merged_windows_dailyavg_base_updated} analyses the LEM coefficients estimated initially using the OLS warm start (Pre OLS) and their values following joint fine-tuning with the RNN/KF through gradient descent (Post GD). The results are provided for selected rolling windows, including the beginning and end of the test sample, midpoints of each test year, and the average values from all windows.\\
    We noticed that the LEM coefficients for the most economically relevant variables remain stable during joint training (comparing Pre OLS and Post GD values). More precisely, the dominant drivers of electricity prices retain their expected economic signs. The load coefficient remains positive across windows before and after joint training, confirming that higher expected demand consistently increases electricity prices. A negative coefficient continues to be observed for wind generation, consistent with the merit-order effect, whereby an increase in renewable supply results in a decrease in electricity prices. Similarly, solar generation has a negative coefficient, albeit smaller in magnitude than wind generation, owing to the absence of solar generation at night.
    After joint training, we observed a decrease in the magnitudes of the coefficients. However, this reduction does not imply a loss of economic meaning (a change in sign). Instead, it shows a shift in explanatory power.
    This change results from a reallocation of linear variation, initially captured by OLS, that is subsequently absorbed by RNN/KF. Specifically, the RNN/KF branch records nonlinear dynamics and regime-dependent dynamics. In this way, the hybrid architecture fulfils its intended purpose: the linear foundation is provided by the LEM, while the nonlinear, regime-dependent dynamics are modelled by the recurrent branch.\\
    Certain secondary variables, including fuel prices and calendar effect dummies, demonstrate less stability across rolling windows. These variables typically have smaller magnitudes and exhibit greater collinearity (see Figure \ref{fig:ols_weights_correlation}, in the Appendix). Occasional sign changes are most plausibly attributed to overlapping economic information, rather than to economically implausible behaviour.\\
  In summary, the results in Table \ref{tab:merged_windows_dailyavg_base_updated} indicate that joint optimization does not substantially affect the economic interpretation of the primary regressors identified by the LEM model. This finding supports the interpretability of the proposed hybrid model.}

  \begin{table}[!htbp]
    \centering
    \scriptsize
    \renewcommand{\arraystretch}{1.15}
    \setlength{\tabcolsep}{3pt}

    \begin{adjustbox}{max width=\textwidth}
      \begin{tabular}{%
          l @{\hspace{16pt}}
          rr @{\hspace{16pt}}
          rr @{\hspace{16pt}}
          rr @{\hspace{16pt}}
          rr @{\hspace{16pt}}
          rr @{\hspace{16pt}}
          rr
        }
        \toprule
        & \multicolumn{2}{c}{\shortstack{Start of Test sample\\(2022-01-15)}}
        & \multicolumn{2}{c}{\shortstack{Year1\_Mid\\(2022-07-13)}}
        & \multicolumn{2}{c}{\shortstack{Year2\_Mid\\(2023-07-13)}}
        & \multicolumn{2}{c}{\shortstack{Year3\_Mid\\(2024-07-12)}}
        & \multicolumn{2}{c}{\shortstack{End of Test sample\\(2025-01-10)}}
        & \multicolumn{2}{c}{\shortstack{Average Coefficients\\over Selected Windows}} \\
        \cmidrule(lr){2-3} \cmidrule(lr){4-5} \cmidrule(lr){6-7}
        \cmidrule(lr){8-9} \cmidrule(lr){10-11} \cmidrule(lr){12-13}
        Feature
        & Pre OLS & Post GD
        & Pre OLS & Post GD
        & Pre OLS & Post GD
        & Pre OLS & Post GD
        & Pre OLS & Post GD
        & Pre OLS & Post GD \\
        \midrule

        Coal\_lag\_2
        & \cellcolor{red!1}-0.003 & \cellcolor{green!5}0.044
        & \cellcolor{red!7}-0.057 & \cellcolor{green!1}0.011
        & \cellcolor{green!5}0.042 & \cellcolor{red!1}$-2.2\cdot 10^{-4}$
        & \cellcolor{green!4}0.038 & \cellcolor{red!1}-0.002
        & \cellcolor{red!5}-0.047 & \cellcolor{red!1}-0.006
        & \cellcolor{red!1}-0.005 & \cellcolor{green!1}0.009 \\

        EUA\_lag\_2
        & \cellcolor{green!3}0.027 & \cellcolor{green!10}0.090
        & \cellcolor{green!1}0.009 & \cellcolor{green!3}0.025
        & \cellcolor{green!3}0.026 & \cellcolor{red!1}-0.002
        & \cellcolor{red!2}-0.014 & \cellcolor{green!1}0.002
        & \cellcolor{green!1}0.012 & \cellcolor{green!1}0.001
        & \cellcolor{green!1}0.012 & \cellcolor{green!3}0.023 \\

        Load\_DA\_lag\_0
        & \cellcolor{green!35}0.298 & \cellcolor{green!15}0.125
        & \cellcolor{green!30}0.256 & \cellcolor{green!5}0.043
        & \cellcolor{green!51}0.439 & \cellcolor{green!1}0.006
        & \cellcolor{green!32}0.275 & \cellcolor{green!1}$5.4\cdot 10^{-4}$
        & \cellcolor{green!39}0.334 & \cellcolor{green!1}0.001
        & \cellcolor{green!37}0.320 & \cellcolor{green!4}0.035 \\

        NGas\_lag\_2
        & \cellcolor{green!49}0.424 & \cellcolor{green!22}0.186
        & \cellcolor{green!37}0.318 & \cellcolor{green!12}0.100
        & \cellcolor{green!2}0.014 & \cellcolor{green!1}0.010
        & \cellcolor{red!8}-0.069 & \cellcolor{red!2}-0.020
        & \cellcolor{green!9}0.081 & \cellcolor{red!1}-0.004
        & \cellcolor{green!18}0.154 & \cellcolor{green!6}0.055 \\

        Oil\_lag\_2
        & \cellcolor{red!6}-0.054 & \cellcolor{green!1}0.008
        & \cellcolor{green!1}0.001 & \cellcolor{red!1}-0.012
        & \cellcolor{red!8}-0.072 & \cellcolor{red!1}-0.004
        & \cellcolor{red!9}-0.077 & \cellcolor{red!1}-0.005
        & \cellcolor{red!8}-0.067 & \cellcolor{red!1}-0.004
        & \cellcolor{red!6}-0.054 & \cellcolor{red!1}-0.003 \\

        Price\_lag\_1
        & \cellcolor{green!10}0.089 & \cellcolor{green!14}0.117
        & \cellcolor{red!1}$-2.1\cdot 10^{-4}$ & \cellcolor{green!1}0.010
        & \cellcolor{green!1}0.008 & \cellcolor{green!1}0.005
        & \cellcolor{red!6}-0.049 & \cellcolor{red!1}-0.004
        & \cellcolor{green!24}0.209 & \cellcolor{green!1}0.008
        & \cellcolor{green!6}0.051 & \cellcolor{green!3}0.027 \\

        Solar\_DA\_lag\_0
        & \cellcolor{red!14}-0.117 & \cellcolor{red!11}-0.090
        & \cellcolor{red!10}-0.089 & \cellcolor{red!6}-0.053
        & \cellcolor{red!6}-0.050 & \cellcolor{red!3}-0.030
        & \cellcolor{red!27}-0.230 & \cellcolor{red!1}-0.013
        & \cellcolor{red!6}-0.051 & \cellcolor{green!1}0.004
        & \cellcolor{red!13}-0.107 & \cellcolor{red!4}-0.036 \\

        WD\_1
        & \cellcolor{green!7}0.056 & \cellcolor{green!4}0.035
        & \cellcolor{green!6}0.052 & \cellcolor{green!1}0.004
        & \cellcolor{green!7}0.061 & \cellcolor{green!1}0.010
        & \cellcolor{red!2}-0.017 & \cellcolor{red!1}$-6.4\cdot 10^{-4}$
        & \cellcolor{green!1}0.008 & \cellcolor{red!1}-0.005
        & \cellcolor{green!4}0.032 & \cellcolor{green!1}0.009 \\

        WD\_6
        & \cellcolor{green!8}0.066 & \cellcolor{green!2}0.019
        & \cellcolor{red!9}-0.073 & \cellcolor{red!1}$-1.5\cdot 10^{-4}$
        & \cellcolor{red!1}-0.010 & \cellcolor{green!1}0.006
        & \cellcolor{red!11}-0.093 & \cellcolor{red!1}-0.003
        & \cellcolor{green!8}0.066 & \cellcolor{red!1}$-3.6\cdot 10^{-4}$
        & \cellcolor{red!1}-0.009 & \cellcolor{green!1}0.004 \\

        WD\_7
        & \cellcolor{green!8}0.067 & \cellcolor{red!5}-0.042
        & \cellcolor{red!6}-0.055 & \cellcolor{red!1}-0.003
        & \cellcolor{red!15}-0.126 & \cellcolor{green!1}0.005
        & \cellcolor{red!22}-0.193 & \cellcolor{red!1}$-9.5\cdot 10^{-5}$
        & \cellcolor{green!13}0.107 & \cellcolor{green!1}0.009
        & \cellcolor{red!5}-0.040 & \cellcolor{red!1}-0.006 \\

        Wind\_DA\_lag\_0
        & \cellcolor{red!43}-0.373 & \cellcolor{red!22}-0.185
        & \cellcolor{red!56}-0.482 & \cellcolor{red!19}-0.164
        & \cellcolor{red!60}-0.513 & \cellcolor{red!7}-0.058
        & \cellcolor{red!70}-0.601 & \cellcolor{red!4}-0.032
        & \cellcolor{red!68}-0.582 & \cellcolor{red!4}-0.030
        & \cellcolor{red!59}-0.511 & \cellcolor{red!11}-0.094 \\

        price\_last
        & \cellcolor{green!33}0.287 & \cellcolor{green!18}0.158
        & \cellcolor{green!21}0.181 & \cellcolor{green!8}0.067
        & \cellcolor{green!36}0.308 & \cellcolor{green!4}0.034
        & \cellcolor{green!19}0.166 & \cellcolor{green!3}0.023
        & \cellcolor{green!11}0.098 & \cellcolor{green!2}0.018
        & \cellcolor{green!24}0.208 & \cellcolor{green!7}0.060 \\

        \bottomrule
      \end{tabular}
    \end{adjustbox}

    \caption{\added{Estimated LEM coefficients obtained from an initial OLS warm start (Pre OLS) and after joint fine-tuning with the RNN/Kalman filter via gradient descent (Post GD). Results are reported for selected rolling windows, including the start and end of the test sample, midpoints of each test year, and the average over all selected windows. Cell shading indicates coefficient sign and magnitude: \textcolor{green}{green} (\textcolor{red}{red}) denotes \textcolor{green}{positive} (\textcolor{red}{negative}) values, with intensity proportional to the absolute coefficient value (common scale across all cells).}}
    \label{tab:merged_windows_dailyavg_base_updated}
  \end{table}

\begin{figure}[h!]
    \centering
    \includegraphics[width=0.9\textwidth]{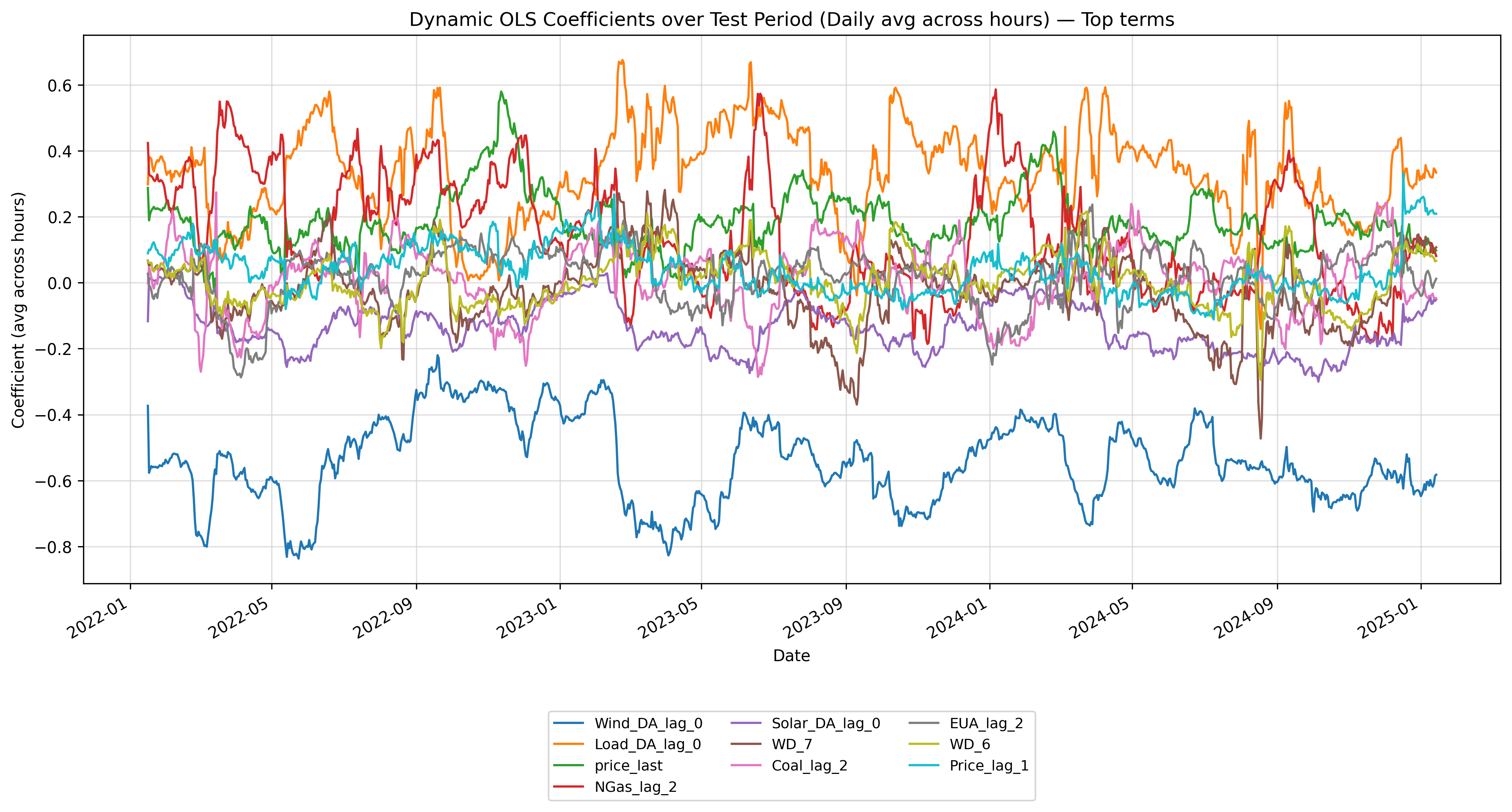}
    \caption{\added{Time-varying OLS coefficients of the LEM component over the test period, averaged daily across hours.}}
    \label{fig:ols_weights_dynamic}
  \end{figure}

  \added{We further explore the temporal stability of feature importance by examining the evolution of LEM coefficients over the entire test period. Figure \ref{fig:ols_weights_dynamic} illustrates the time-varying OLS coefficients of the LEM coefficients over the test period, averaged daily across hours. We noticed that the load coefficient is positive over time and has a higher amplitude than the other regressors. This indicates that high demand leads to higher prices across the entire period. Furthermore, demand seasonality is reflected in its magnitude and temporal variation (high in winter, low in summer). Regarding renewable generation, we observed that the wind generation coefficient is consistently negative and has the largest magnitude among the renewable variables. The coefficient of solar generation is also negative, but of smaller magnitude than that of wind, due to its concentration during daylight hours. While the natural gas coefficient remains positive, its magnitude varies as its marginal contribution to electricity price formation changes.
    The coefficients for coal and EUA prices are smaller and variable over time. This indicates that although these costs affect electricity prices, their impact is primarily indirect and depends on their interactions with other fuels in the generation mix.
  Calendar effect coefficients tend to be time-varying but remain secondary, suggesting that they act more as regime shifters than primary price drivers.}

  \added{To understand the relationships among the LEM coefficients, we illustrate the correlation structure of the dynamic OLS coefficients in Figure \ref{fig:ols_weights_correlation} (in the Appendix). The heatmap reveals the co-movement of the estimated effects of different regressors over the test period.\\
    Several interpretable patterns emerge. The coefficients for Saturday and Sunday are strongly correlated, indicating a coherent weekend pricing regime. Their high correlation with the load coefficient suggests that changes in demand sensitivity are closely linked to calendar effects, reflecting systematic differences in consumption and bidding behaviour between weekdays and weekends.
    The fuel-related coefficients show notable negative correlations, especially between natural gas and coal, and between natural gas and EUA. This pattern reflects substitution effects within the generation technolgies: when gas plays a stronger role in marginal price formation, the contribution of coal or carbon costs becomes less pronounced, and vice versa.
    Moderate positive correlations between renewable coefficients (wind and solar) and price persistence terms indicate that periods with stronger renewable price effects often coincide with more persistent price dynamics. Weaker persistence is associated with more volatile regimes.
  Overall, the correlation matrix shows that coefficient movements are structured and economically meaningful. They reflect underlying interactions between demand, fuel substitution, renewables, and calendar effects, rather than random instability.}

\begin{figure}[h!]
    \centering
    \includegraphics[width=0.9\textwidth]{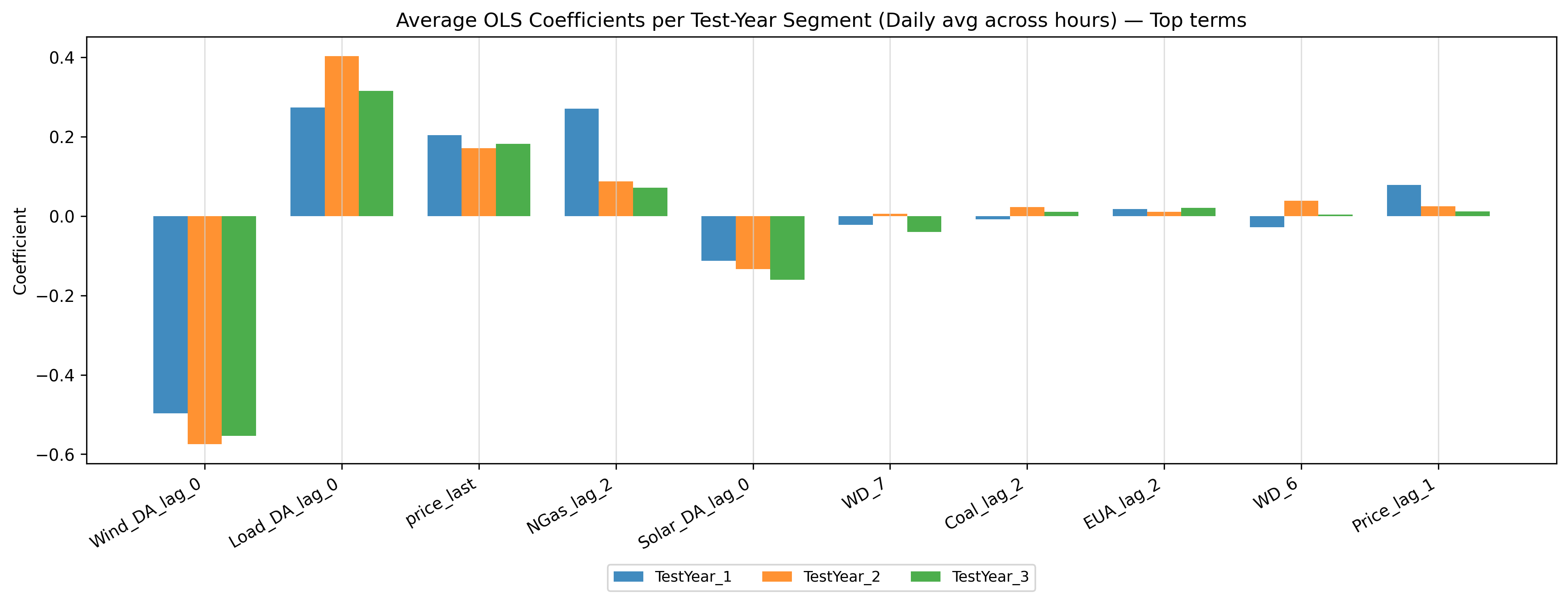}
    \caption{\added{Feature importance from the LEM component of the LEM-KF-RNN model for separate segments of the test period in \textcolor{blue}{2022}, \textcolor{orange}{2023}, and \textcolor{darkgreen}{2024}.}}
    \label{fig:ols_weights_segment_means}
  \end{figure}

  \added{To understand how feature importance varies across different segments of the test period, we quantify the LEM coefficients over the three years of the test sample separately. Figure \ref{fig:ols_weights_segment_means} presents average OLS coefficients calculated for each test year: 2022 (energy crisis period), 2023, and 2024 (post-crisis period), allowing direct economic comparison across different market regimes.\\
    The most notable finding relates to natural gas. In Test Year 1 (2022), at the height of the European energy crisis, the gas coefficient is significantly larger than in later years. This highlights the dominant influence of natural gas generation on marginal electricity prices during this period, driven by supply disruptions, geopolitical uncertainty, and extreme fuel price volatility. As a result, electricity prices in 2022 were highly sensitive to gas price movements, consistent with observed market behaviour. Figure \ref{fig:data} further confirmed this finding, showing that price peaks for both gas and electricity align in 2022. Conversely, the gas coefficient is substantially lower in test years 2023 and 2024. Consequently, electricity prices are less dependent on gas costs because markets have stabilised, alternative generation sources have re-emerged in the merit order, and policy interventions have mitigated extreme price movements.\\
  Other effects remain structurally consistent across the years. Over the three test years, wind generation consistently showed the highest magnitude and the negative impact. Despite its varying magnitude, load remains the dominant positive driver. Compared with 2022, both wind and load effects increased in the second test year. Additionally, solar generation has a negative impact, which is more pronounced in the final year of the test period. Indeed, the increase in wind and solar contributions in the later test years is consistent with increased renewable penetration, which likely comes to reduce the role of gas in electricity price formation relative to 2022. This figure directly demonstrates that the model captures regime-specific price-formation mechanisms.}

  \subsection{Hyperparameter optimization results}
  \label{hyperparam_results_section}
  \added{In this section, we present and interpret the hyperparameter optimisation results obtained from the validation sample. Thereafter, we perform a stability analysis to assess the robustness of the selected hyperparameters across different market regimes.}
  \subsubsection{Hyperparameters interpretation}

  Table \ref{tab:hyperparams_all_models} in the Appendix summarises hyperparameter optimisation results from the validation sample for all model architectures. This set of optimal hyperparameters will serve as initial values for forecasting the test sample. \deleted{Despite the pure RNN achieving the lowest validation RMSE ($28.12$), the difference between it and the best hybrid configuration, especially the RNN-LEM-KF model ($28.16$), is marginal ($\approx 0.05$ RMSE).} \added{The hybrid configurations deliver the lowest RMSE across all settings, followed by the nonlinear RNN, whereas the linear model alone is insufficient to capture the underlying price dynamics and therefore exhibits higher errors.}
  The low RMSE value associated with the RNN model can be the result of finely tuned hyperparameters. More precisely, the optimal initial learning rate is \deleted{small} \added{moderate} (\deleted{$\eta_{\mathrm{init}} = 2.48\times10^{-5}$} \added{$\eta_{\mathrm{init}} = 5.51\times10^{-4}$), allowing efficient early optimization}. The initial training window is large \deleted{($D_{\mathrm{init}} = 359$)} \added{ ($D_{\mathrm{init}} = 540$)}, and mild regularisation (\deleted{$\lambda_{w,\mathrm{init}} = 5.31\times10^{-4}$} \added{$\lambda_{w,\mathrm{init}} = 4.77\times10^{-3}$}) that preserves nonlinear temporal dependencies while stabilising recurrent gradients. Moreover, the RNN model benefits from a larger number of hidden units ($H=110$) enabling it to capture complex nonlinear relationships.\\
  \added{The difference between the $H_{KF}$ and $H_{RNN}$ is due to the fact that the linear/KF branch is adopted to capture stable linear effects, while the nonlinear RNN branches aim to handle more complex nonlinear temporal relationships. Hence, a larger KF hidden-layer dimension is unnecessary as it risks interfering with the nonlinear representations learned by the RNN components.}\\
  The linear LEM and KF models depend on explicit feature relationships and limited memory, as evidenced by their learning rates (\deleted{$\eta_{\mathrm{init}} = 1.71\times10^{-5}$ and $1.23\times10^{-3}$} \added{$\eta_{\mathrm{init}} = 1.85\times10^{-5}$ and $7.56\times10^{-4}$}, respectively), abbreviated \deleted{update intervals  (e.g., $D_{\mathrm{all}} = 33$ for KF model)}, and more stringent penalties (e.g., \deleted{$\lambda_{w,\mathrm{init}} = 3.09\times10^{-3}$} \added{$\lambda_{w,\mathrm{init}} = 1.28\times10^{-3}$} for KF model).\\
  The optimal sequence length of all models is set to 1 ($L=1$), signifying that extended sequences were unnecessary. \deleted{Furthermore, across models with an LEM component, the consistent selection of the OLS-based initialisation indicates that this is crucial to enhance interpretability and convergence stability. However, the relative ranking of models during the test phase may not necessarily hold even though the RNN has the lowest numerical validation RMSE. } \added{Moreover, the updated training window ($D_{\mathrm{all}}$) is consistently short across all models, indicating that frequent model updates using recent data are beneficial for capturing evolving market dynamics.}\\
  \added{The hyperparameter regimes of hybrid models are intermediate and more balanced. The LEM–RNN configuration reduces RNN capacity relative to the pure RNN ($H=76$), limiting model complexity and mitigating overfitting while preserving nonlinearity. In addition, it adopts a shorter training window ($D_{\mathrm{init}} = 114$), indicating that it can adapt to recent dynamics more quickly than relying on long historical data. Further, lower initial learning rates ($\eta_{\mathrm{init}} = 6.72\times10^{-5}$ ) promote stabilisation early in the optimisation process.\\
  Additionally, the fully integrated RNN–LEM–KF model achieves a compact, well-regularised configuration, characterised by a smaller hidden size ($H=53$) and a shorter initial training window  ($D_{\mathrm{init}} = 263$) than the standalone RNN and KF models. As a result, joint optimisation across all components promotes parameter efficiency over maximum capacity.}

  \begin{figure}[h!]
    \centering
    \includegraphics[width=0.9\textwidth]{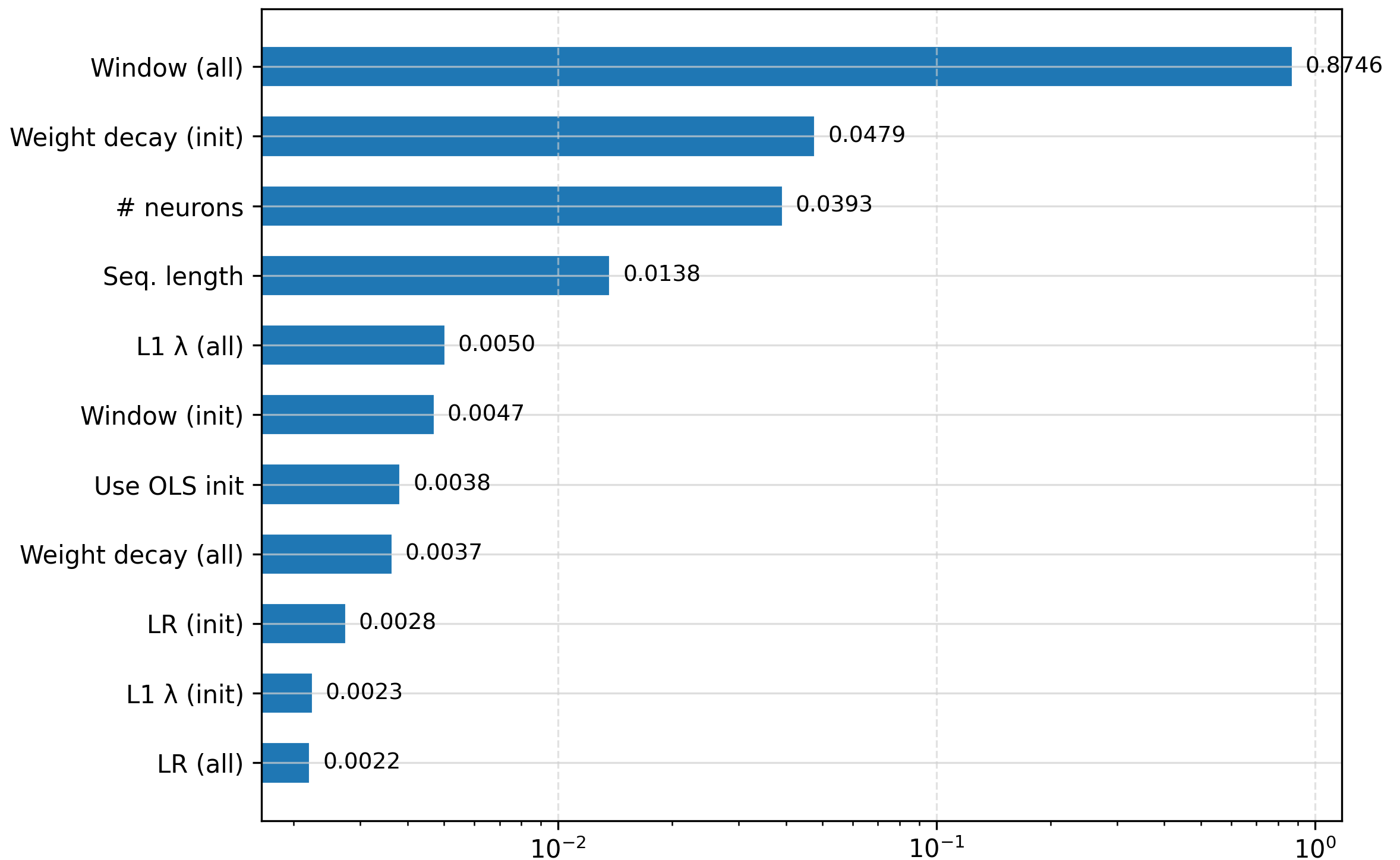}
    \caption{\added{Important variables identified during hyperparameter optimization. Importance score in X-axis are in log scale for better visualisation.}}
    \label{fig:important_variables}
  \end{figure}
  To assess the importance of hyperparameters in the learning process, we performed a fANOVA analysis (\cite{hutter2014efficient}). Results in Figure \ref{fig:important_variables} indicate that the length of the rolling window is the most powerful hyperparameter (\deleted{$\approx 0.74$} \added{$\approx 0.87$}), highlighting the importance of historical data in enhancing model accuracy. Followed by \deleted{learning rate} \added{weight decay}, which, with lower significance (\deleted{$\approx 0.20$} \added{$\approx 0.05$}), is essential for \deleted{achieving steady convergence} \added{regularization and stable convergence}. Other hyperparameters showed similar minor effects. This indicates that historical data and learning rate play a greater role in forecasting accuracy than other architectural or regularisation adjustments; the former plays a primary role, and the latter plays a secondary role.\\
  \begin{figure}[h!]
    \centering
    \includegraphics[width=1\textwidth]{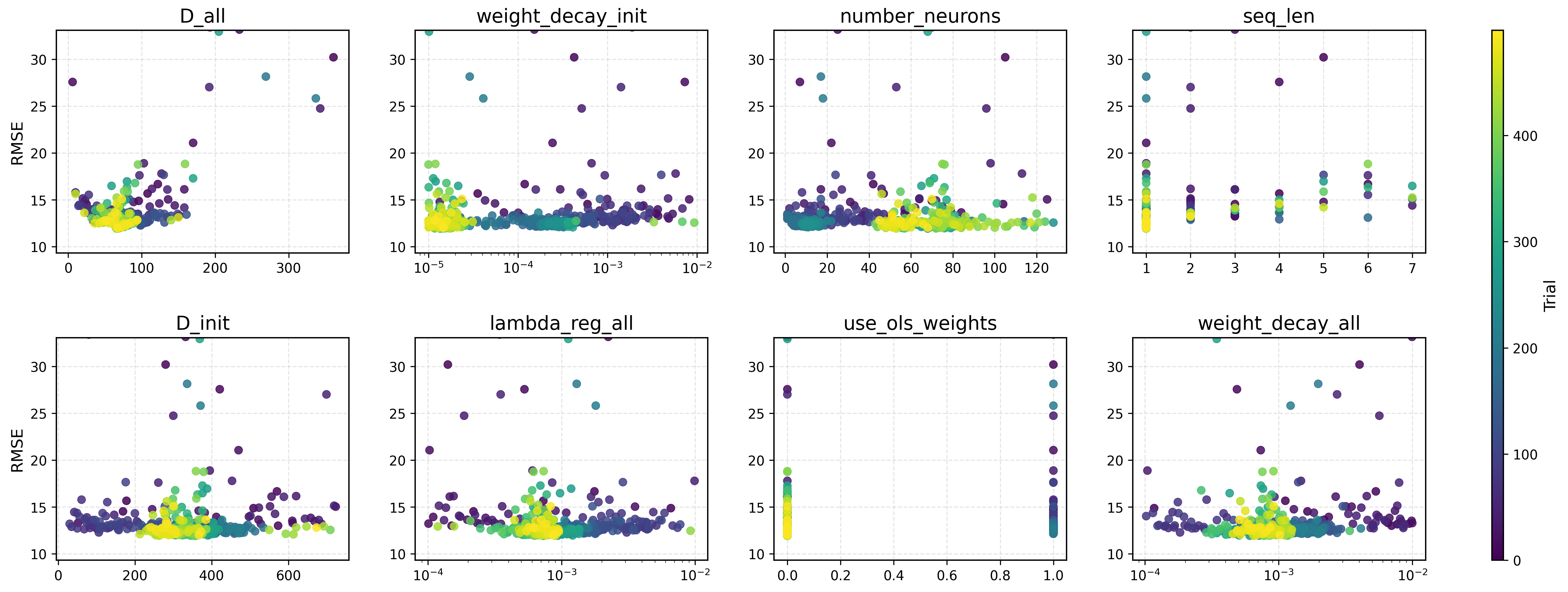}
    \caption{\added{Top hyperparameter values for best-performing trials.}}
    \label{fig:hp_slice_top_params}
  \end{figure}
  Figure \ref{fig:hp_slice_top_params} illustrates the impact of each hyperparameter (initialisation and rolling-update values) on the model's performance (RMSE) during 500 iterations of the training period. \added{The length of the training window exhibits distinct optimal ranges for both the initial and updated windows. The initial window (D$_{init}$) performs better for larger windows. This means that a longer historical context is needed to properly set the model parameters. However, the updated window (D$_{all}$) achieves lower RMSE for shorter training window. This suggests that updates rely more on recent information and should not include old observations.}. \deleted{For the learning rate, both the initial ($\eta_{init}$) and updated values ($\eta_{all}$) perform better in the interval [$10^{-4}, 10^{-3}$], meaning in mid-range values. However, the RMSE increased when learning speeds were extremely high or low, which indicates that the learning rate ought to be balanced}. The weight decay,($\lambda_{w, init}$ and $\lambda_{w, all}$) \deleted{a mild regularisation of about $10^{-4}$ and $10^{-3}$ range tends to reduce overfitting. Thus, a moderate L2 regularisation value is beneficial during initialisation and rolling adaptation} \added{do not indicate a sharp optimum. Across the tested range, RMSE values are broadly distributed, indicating that performance is relatively insensitive to this hyperparameter. This is due to the fact that the weight decay has a secondary role relative to other hyperparameters, serving primarily as a stabilising regulariser rather than as a principal performance factor}. \added{Similarly, the L1 regularization parameters ($\lambda_{1, all}$) shows no sharply defined optimum.} \deleted{Concerning the L1 regularization parameters (\deleted{$\lambda_{1, init}$} and $\lambda_{1, all}$), a moderate value of L1 regularization assists the model in focusing on salient features rather than irrelevant ones.} In terms of sequence length, models with short input sequences of one to three time steps or days yielded the best performance (lower RMSE), indicating that long historical sequences do not provide additional useful patterns. \deleted{Finally, the rolling window size ($D_{all}$) significantly improved accuracy over longer training periods, especially between 300 and 365 days, implying that the model is capable of understanding more complex patterns over longer periods.}\\

  \deleted{These findings are reinforced by the results in Figure \ref{fig:hp_sampling_distributions} in the Appendix, which illustrates how each hyperparameter was explored during training and where the search trials were most concentrated. It indicates that the updated training window $D_{all}$ is clearly left-skewed (most trials are packed near the high end $\sim 330\text{--}365$ with a tail towards small windows). Both L1 regularisation parameters peaked near 0 ($\approx 10^{-3}\text{--}10^{-4}$) with a tail for large values. This is also similar for the learning rate and the weight decay. A left-skewed distribution of neurons, with a strong tendency towards larger sizes ($100\text{--}128$), suggests that a model with larger hidden layers is able to capture more complex patterns and nonlinear relationships in the data, thus improving forecasting. The sequence length is strongly right-skewed (over 400 trials chose 1). $\alpha$ is left-skewed (mass near $1.5\text{--}2.0$), indicating that stronger OLS scaling was favoured. In addition, use of OLS weights are dominated by True.\\}

  \begin{figure}[H]
    \centering
    \includegraphics[width=0.9\textwidth]{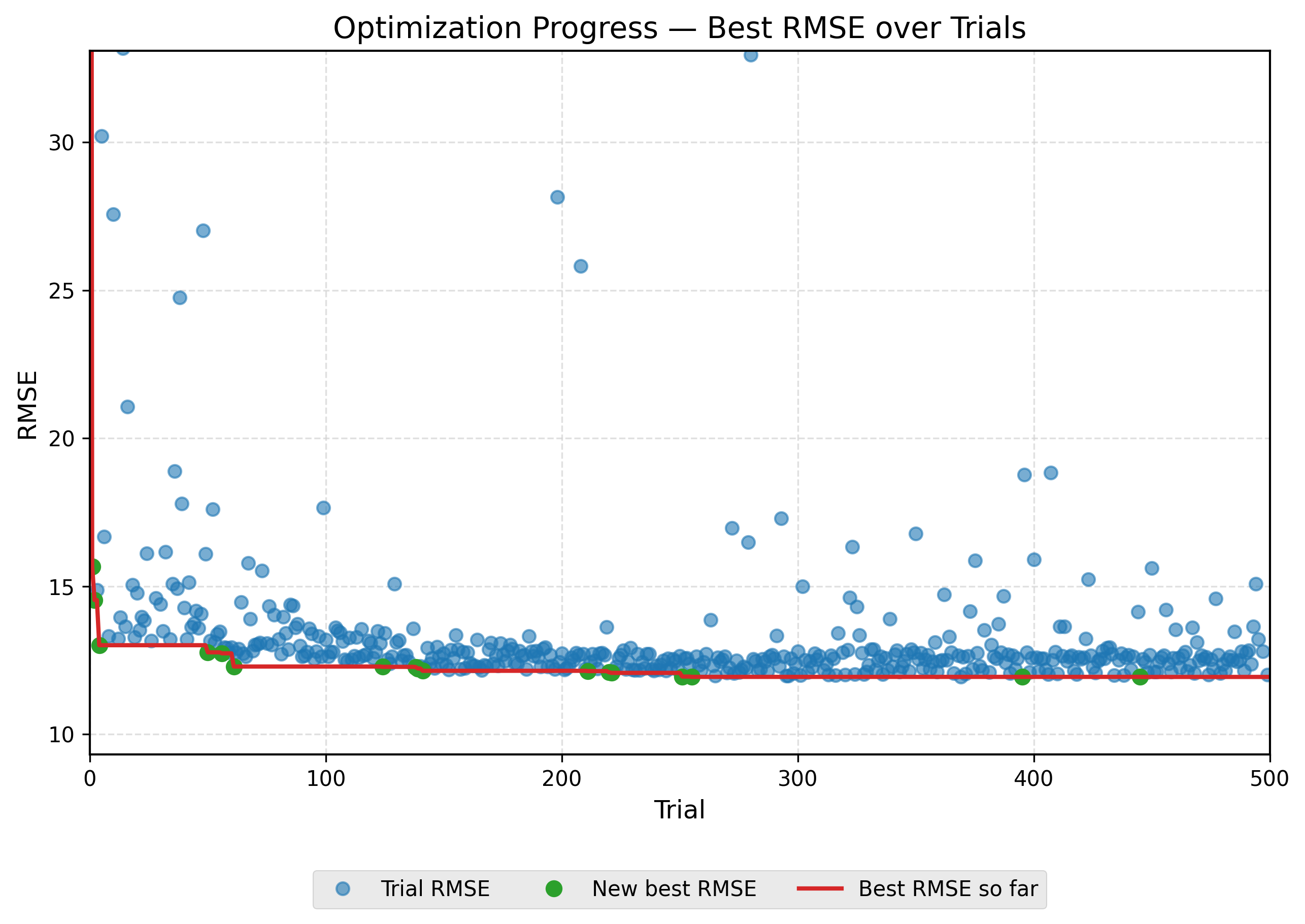}
    \caption{\added{Hyperparameter optimization progress across 500 trials, where \textcolor{blue}{the blue dots represent the RMSE values achieved for each trial}, \textcolor{darkgreen}{the new best RMSE is indicated by the green dots}, \textcolor{red!80!black}{the red line links best RMSE values over trials.}}}
    \label{fig:hp_optimization_history}
  \end{figure}
  Figure \ref{fig:hp_optimization_history} illustrated that the lowest RMSE was achieved early during the optimization process, suggesting that effective hyperparameter values were found rather quickly.\added{ In the following phases of optimization, most trials yield comparable low RMSE values, reflecting a stable region of near-optimal solutions.}  While the true global minimum was found much later, around trial \deleted{400} \added{450}, \deleted{. The randomly scattered results of the trials and the lack of an overall downward trend provide evidence for the stochasticity of the search process, and imply that} further trials (beyond 500) could provide marginal returns.

  \subsubsection{Hyperparameter stability analysis}
  \label{hyperparam_stability_section}
  \added{To assess the stability of hyperparameter optimisation results, we conducted a stability analysis of the best hyperparameters across different validation periods within different regimes. Two distinct validation periods were defined: (i) Validation Period 1 (13 Jan 2021 – 12 Jan 2023), which encompasses both pre-crisis and crisis periods, and represents the whole validation period (2 years) used as a benchmark to understand how the hyperparameters drifted if it's optimised in high volatility periods, which leads us to the (ii) Validation Period 2 (13 Jan 2022 – 12 Jan 2023) that includes the peak of the energy crisis. The test period is the same for both validation periods (13 Jan 2023 – 10 Jan 2025), i.e., the last 2 years of the data.}
  \begin{table}[H]
    \centering
    \small
    \caption{Best hyperparameters and performance of the LEM-KF-RNN model under different validation periods.}
    \label{tab:rnn_lem_kf_stability}

    \resizebox{\textwidth}{!}{%
      \begin{tabular}{lcc}
        \toprule
        \textbf{Symbol} &
        \textbf{LEM-KF-RNN} &
        \textbf{LEM-KF-RNN} \\
        & \textbf{(Validation Period 1)} & \textbf{(Validation Period 2)} \\
        \midrule
        $H$                     & 118 & 117 \\
        $L$                     & 1   & 1 \\
        $D_{\mathrm{init}}$      & 302 & 351 \\
        $D_{\mathrm{all}}$       & 358 & 359 \\
        $\eta_{\mathrm{init}}$   & $8.05\times10^{-4}$ & $2.04\times10^{-3}$ \\
        $\eta_{\mathrm{all}}$    & $1.09\times10^{-4}$ & $7.24\times10^{-4}$ \\
        $\lambda_{w,\mathrm{init}}$ & $9.48\times10^{-4}$ & $7.76\times10^{-4}$ \\
        $\lambda_{w,\mathrm{all}}$  & $1.67\times10^{-3}$ & $1.30\times10^{-4}$ \\
        $\lambda_{1,\mathrm{init}}$ & $1.41\times10^{-4}$ & $1.79\times10^{-4}$ \\
        $\lambda_{1,\mathrm{all}}$  & $1.79\times10^{-4}$ & $9.58\times10^{-4}$ \\
        \texttt{use\_ols\_weights} & True & False \\
        \midrule
        \textbf{Validation RMSE} & \textbf{28.162} & \textbf{34.844} \\
        \midrule
        \textbf{Test RMSE} & \textbf{22.754} & \textbf{25.046} \\
        \textbf{Test MAE}  & \textbf{13.006} & \textbf{13.816} \\
        \bottomrule
      \end{tabular}%
    }
  \end{table}

  \begin{figure}[h!]
    \centering
    \begin{subfigure}[t]{0.48\textwidth}
      \centering
      \includegraphics[width=\textwidth]{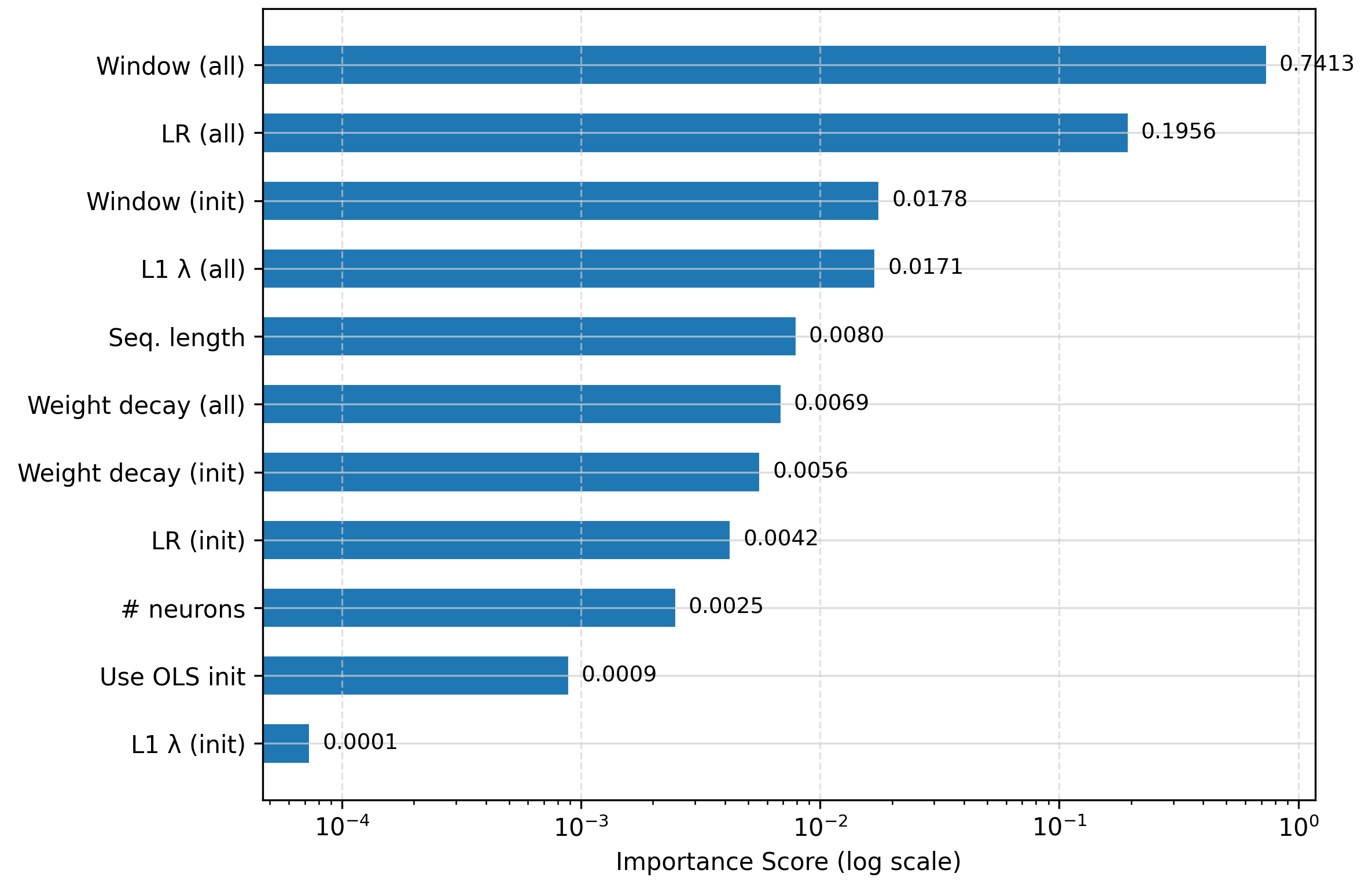}
      \caption{Validation Period 1}
      \label{fig:important_variables_val1}
    \end{subfigure}
    \hfill
    \begin{subfigure}[t]{0.48\textwidth}
      \centering
      \includegraphics[width=\textwidth]{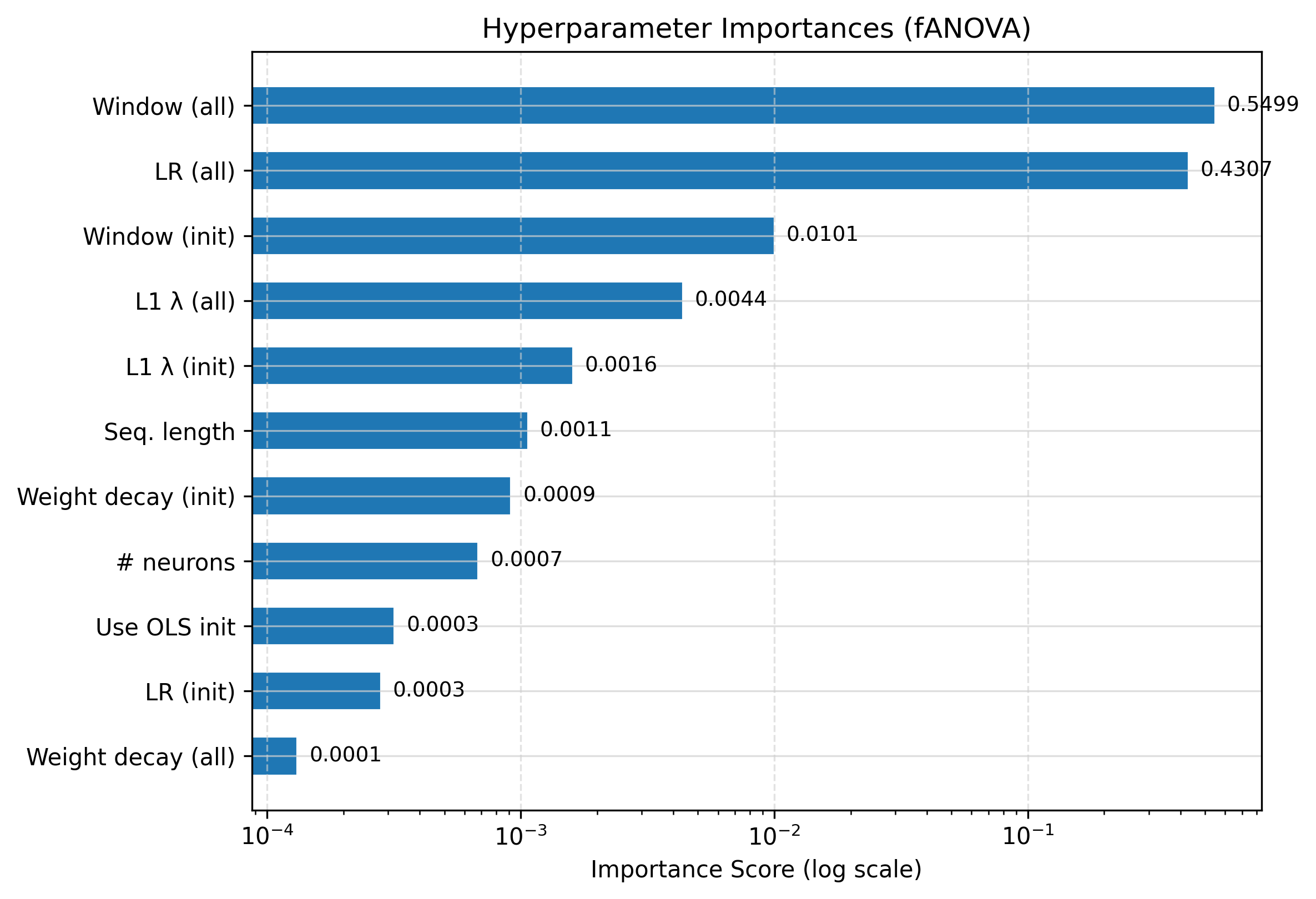}
      \caption{Validation Period 2}
      \label{fig:important_variables_val2}
    \end{subfigure}

    \caption{\added{Important variables identified during hyperparameter optimization for the LEM-KF-RNN model for validation period 1 (subfigure a) and validation period 2 (subfigure b). Importance scores on the x-axis are shown on a logarithmic scale for improved visualisation.}}
    \label{fig:important_variables_stability}
  \end{figure}

  \added{Table \ref{tab:rnn_lem_kf_stability} summarises the best hyperparameters and performance metrics of the LEM-KF-RNN model under the two different validation periods. The results indicate that the optimal hyperparameters are relatively stable across both validation periods, with only minor variations observed. Specifically, the hidden layer size ($H$) and sequence length ($L$) remain stable across both validation samples, suggesting that the model's capacity and input structure (model architecture hyperparameters) are stable under market conditions. In particular, the hidden layer size ($H$) and the sequence length ($L$) are fixed in both validation schemes, implying that the model's capacity and input configuration are invariant across different market regimes. However, we noticed an increase in learning rates ($\eta_{init}$ and $\eta_{all}$) and the regularisation parameters ($\lambda_{1, all}$) in the second validation sample, indicating that an adjustment of hyperparameters related to the model optimisation is required to accommodate changes in volatility profiles across validation periods.}\\
  \added{Regarding the evaluation metrics (RMSE and MAE) on both the validation sets and the test set, we observed better performance when validated on the longer period that includes pre-crisis data than on the period limited to the crisis. This suggests that incorporating a broader range of market conditions during validation can enhance model generalizability.}\\
  \added{ Figure \ref{fig:important_variables_stability} illustrates the importance of hyperparameters, as determined by fANOVA analysis, during hyperparameter optimization for the LEM-KF-RNN model across the two validation periods.  Overall, the most important hyperparameters, update window size and learning rate ($D_{all}$ and $\eta_{all}$), exhibit high importance scores. The sum of their scores is 93.66\% for the first validation period and 98.06\% for the second validation period, due to the increased updated learning rate ($\eta_{all}$).  However, the L1 regularisation ($\lambda_{1, all}$) appears to be more important in the first validation sample. This may be due to the increased importance of the learning rate in the second validation period, which reduces the importance of the other hyperparameter. Hence, the most sensitive hyperparameters to the highly volatile market regime are the learning rate, since it governs how rapidly the model adapts to sudden structural changes while maintaining numerical stability during rolling retraining.}

  \subsection{Forecasting results}
  An interpretation and discussion of the evaluation metrics, the GW statistical test, and the forecast decomposition analysis are provided in this section.
  \subsubsection{Forecasting accuracy metrics}

  A summary of forecasting accuracy metrics for all models over the \deleted{two-year} \added{three-years} testing period \added{for Germany–Luxembourg market} is illustrated in Table \ref{tab:forecasting_metrics_Germany}.

  \begin{table}[h!]
    \centering
    \renewcommand{\arraystretch}{1.2}
    \resizebox{\textwidth}{!}{%
      \begin{tabular}{lcccccc}
        \toprule
        \textbf{Model} & \textbf{MAE} & \textbf{RMSE} & \textbf{rMAE} &
        \textbf{rRMSE$_{\text{naive}}$} & \textbf{rRMSE$_{\text{mean}}$} & \textbf{DA} \\
        \midrule
        Naive          & 50.551 & 78.083 & --    & --    & 0.577 & 0.640 \\
        RNN            & 18.549 & 29.811 & 0.366 & 0.381 & 0.220 & 0.835 \\
        KF             & 19.020 & 30.371 & 0.376 & 0.389 & 0.224 & 0.840 \\
        LEM            & 19.201 & 30.819 & 0.380 & 0.395 & 0.228 & 0.835 \\
        LEM-RNN        & 18.362 & 29.490 & 0.362 & 0.377 & 0.218 & 0.842 \\
        KF-RNN         & 18.094 & 29.136 & 0.356 & 0.372 & 0.215 & 0.845 \\
        LEM-KF-RNN     & \textbf{17.625} & \textbf{28.429} & \textbf{0.349} & \textbf{0.364} & \textbf{0.210} & \textbf{0.848} \\
        Ens--LEM-RNN   & 17.632 & 28.583 & 0.349 & 0.366 & 0.211 & 0.848 \\
        Ens--KF-RNN   & 17.979 & 29.014 & 0.356 & 0.372 & 0.214 & 0.844 \\
        DNN     & 19.697 & 31.861 & 0.390 & 0.409 & 0.235 & 0.817 \\
        DNN/-Fuel           & 20.318 & 32.202 & 0.402 & 0.413 & 0.238 & 0.825 \\
        LEAR     & 23.045 & 35.892 & 0.457 & 0.460 & 0.265 & 0.827 \\
        LEAR/-Fuel           & 24.227 & 38.320 & 0.480 & 0.491 & 0.283 & 0.822 \\
        \bottomrule
    \end{tabular}}
    \caption{\added{Forecasting performance of all models tested for the Germany-Luxembourg market. Lower values indicate better performance, while higher directional accuracy (DA) is preferred.}}
    \label{tab:forecasting_metrics_Germany}
  \end{table}

  For the single-type models, we noticed that the RNN model outperformed the stand-alone linear models, with an RMSE \deleted{of 23.078} \added{of 28.429}. This indicates its strength in capturing the nonlinear dynamics and time dependences that characterise electricity price time series. However, purely linear models, i.e., LEM and KF, generate higher forecasting errors \deleted{(RMSE = 24.465 and RMSE = 25.352, respectively)} \added{(RMSE = 30.819 and RMSE = 30.371, respectively)}, with the KF model slightly outperforming the LEM model. Despite their ability to capture structural trends and mean-reverting patterns, linear methods often struggle with nonlinearity, spikes, and volatility. 

  For the combined models, we found that the hybrid models significantly improved the forecasting accuracy of the linear components (LEM and KF) by incorporating a nonlinear component (RNN). Specifically, KF-RNN achieved lower errors \deleted{(RMSE = 23.503)} \added{(RMSE = 29.136)} compared to LEM-RNN \deleted{(RMSE = 24.940)} \added{(RMSE = 29.490)}reflecting the additional benefit of modeling linear temporal dynamics through state-space representations. Furthermore, the LEM-KF-RNN model achieved the best forecasting results \deleted{(RMSE = 22.754)} \added{across all metrics (achieving the lowest MAE (17.625), RMSE (28.429), rMAE (0.349), and rRMSE$_{\text{naive}}$ (0.364))}.
  \added{In addition, the LEM-KF-RNN model has the lowest rRMSE$_{\text{mean}}$ (rRMSE$_{\text{mean}}$ =0.210), indicating that its typical forecasting error is approximately 21\% of the average electricity price during the test period. Their improvements are therefore significant, even relative to average prices.\\
  Additionally, the LEM-KF-RNN model consistently achieved the highest DA value (DA=0.848), indicating superior predictive ability to capture the direction of prices in 84.8\% of cases. In this regard, the LEM-KF-RNN model provides more accurate magnitude and direction forecasts, thereby reinforcing its relevance for decision-making applications.}\\
  This strong performance comes from the model's ability to capture both smooth structural components through LEM and KF, and rapid, nonlinear changes through RNN. Combining these elements allows the model to respond well to different market conditions and keep a good balance between bias and variance. 

\added{On one side, the linear-RNN ensemble combinations (Ens--LEM-RNN and Ens--KF-RNN)} \footnote{The Ensemble model is an equal-weight average of single forecasts}) \added{provided improvements over their single counterparts, indicating that model averaging can further stabilise predictions and reduce variance. On the other side, the proposed LEM-KF-RNN model demonstrates a slightly superior performance compared with Ens--LEM-RNN and Ens-KF-RNN, suggesting that joint training within a unified architecture is more effective than post-hoc averaging of forecasts. This superiority is further supported by computational costs (reported in Table \ref{tab:computational_time}). More precisely, an ensemble approach requires training and executing each model independently, with a total computational cost at least as high as the sum of the individual models' runtimes. As an example, combining a LEM with an RNN requires about 21 hours of runtime ($6.27 h + 14.99 h$), but training a joint LEM–KF–RNN model takes only 10.3 hours.}

To assess the accuracy of the proposed model compared to other models in the literature, we first consider the naive model (see Equation \ref{naive}). Furthermore, we benchmark our models with the state-of-the-art models proposed by \cite{lago2021forecasting}. This includes:

\begin{enumerate}
  \item A parameter-rich autoregressive model with exogenous variables, known as the Lasso Estimated AutoRegressive (LEAR) model. Estimation of the proposed LEAR model is based on L1-regularisation (LASSO, Least Absolute Shrinkage and Selection Operator). 
  \item A Deep Neural Network (DNN) model, which extends the traditional multilayer perceptron (MLP). This network consists of four layers, uses a multivariate framework (single model with 24 outputs), is estimated using Adam \cite{adam2014method}, and its hyperparameters and input features are optimised using a Bayesian optimisation algorithm called the tree Parzen estimator \cite{bergstra2011algorithms}.
\end{enumerate}
As input features, both models use the same set of information: 
\begin{enumerate}[a.]
  \item Electricity price lagged values (historical day-ahead prices of the prior three days $p_{d-1}$, $p_{d-2}$, $p_{d-3}$, and one week ago $p_{d-7}$). 
  \item A day-ahead load forecast $\mathbf{x}^1_d$ along with the previous day $\mathbf{x}^1_{d-1}$ and one week ago $\mathbf{x}^1_{d-7}$. 
  \item Similarly, the forecasted day-ahead solar and wind generation $\mathbf{x}^2_d$ plus the previous day value $\mathbf{x}^2_{d-1}$ and the value a week ago $\mathbf{x}^2_{d-7}$. 
  \item  The weekday is represented as a dummy variable. 
\end{enumerate}
Detailed information about the two state-of-the-art models can be found in \cite{lago2021forecasting}. The python codes are open access and available in Github\footnote{https://github.com/jeslago/epftoolbox}.\\

\added{To ensure a fair comparison, the LEAR and DNN models were augmented with fuel prices (natural gas, coal, oil and CO$_2$) as additional input features, similar to our proposed models (as they are defined in Equation \ref{detailedlm}). This is important because fuel prices have a direct effects on electricity prices (\cite{ghelasi2024day}), especially during the energy crisis of 2022 (see Figure \ref{fig:data}). We also evaluated the LEAR and DNN models without fuel prices (noted as LEAR/-Fuel and DNN/-Fuel, i.e., the original versions) to assess their performance in the absence of this critical information.\\}

Examining the benchmark models, the Weekly Naive baseline performs as expected, with the highest errors \deleted{(RMSE = 51.617)} \added{(RMSE = 78.083)}, since it does not account for structural or temporal variability beyond simple weekly repetition. We also examined the state-of-the-art LEAR and DNN models. Even though they are sophisticated \added{and used the same set of features}, they still have relatively low accuracy \deleted{(RMSE = 26.587 and 25.976, respectively)} \added{(RMSE = 35.892 and 31.861, respectively)}.
\added{We also noticed that when fuel prices are excluded from the input features (LEAR/-Fuel and DNN/-Fuel), their performance further degrades (RMSE = 38.320 and 32.202, respectively), highlighting the importance of including relevant exogenous variables in electricity price forecasting models. \\
Despite the fact that fuel variables improve the performance of state-of-the-art models, the proposed LEM–KF–RNN architecture consistently outperforms them. The results indicate that the improvement in accuracy did not simply result from the addition of richer input information, but from the structured integration of linear expert knowledge, state-space dynamics, and nonlinear temporal modelling within a single, jointly optimised framework. The proposed model's superiority can therefore be attributed to its architectural innovation and coherent integration rather than to the added complexity or features.\\}
In summary, These findings highlight the benefits of hybrid models combining linear, filtering, and nonlinear components for accurate forecasting. This can be achieved through puzzling out various features in electricity prices through each model's strengths.

\added{To assess the robustness of our proposed models across different market regimes, we split the test sample into three-year intervals and computed evaluation metrics for each year. Table \ref{tab:forecasting_by_year} (in the Appendix)provides a segmented analysis of forecast performance across three test years. A stressed regime is evident in 2022 due to the energy crisis, characterised by extreme volatility and abrupt price changes. Meanwhile, 2023 and 2024 represent post-crisis periods of increased stability and predictability.\\
  Across all the test years, we observed that MAE and RMSE are substantially higher in 2022 than in 2023 and 2024. Therefore, forecasting for 2022 will be significantly more challenging owing to heightened volatility. In contrast, relative error measures (rMAE, rRMSE$_\text{naive}$, and rRMSE$_\text{mean}$) are lower in 2022. According to relative metrics, forecast errors are normalised either by the naive model's MAE/RMSE or by the price level, both of which are significantly higher in 2022. As a result, although absolute errors increase, they represent a smaller share of the average price. There is a marked decrease in error levels in 2023, reflecting more stable and predictable market conditions. However, they rise again in 2024, indicating that price dynamics are more volatile than in 2023 (see Figure \ref{fig:data}).\\
  Based on evaluations across sub-periods, all models performed worse in 2022, with accuracy improving in subsequent years. More precisely, from 2022 to 2023, MAE and RMSE decreased by almost half for all models.\\
  Across all years, hybrid model specifications outperform single-model approaches. Based on 2022 results, the LEM-KF-RNN model had the lowest errors and the highest directional accuracy, followed by ensemble hybrids, which emphasise the value of model combinations in challenging market conditions. Despite narrowing performance gaps in 2023, LEM-KF-RNN continues to dominate across all metrics. In 2024, the ensemble variants exceeded LEM-KF-RNN by a slight margin.\\
The LEM-KF-RNN consistently demonstrated the highest robustness across all three subperiods, delivering excellent results in both volatile and stable market regimes. Although ensemble models have occasionally matched or slightly outperformed the LEM-KF-RNN model over the past year, the latter has maintained superior stability and accuracy, demonstrating its suitability for operational forecasting under heterogeneous market conditions.}

\subsubsection{The multivariate GW test results}
\label{subsec:GW test}
According to the results of the multivariate GW test using the $L_1$ norm in Equation \ref{eqmultivariateDM}, we have the following loss differential series:

\begin{equation}
  \Delta_d^{\mathrm{A}, \mathrm{B}}=\sum_{h=1}^{24}\left|\varepsilon_{d, h}^{\mathrm{A}}\right|-\sum_{h=1}^{24}\left|\varepsilon_{d, h}^{\mathrm{B}}\right|
  \label{GW}
\end{equation}

To illustrate the range of obtained $p-values$, we use heat maps arranged as chessboards. Given that this test is run at a 5\% significance level, results are interpreted based on $p-values$ as follows:

\begin{itemize}

  \item {\deleted{$p-values$ $<$ 0.05 (green) indicate significant outperformance of a model on the X-axis (better) compared to the model on the Y-axis (worse).}\added{\textbf{Red cells}: The row model (Y-axis) significantly outperforms the column model (X-axis)($p < 0.05$). In the gradient version, deeper red indicates stronger significance (smaller $p$-values)}}
  \item {\deleted{$p-values$ $>$ 0.05 indicate significant underperformance of the model on the X-axis (worse) compared to the model on the Y-axis (better).}\added{\textbf{Blue cells}: The column model (X-axis) significantly outperforms the row model (Y-axis) ($p < 0.05$). In the gradient version, deeper blue indicates stronger significance (smaller $p$-values).}}
  \item \added{\textbf{White cells}: No significant difference between models ($p \geq 0.05$) or not applicable (diagonal).}
\end{itemize}


\begin{figure}[h!]
  \centering
  \includegraphics[width=0.9\textwidth]{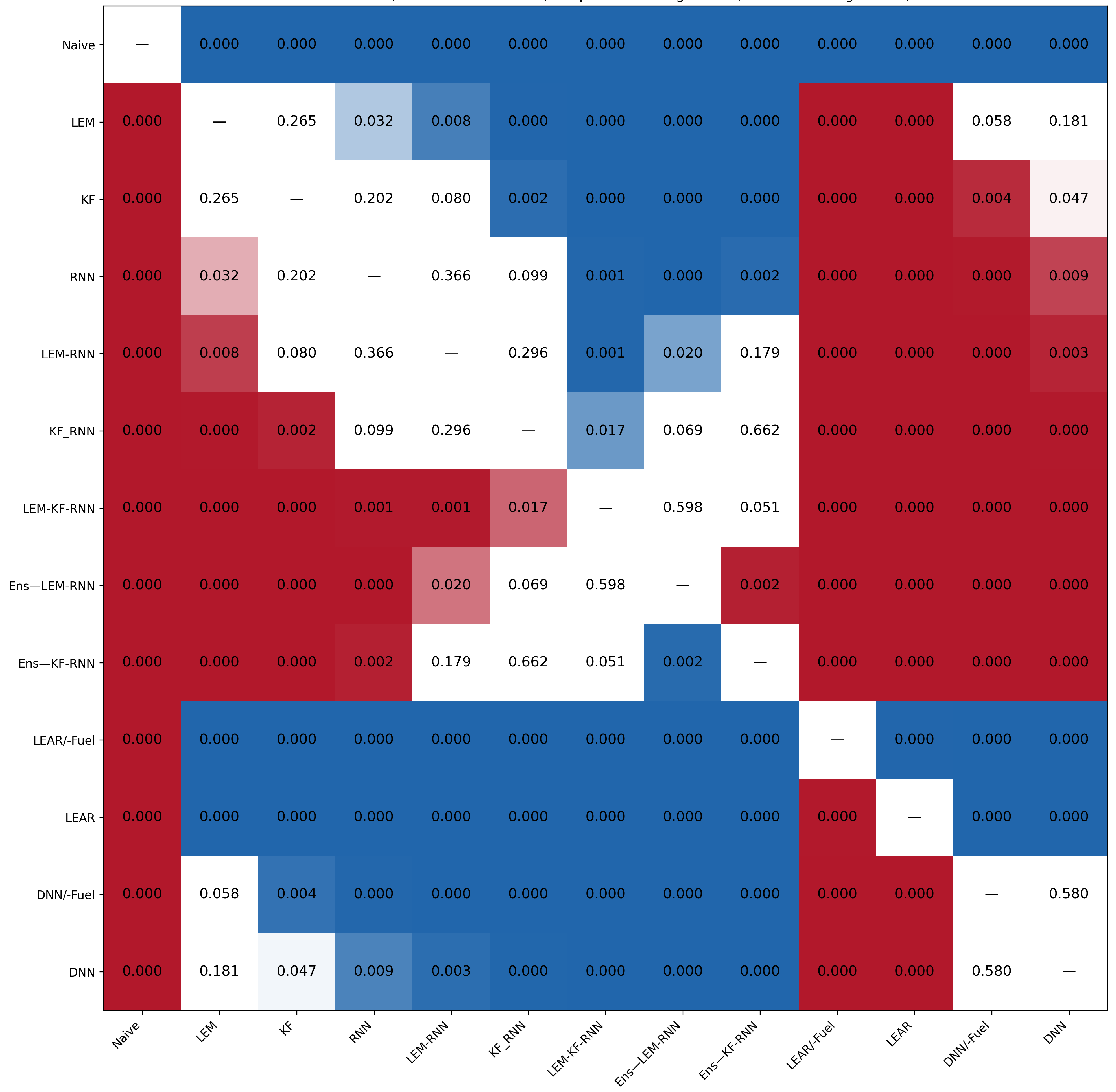}
  \caption{\added{Giacomini–White test results for all models over the two-year test period. For each pair of models, two one-sided tests are run, and a heat map shows the p-values. \textcolor{red}{Red cells} indicate the row model (Y-axis) is significantly better, \textcolor{blue}{blue cells} indicate the column model (X-axis) is significantly better, and white means not significant/NA. In the gradient version, deeper red/blue indicates greater significance (smaller p-values), while pale colors approach the non-significant range.} }
  \label{fig:GW-test-all}
\end{figure}
The GW results are illustrated in Figure \ref{fig:GW-test-all}, the following findings were gathered:
\deleted{As indicated by the green cells in the RNN column, RNN outperforms KF and LEM, which is consistent with its lower RMSE and MAE values. It also significantly outperforms the LEAR and DNN state-of-the-art models.}
\deleted{Both KF and LEM columns are in black, meaning that all other models consistently outperform them, and their forecasting accuracy is not statistically significant, while there is no statistical significance between their forecasting accuracy. Additionally, these models significantly outperform the benchmark models (LEAR and DNN), except for the DNN and KF models, where the latter does not significantly outperform the former.}
\deleted{The combination models (LEM–RNN, KF–RNN, and LEM–KF–RNN) outperformed all single model components and all benchmark models, proving the effectiveness of hybridisation. There is an exception, however, in the RNN and LEM-RNN pair, which does not show a significant difference at 5\%, suggesting that LEM does not substantially improve RNN.}
\deleted{It is worth mentioning that the differences between the hybrid models themselves are mostly statistically insignificant (black cells), meaning that the LEM–KF–RNN, KF–RNN, and LEM–RNN are all comparable in forecasting accuracy when compared to each other.}
\deleted{In summary, the performance of hybrid models is clearly superior to that of single models. The best architectures, KF-RNN and LEM-KF-RNN, show similar statistical results. Consequently, the most effective combinations appear to have reached a plateau in performance.\\}

\begin{itemize}
  \item \added{Compared to the Naive benchmark, all models perform significantly better. More precisely, across all alternatives, the Naive row is uniformly blue, confirming that all proposed and benchmark models deliver statistically significant improvements.}

  \item \added{Among single models, KF and LEM are statistically indistinguishable, while RNN is only clearly better than LEM. According to the heatmap, there is no significant difference between KF and LEM ($p=0.265$). The RNN significantly outperforms LEM ($p=0.032$), but the difference between RNN and KF is not significant ($p=0.202$).}

  \item \added{Adding the RNN branch to a linear component improves the linear model, but does not consistently beat the single RNN model. To illustrate, LEM-RNN significantly improves over LEM ($p=0.008$), and KF-RNN significantly improves over KF ($p=0.002$). However, neither LEM-RNN nor KF-RNN is significantly different from the standalone RNN ($p=0.366$ and $p=0.099$, respectively). This indicates that hybridisation enhances linear baseline accuracy to reach the level of the RNN model.}

  \item \added{LEM-KF-RNN is the most robust hybrid and shows statistically significant gains over most competitors.  To exemplify, the LEM-KF-RNN significantly outperforms every single component (vs LEM: $p=0.000$; vs KF: $p=0.000$; vs RNN: $p=0.001$) and also outperforms LEM-RNN ($p=0.001$). It also yields a statistically significant improvement over KF-RNN ($p=0.017$), supporting the benefit of combining \emph{both} the linear expert and state-space dynamics with the nonlinear branch.}

  \item \added{Top-performing hybrids and ensembles are often statistically tied: The difference between LEM-KF-RNN and Ens—LEM-RNN is not significant ($p=0.598$). The comparison with Ens—KF-RNN is borderline ($p=0.051$), and thus not significant at the 5\% level (but significant at the 10\% level). Additionally, Ens—LEM-RNN significantly outperforms Ens—KF--RNN ($p=0.002$). Overall, this suggests that once strong hybrid structures are used, several approaches become statistically comparable.}

  \item \added{Fuel augmentation improves LEAR substantially, while the effect is not significant for DNN in this setting. Moreover, several proposed models significantly outperform DNN (e.g., KF vs DNN: $p=0.047$; RNN vs DNN: $p=0.009$; and hybrids vs DNN: typically $p\approx 0.000$).}
\end{itemize}

\added{Overall, the GW test confirms that the proposed LEM-KF-RNN model has a statistically significant performance advantage over single linear components. Comparing LEM-KF-RNN to the ensemble variants, the heatmap indicates that they are often statistically tied.}

\begin{figure}[h!]
  \centering
  \includegraphics[width=\textwidth]{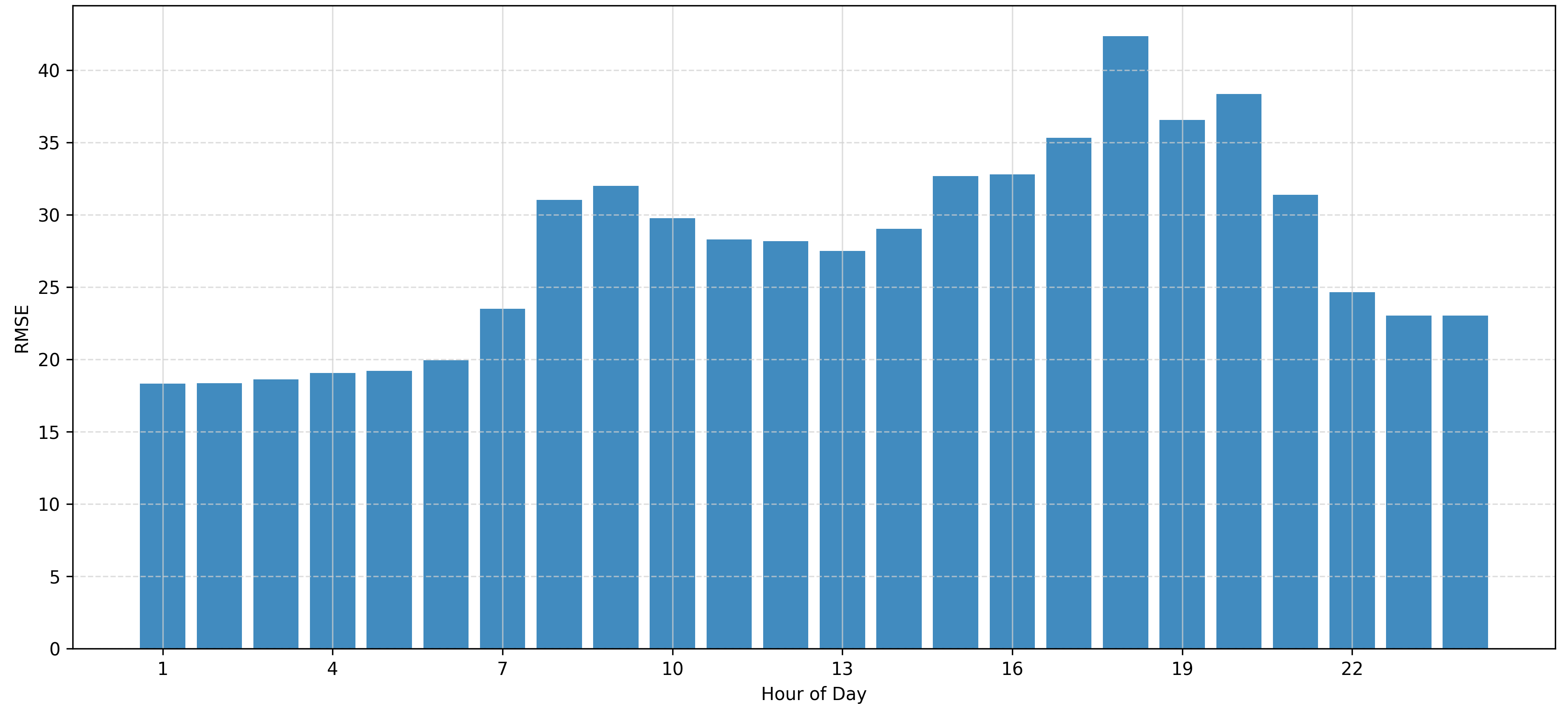}
  \caption{\added{RMSE by Hour (Hourly Test Set) for the LEM-KF-RNN model of the Germany–Luxembourg market.}}
  \label{fig:rmse_by_hour}
\end{figure}

Figure \ref{fig:rmse_by_hour} visualises the hourly RMSE during the test period. It indicates that the LEM-KF-RNN model replicates the daily price cycle accurately, demonstrating a precise understanding of both magnitude and phase as it shows lower prices during the morning, an increase at noon, and a significant peak in the evening. More precisely, the figure shows a varying RMSE value throughout the day, peaking in the late afternoon at about \deleted{35} \added{45}, when peak demand occurs. During the night hours, however, a lower value \deleted{(10–15)} \added{$\approx 18$} is observed (base hours).

\subsection{Computational time}
\added{The computational time required for training and testing each model is summarised in Table
\ref{tab:computational_time}. Experiments were conducted on a machine equipped with an Intel Core i9-14900HX CPU (up to 5.60 GHz), 64 GB RAM, and an NVIDIA GeForce RTX 4070 Max-Q GPU using Python 3.12.8\\}
\begin{table}[h!]
  \centering
  \setlength{\tabcolsep}{10pt}
  \renewcommand{\arraystretch}{1.2}
  \begin{tabular}{lccc}
    \toprule
    \textbf{Model} & \textbf{Training Time (min)} & \textbf{Testing Time (s)} & \textbf{Total Time (h)} \\
    \midrule
    RNN         & 899.550       & 108.007   & 14.99  \\
    Kalman Filter & 649.280  & 80.444    & 10.83 \\
    LEM         & 376.060     & 55.880    & 6.27 \\
    LEM-RNN     & 822.570   & 130.218   & 13.71 \\
    KF-RNN      & 1664.400 & 265.341   & 27.75 \\
    LEM-KF-RNN  & 617.980    & 36.850    & 10.30 \\
    LEAR        & -- &194 \footnote{For LEAR, training and testing are performed jointly; therefore, the reported time corresponds to the total runtime for both phases.} & 3.23 \\
    LEAR Fuel+   & -- & 438          & 7.30 \\
    DNN         & 170     & 42660 & 14.70 \\
    DNN Fuel+   & 140   & 380        &  8.67\\
    \bottomrule
  \end{tabular}
  \caption{\added{Computational time for training and testing each model for the Germany–Luxembourg market under the rolling-window retraining scheme.}}
  \label{tab:computational_time}
\end{table}

\added{The computational time required to train and test each model under the rolling-window retraining scheme is presented in Table \ref{tab:computational_time}. Results indicate that linear models, including LEMs and LEARs, are computationally efficient, whereas nonlinear architectures, such as RNNs and DNNs, require substantial training resources. In addition, due to its recursive state and covariance updating, the KF model requires more training time than the LEM, which uses a linear estimation method.\\
  Interestingly, the LEM–KF–RNN hybrid does not have the highest computational cost. This hybrid model incorporates three components but requires a smaller training time (approximately 10.3 hours) than a stand-alone RNN and is significantly shorter than the KF-RNN hybrid model. This result shows that warm-starting and linear expert branches stabilise the nonlinear component by reducing its optimisation requirements.\\
  Based on Figure \ref{fig:trial_durations_bar} (in the Appendix), the training-duration distribution for the LEM–KF–RNN model further indicates that although some Optuna trials require substantial computational resources, most fall within the moderate-to-stable run-time range. This pattern suggests that the optimisation process is reliable with respect to convergence.\\
Across all models we developed (hybrid and standalone models), we observed that testing times were negligible relative to training times. Specifically, the LEM-KF-RNN model's testing time is significantly lower than that of other models, indicating its suitability for real-time forecasting applications. Given its performance and efficiency, the LEM-KF-RNN model is well-suited for electricity price forecasting.}

\added{Beyond the empirical runtime results reported above, it is also informative to discuss how the computational complexity of the proposed framework scales with data resolution and input dimensionality. The scalability of the proposed framework can be discussed along two dimensions. First, the transition from hourly prices (24 observations per day) to higher-frequency prices (e.g., 30-minute or 15-minute prices) primarily increases the sequence length of the RNN branches, thereby increasing computational time.\\
  Second, in both linear (LEM) and Kalman filters, the scale is linear in the number of input features (generally a quadratic row), whereas RNNs scale with both the input dimension and the hidden state size. With the addition of more covariates, regularisation techniques, such as weight decay, as well as feature selection methods, have become increasingly important for controlling overfitting.\\
As a result, the LEM–KF–RNN architecture can be easily adapted to other electricity markets with different data frequencies, since only the input dimensionality and sequence length need to be changed.}

\subsubsection{Decomposed forecasts}

\label{sec:decomposed_forecasts}
\begin{figure}[h!]
  \centering
  \begin{subfigure}[b]{0.48\textwidth}
    \centering
    \includegraphics[width=\textwidth]{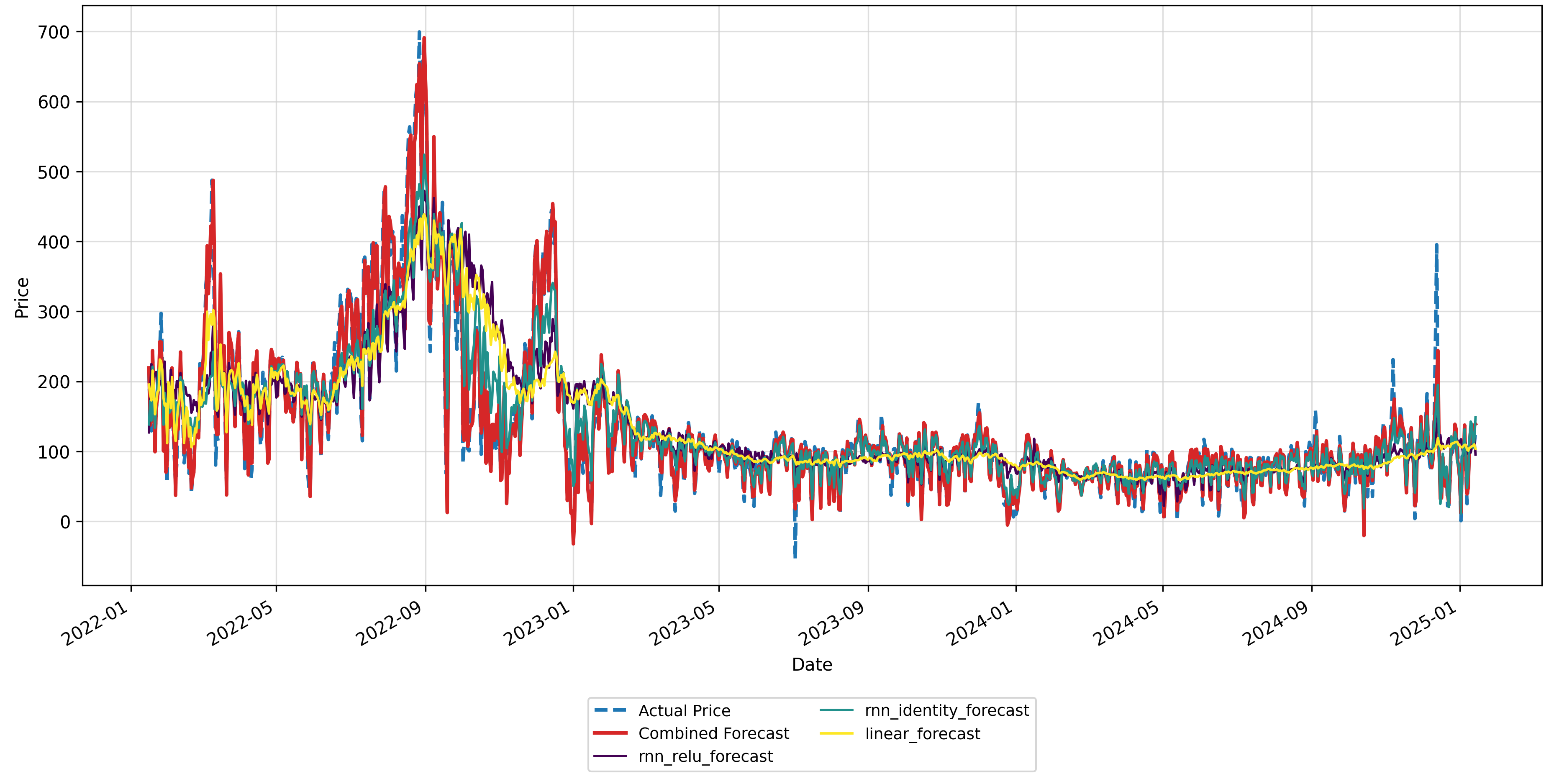}
    \caption{Average Daily Forecasts.}
    \label{fig:decomposed_forecasts_daily}
  \end{subfigure}
  \hfill
  \begin{subfigure}[b]{0.48\textwidth}
    \centering
    \includegraphics[width=\textwidth]{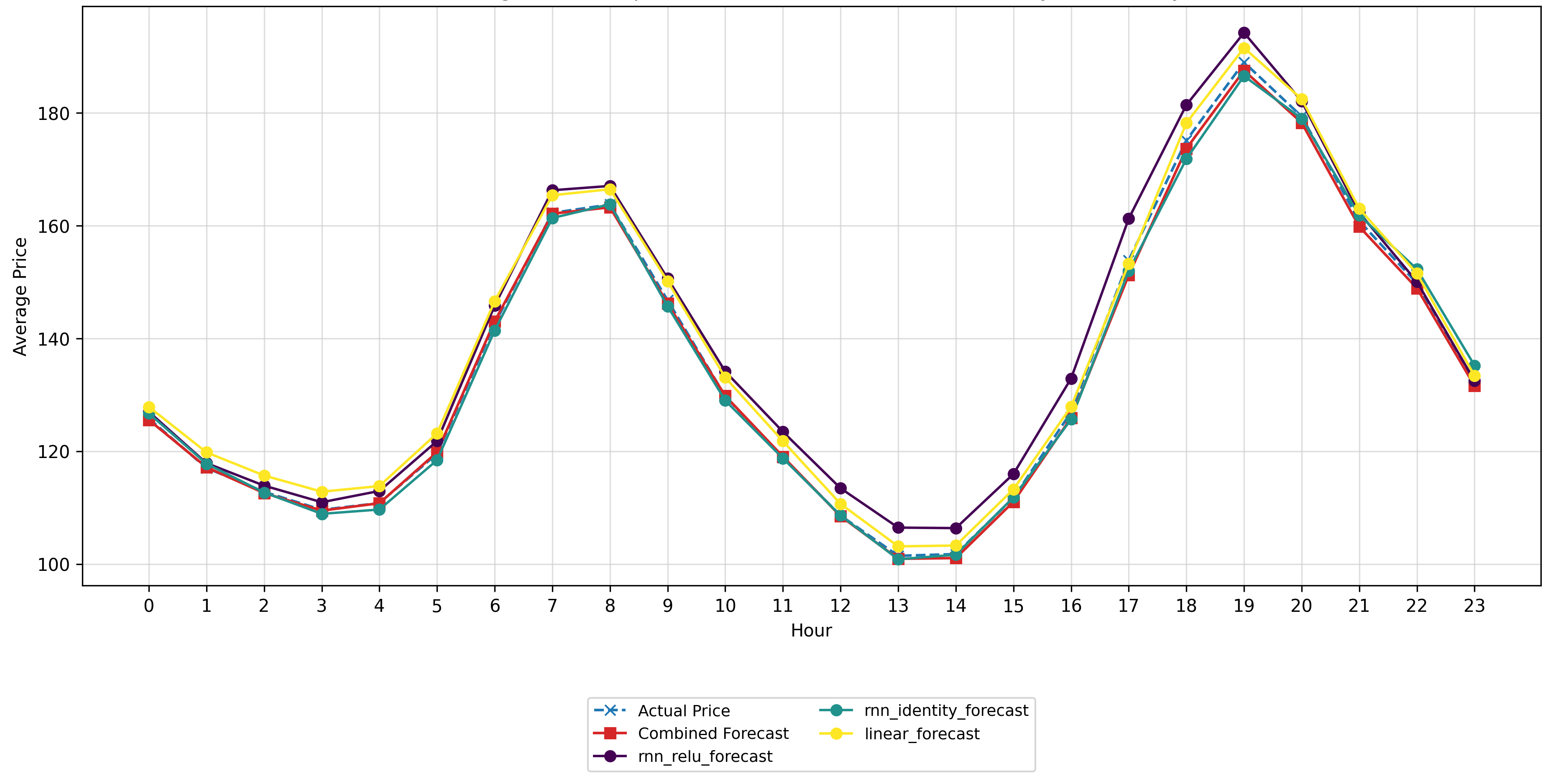}
    \caption{Average Hourly Forecasts.}
    \label{fig:decomposed_forecasts_hourly}
  \end{subfigure}

  \caption{%
    \added{Decomposed, Combined \& Real Electricity Prices for the Test Period.
      Subfigure~(a) shows the \textcolor{blue}{real price values (dashed blue line)} compared with forecasted values from
      the \textcolor{red!90!black}{Combined LEM–KF–RNN model}, which represents the aggregated forecast from all model branches:
      \textcolor{darkpurple}{RNN with ReLU forecast} and \textcolor{teal}{RNN with Identity forecast} correspond to the two RNN branches (ReLU and linear activations),
      and \textcolor{yellow!80!black}{the LEM model component}.
      Subfigure~(b) presents the same decomposition at the hourly level for the test set.
    All values are unstandardized.}
  }
  \label{fig:decomposed_forecasts_combined}
\end{figure}

\begin{figure}[H]
  \centering
  \begin{subfigure}[b]{0.48\textwidth}
    \centering
    \includegraphics[width=\textwidth]{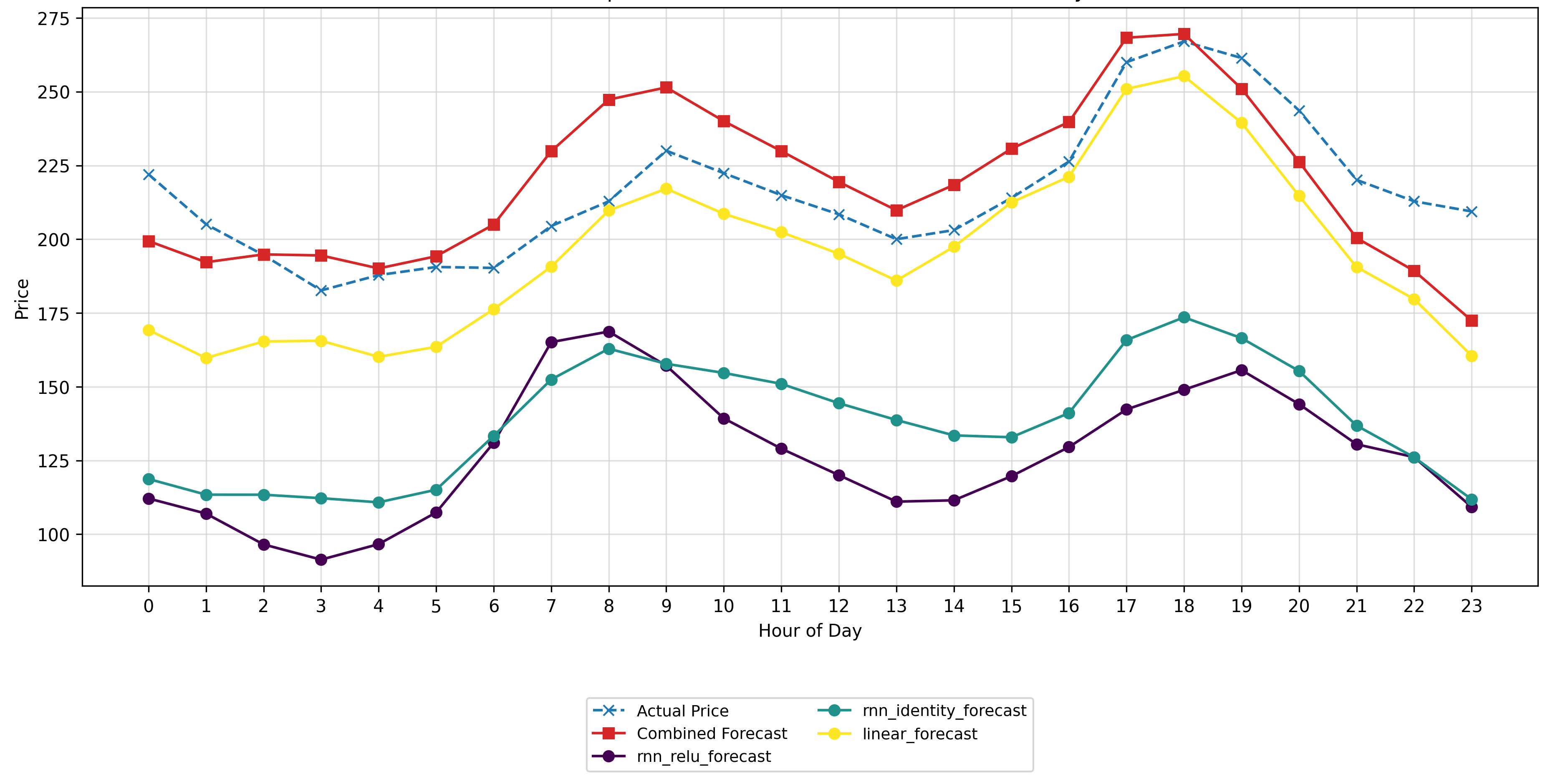}
    \caption{Decomposed Forecasts - First Day Test Set.}
    \label{fig:decomposed_forecasts_first_day}
  \end{subfigure}
  \hfill
  \begin{subfigure}[b]{0.48\textwidth}
    \centering
    \includegraphics[width=\textwidth]{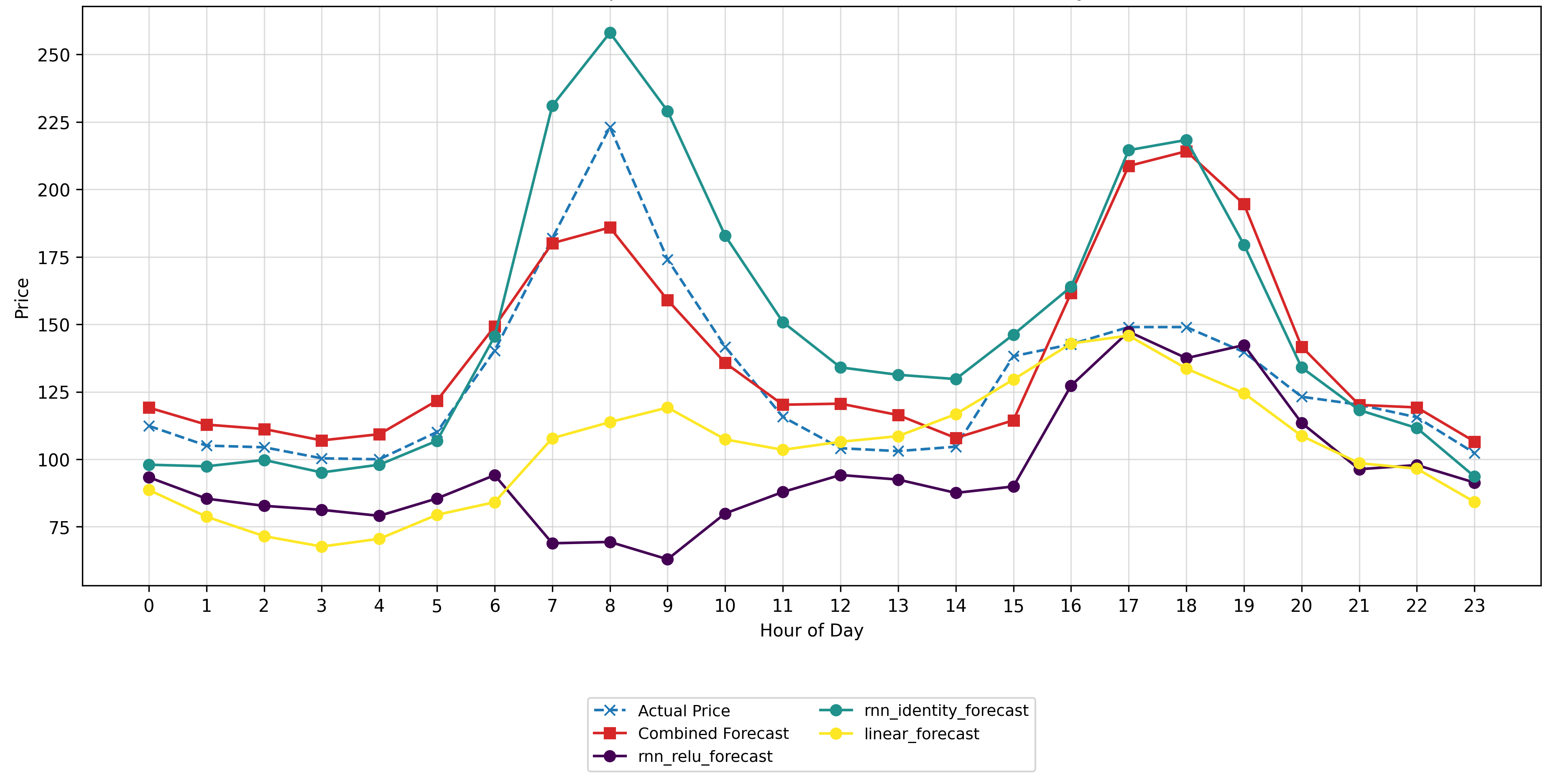}
    \caption{Decomposed Forecasts - Last Day Test Set.}
    \label{fig:decomposed_forecasts_last_day}
  \end{subfigure}
  \caption{\added{Decomposed Forecasts for the First and Last Day of the Test Set for the German-Luxembourg Market.}}
  \label{fig:decomposed_forecasts_days}
\end{figure}

A comprehensive view of the behaviour of the hybrid LEM-KF-RNN model and its comparison to the real electricity price and different forecast components for two years out-of-sample testing period is presented in Figures \ref{fig:decomposed_forecasts_combined} and \ref{fig:decomposed_forecasts_days}. The Figures illustrate how the combined forecast gains the advantage from its linear and nonlinear components in enhancing forecasting accuracy on both a daily and hourly basis.\\
To better visualize the individual forecasts alongside the combined forecast and actual prices, each component was unstandardized separately following Equation (\ref{eq:destandardization}).\\
Let $\widehat{\boldsymbol{Y}}^{\text{std}}_t(c)$ denote the standardized forecast of component $c$ ($c={\text{LEM}, \text{KF}, \text{RNN}}$); then the corresponding unstandardized forecast $\widehat{\boldsymbol{Y}}_t(c)$ is obtained as:
\begin{equation}
  \widehat{\boldsymbol{Y}}_t(c) = \widehat{\boldsymbol{Y}}^{\text{std}}_t(c) \sigma_y + \mu_y
  \label{eq:dest_comp}
\end{equation}
It is important to note that the de-standardized combined forecast (Equation \ref{unstadard_comb}) is not the direct sum of the de-standardized components, since the summation is performed on the standardized values. Consequently, the combined forecast time series (shown in red in Figures \ref{fig:decomposed_forecasts_combined} and \ref{fig:decomposed_forecasts_days}) does not represent a simple sum of the individual unstandardized forecasts.\\
First, Figure \ref{fig:decomposed_forecasts_daily} illustrates the difference in average daily forecasts for stand-alone and combined models. It illustrates the fact that the combined forecast (red line) closely tracks the actual price (dashed line), capturing both long-term trends and short-term fluctuations. As for the model's components, we noticed that the LEM component (yellow) can model baseline structural movements, which is consistent with its ability to track smooth, low-frequency variations in electricity prices. Alternatively, the RNN components (ReLU and identity activations) capture high-frequency deviations such as peaks, drops, and volatility spikes, with the RNN model with ReLU showing greater superiority over the KF component. Therefore, combining these different components results in a model that is strong and interacts well, enabling the aggregated output to track persistent trends and transient shocks, thus predicting prices with high accuracy regardless of market conditions.\\
Secondly, Figure \ref{fig:decomposed_forecasts_hourly} shows forecasts by hour of the day, confirming that the combined model follows the intraday demand-price cycle accurately. There is a great deal of agreement between the predicted and actual hourly patterns, especially during peak hours in the afternoon and evening, when volatility is high. The intraday price cycle was also captured by the model's components with variation in amplitude and accuracy. 

Lastly, Figure \ref{fig:decomposed_forecasts_days} examine component contributions during the first and last test days.
In the first test day plot (Figure \ref{fig:decomposed_forecasts_first_day}) for Sunday, 15 January \deleted{2023} \added{2022}, real electricity prices remained relatively stable \deleted{and very low.} This may have been the result of a calm market on the weekend or an increased share of renewables, which reduces price volatility. \deleted{In spite of this, the linear component still produces a typical daily cycle with pronounced peak and base hours, thus predicting variation that does not occur in reality.
Alternatively, the RNN branches respond more adaptively: the identity-based branch exhibits inverse movement of the LEM model, thus cancelling the overestimation of the latter and reducing artificial peaks introduced by it, a process further supported by the ReLU-based branch. By combining these adjustments, the combined model can generate a more realistic forecast that reflects the dynamics of a low-volatility day. Therefore, the combined forecast seems close to that day's stable market behaviour.\\}
\added{The linear component (LEM) generally seems close to that day's stable market behaviour. In addition, the identity and ReLU branches of RNNs produce forecasts that are substantially lower than observed prices, indicating they are not capturing the high price level on this particular day. It appears, however, that the combined model corrects the common underestimation bias in the standalone components by tracking the actual series more closely throughout the day, especially during the evening ramp.\\}
According to the last test day plot (Figure \ref{fig:decomposed_forecasts_last_day}) (Monday, 13 January 2025), prices show clear intraday variability, with peaks in early morning and late evening, which will be typical of winter demand on a working day. \deleted{In general, the linear model captures the shape of these peaks, but underestimates their intensity. Alternatively, RNNs seem more capable of forecasting the amplitude of price peaks and troughs, resulting in a forecast that's more accurate at capturing short-term volatility and follows actual price dynamics.\\} \added{The linear component (LEM) primarily captures a smooth, low-frequency structure and fails to capture the abrupt intraday spike, leading to a clear underestimation of peak intensity.
We also noticed that the RNN branches counteract each other by attenuating extreme movements. This occurs in the mornings (8 am) when the identity-based branch overestimates the morning spike, whereas the ReLU-based branch attenuates this behavior by showing inverse movement. As a result of their interaction, the combined forecast shows a moderate spike response that follows the observed dynamics without excessive amplitude or damping.}\\
In summary, the decomposition plots illustrate how different components contribute to the combined model depending on market conditions. \deleted{Although LEM models can impose typical daily patterns even on stable days, RNN and KF components respond adaptively, suppressing unnecessary cycles when the market is calm and emphasising them when strong intraday dynamics appear. This adaptive equilibrium renders the hybrid forecast more realistic and more aligned with actual price fluctuations.} \added{They illustrat how the hybrid architecture leverages nonlinear interactions to stabilise forecasts under high-volatility intraday conditions, rather than relying on the linear component to explain short-term dynamics.}

\begin{figure}[h!]
  \centering

  \begin{subfigure}[b]{0.32\textwidth}
    \centering
    \includegraphics[width=\textwidth]{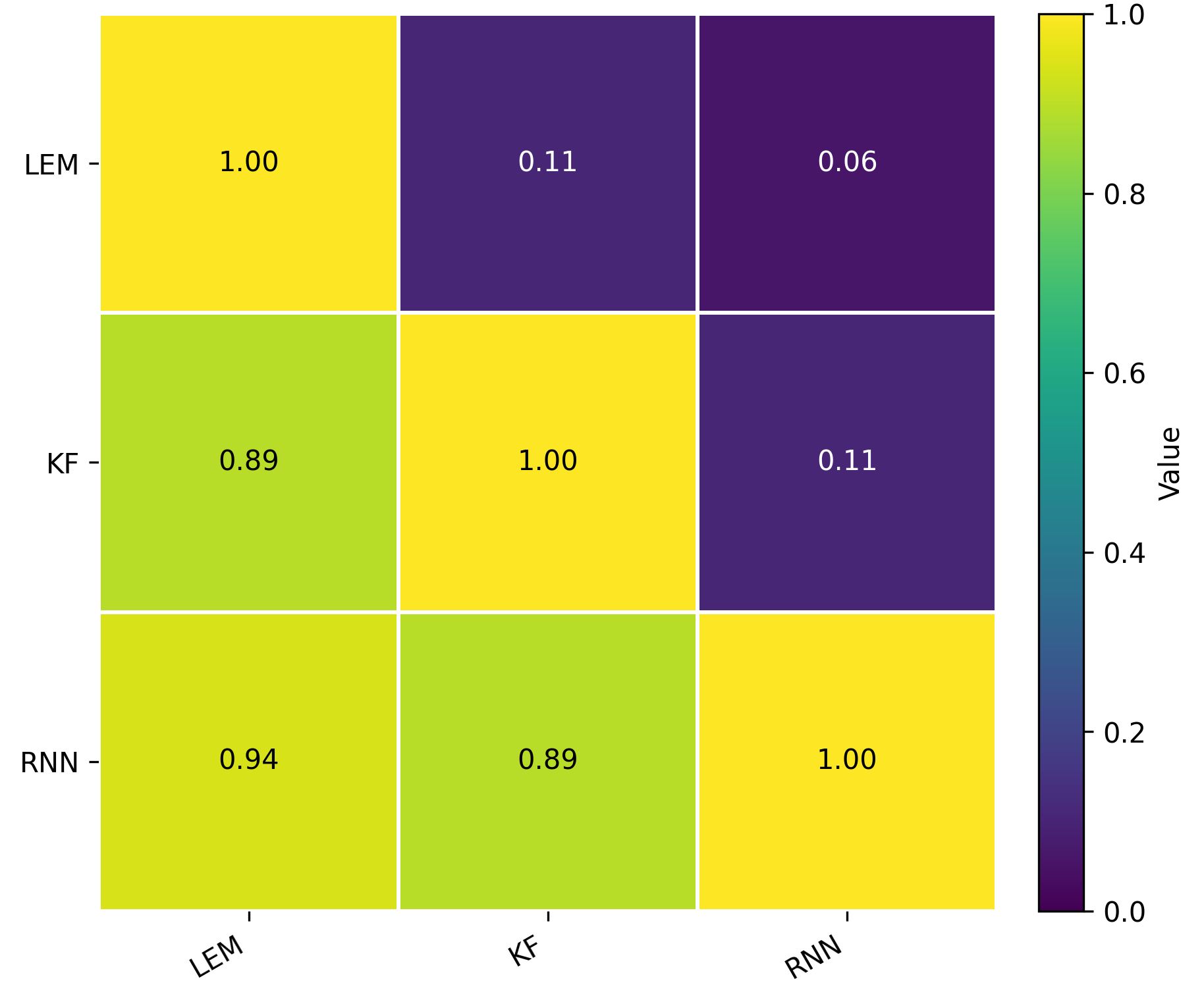}
    \caption{Hour 04}
    \label{fig:corr_dist_err_h04}
  \end{subfigure}
  \hfill
  \begin{subfigure}[b]{0.32\textwidth}
    \centering
    \includegraphics[width=\textwidth]{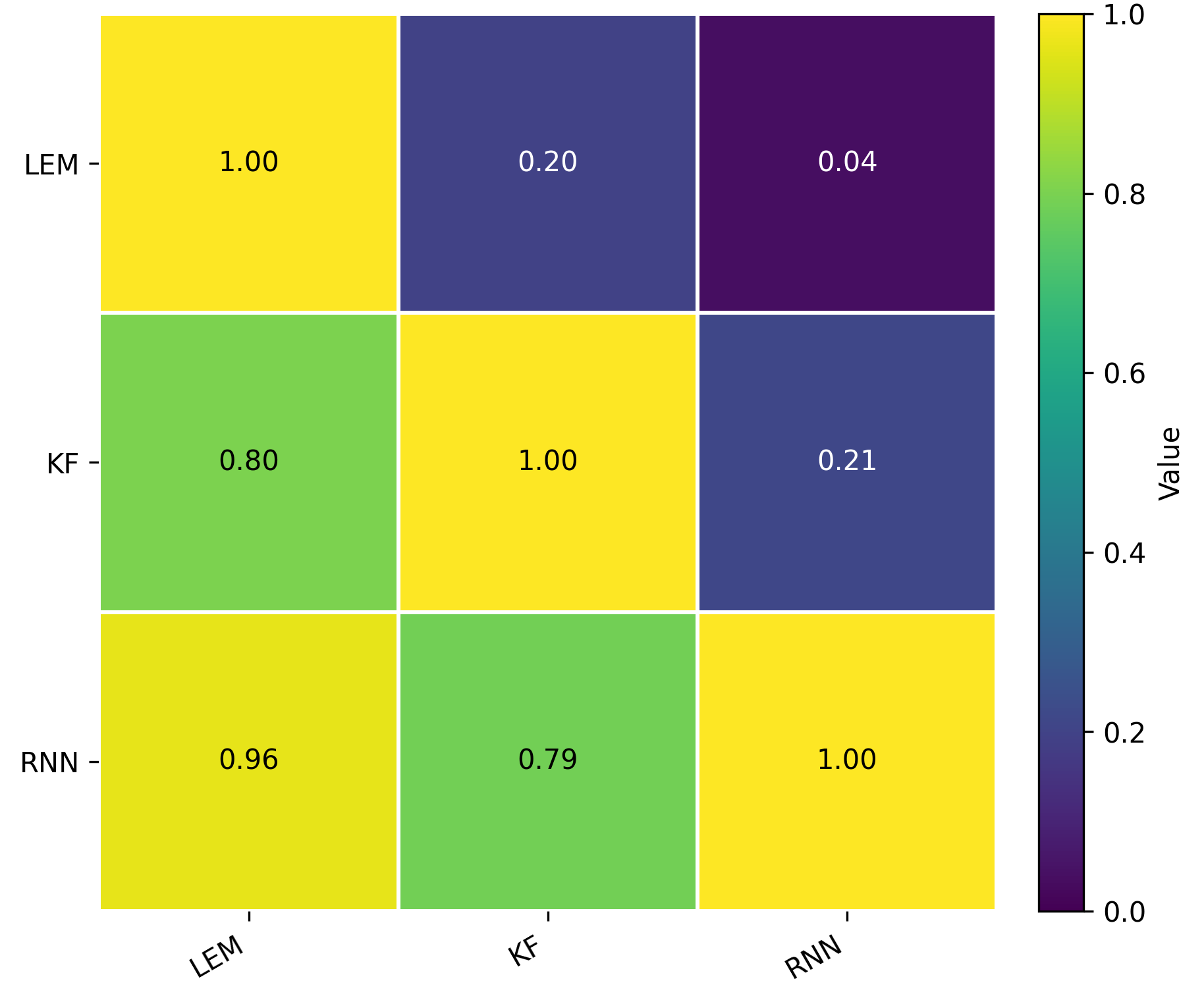}
    \caption{Hour 08}
    \label{fig:corr_dist_err_h08}
  \end{subfigure}
  \hfill
  \begin{subfigure}[b]{0.32\textwidth}
    \centering
    \includegraphics[width=\textwidth]{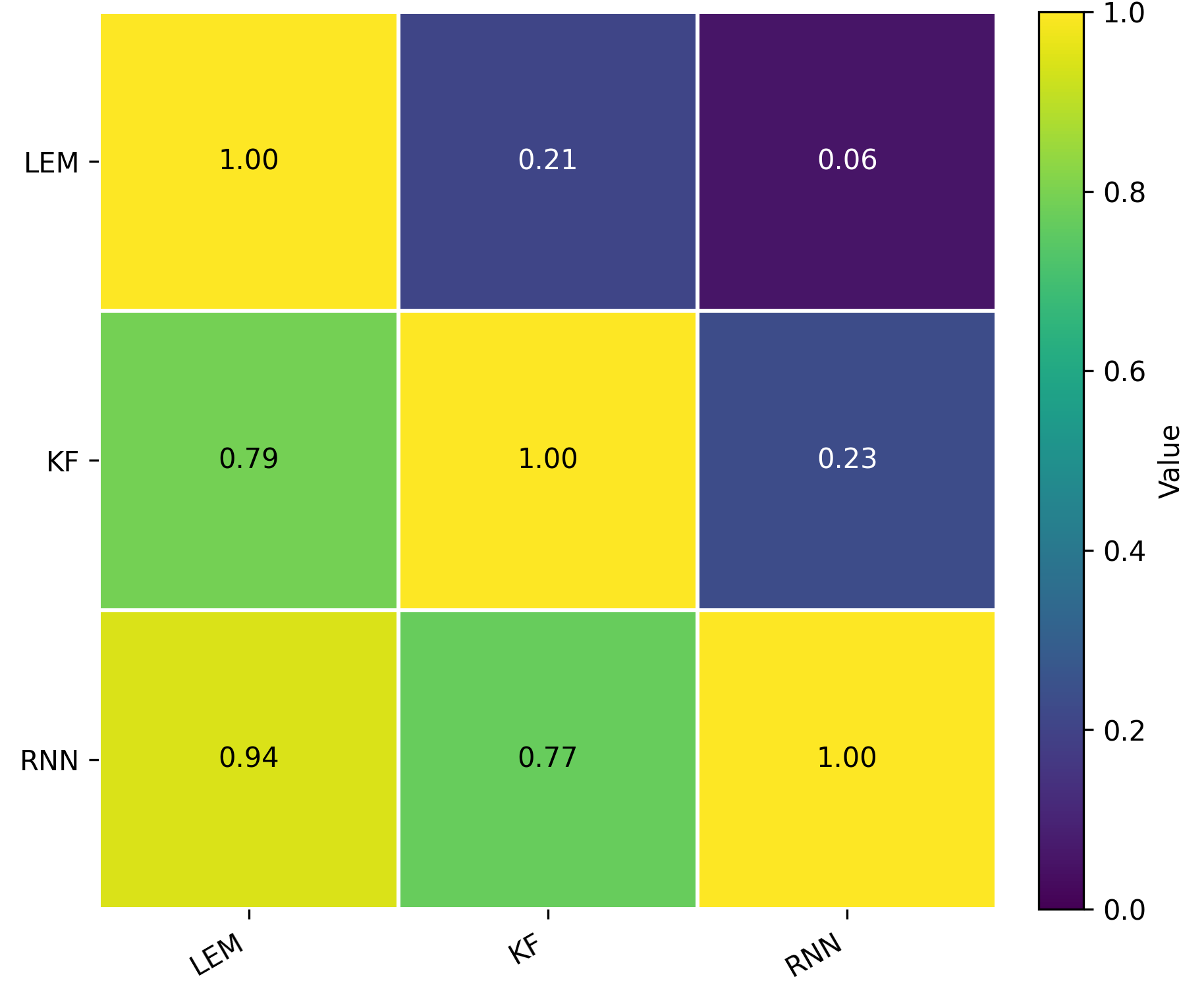}
    \caption{Hour 12}
    \label{fig:corr_dist_err_h12}
  \end{subfigure}

  \vspace{0.4cm}

  \begin{subfigure}[b]{0.32\textwidth}
    \centering
    \includegraphics[width=\textwidth]{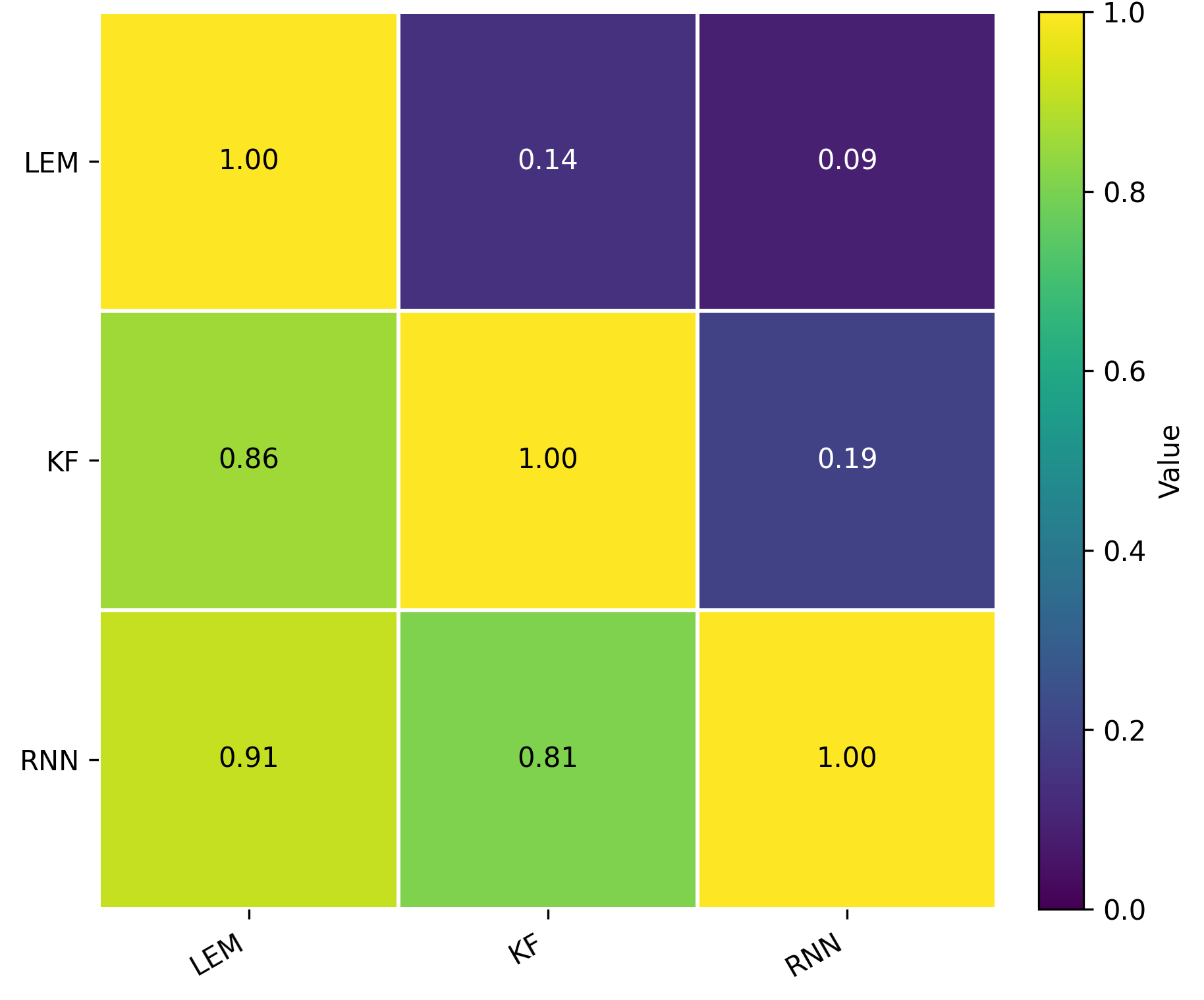}
    \caption{Hour 16}
    \label{fig:corr_dist_err_h16}
  \end{subfigure}
  \hfill
  \begin{subfigure}[b]{0.32\textwidth}
    \centering
    \includegraphics[width=\textwidth]{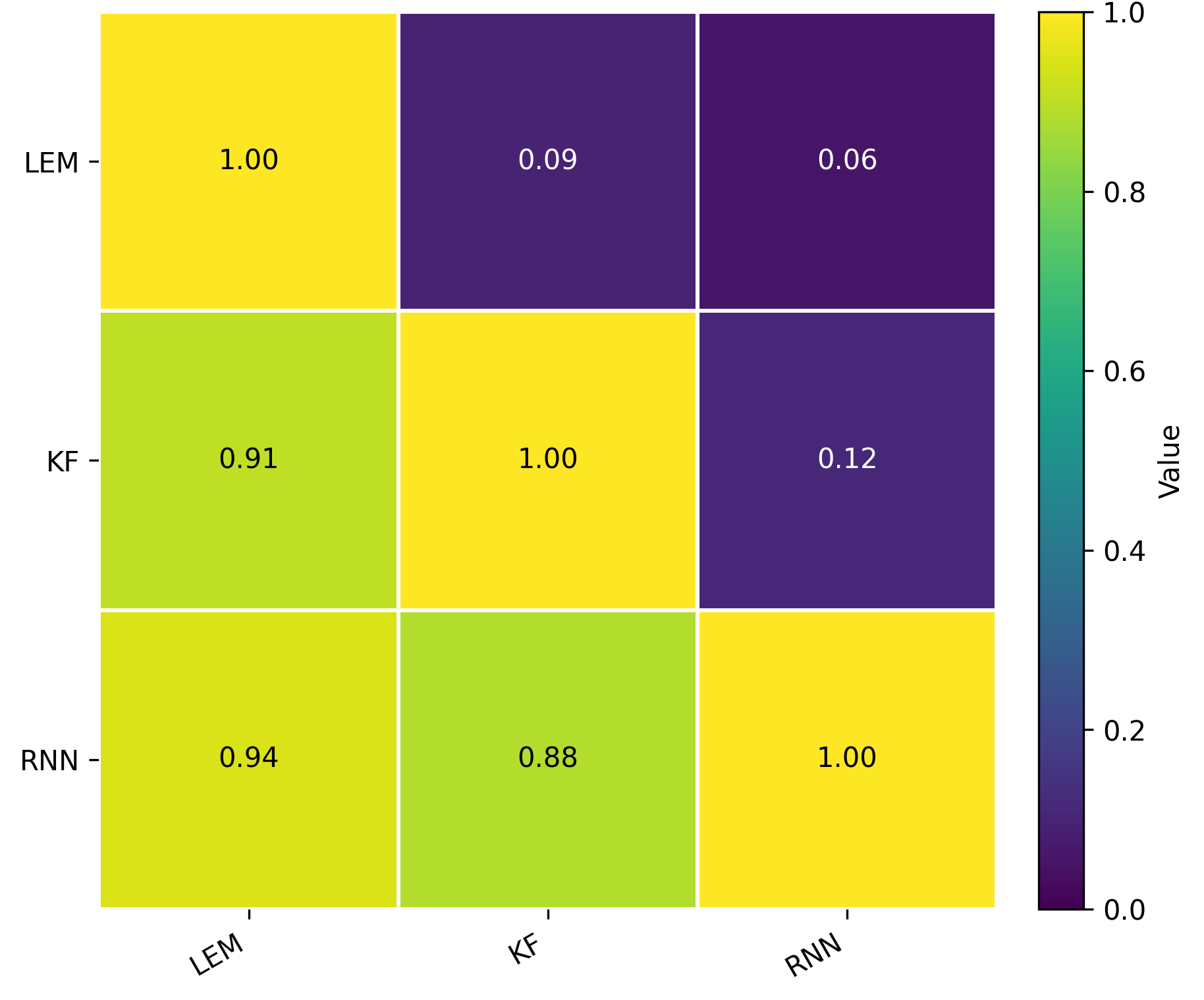}
    \caption{Hour 20}
    \label{fig:corr_dist_err_h20}
  \end{subfigure}
  \hfill
  \begin{subfigure}[b]{0.32\textwidth}
    \centering
    \includegraphics[width=\textwidth]{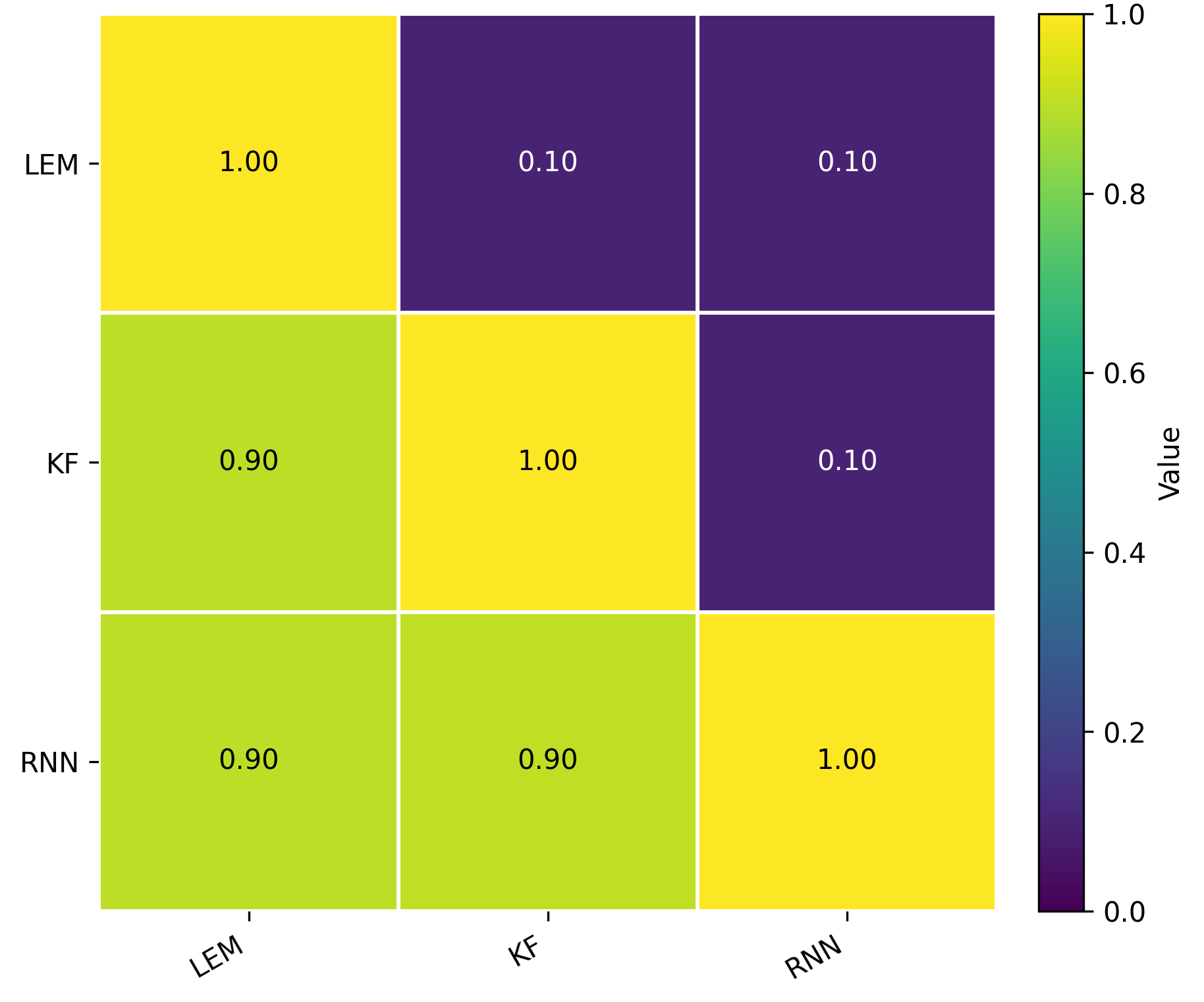}
    \caption{Hour 24}
    \label{fig:corr_dist_err_h24}
  \end{subfigure}

  \caption{\added{Merged correlation and distance heatmaps of forecast errors for the LEM-KF-RNN model. The lower triangular part of each matrix shows the Pearson correlation between component-wise forecast errors, while the upper triangular part shows the corresponding correlation distance $1-\mathrm{corr}$. Results are shown for selected hours of the day.}}
  \label{fig:merged_corr_dist_errors_model7}
\end{figure}
\added{To further investigate how the different components of the hybrid LEM-KF-RNN model interact, we computed Pearson correlation coefficients and correlation distances for each component's error series at selected hours of the day. Figure \ref{fig:merged_corr_dist_errors_model7} presents heatmaps combining correlation and distance information for hours 04, 08, 12, 16, 20, and 24 (averaged across test sample). The lower triangular part of each matrix displays the Pearson correlation coefficients between the forecast errors of the LEM, KF, and RNN components, while the upper triangular part shows the corresponding correlation distances (1 - correlation).\\
  By examining how component forecast errors interact, these measures provide insights into whether components operate independently or in a coordinated manner. Results indicate that the error components are positively correlated and have low correlation distances. \\
  The magnitude varies across hours of the day and among pairs of error components. The lower value appears most often between KF and RNN, especially during hour 12, as with KF and LEM. The correlation distances remain non-negligible, indicating that while the errors are aligned, they are not identical. Therefore, the error correlation analysis indicates that the hybrid model components are aligned rather than orthogonal, so the model components cannot be considered separate experts identifying distinct error patterns.\\
  This pattern likely reflects the nature of electricity prices rather than redundancy. Electricity prices are influenced by systemic disruptions and structural factors, such as supply-demand mismatches, rising fuel costs, and severe market pressures. All well-designed forecast models, linear or non-linear, will likely produce similar forecast errors under these conditions. Low or negative error correlations would be unexpected and may indicate that a component is missing key market trends.\\
According to these findings, the hybrid model works within a coherent system, in which components remain aligned in response to shared electricity price shocks while contributing complementary refinements through distinct modelling mechanisms. In the linear component, stable economic relationships and smooth temporal dynamics are captured, whereas in the RNN, these signals are refined by modelling nonlinear and higher-order temporal effects. Its important to note that despite its strong coherence, evaluation results (in Table \ref{tab:forecasting_metrics_Germany}) of the hybrid model indicate a lower RMSE than any individual component, suggesting that performance improvements arise not from diversification but from bias correction.\\}

\subsubsection{Probabilistic forecast extension}
\label{probabilistic}
\added{EPF is inherently subject to substantial uncertainty due to demand variability, renewable generation intermittency, fuel price shocks, and structural market changes. While point forecasts remain useful for benchmarking, many operational decisions in electricity markets (such as risk management, bidding strategies, reserve allocation, and portfolio hedging) require an explicit quantification of forecast uncertainty. In response, we extend our framework beyond point forecasting and introduce a probabilistic forecasting layer that provides full predictive intervals alongside the median forecast.\\
  In our point forecast architecture, we have already incorporated rolling retraining, L1/L2 regularisation, and a rolling-window-based retraining infrastructure that favours easy probabilistic framework extension, making them readily amenable to probabilistic extensions. For this purpose, the MSE objective function is replaced by the pinball loss in the quantile regression model, enabling direct estimation of conditional quantiles. For every forecasting step, the model produces the $10\%$, $50\%$, and $90\%$ conditional quantiles ($P10$, $P50$, $P90$), where the median is used as the point forecast and the range P10-P90 represents an 80\% forecasting range. Specifically, we extend the probabilistic forecast for the LEM-KF-RNN model.\\
Numerous advantages can be gained from probabilistic forecasting. First, it is a good fit for electricity prices, which are known to exhibit heavy tails and asymmetric shocks, because quantile regression requires no parametric assumptions about the predictive distribution. Further, the prediction interval ensures that uncertainty is transparently interpreted. It's worth noting that volatility clustering in the data underscores the importance of probabilistic forecasting. A closer look at Table \ref{tab:descriptive_stats} indicates that the price standard deviation has increased from  53 to 115 in the post-crisis period, while extreme outliers remain prevalent. Given that this volatility and tail risk persistence are signs of heteroskedastic dynamics, which in fact cannot be captured by point forecasting.\\}
\begin{figure}[H]
  \centering
  \begin{subfigure}{0.9\textwidth}
    \centering
    \includegraphics[width=\textwidth]{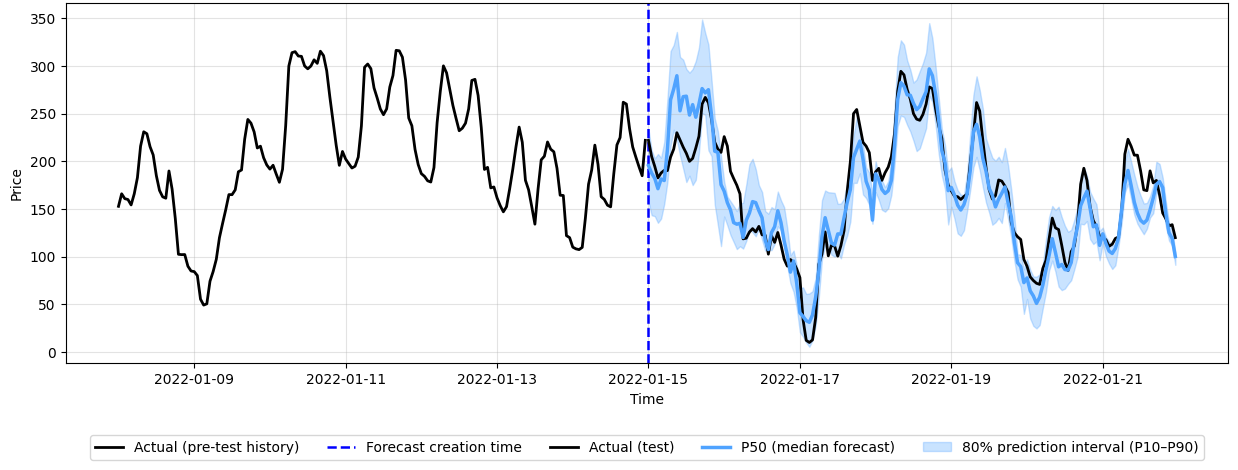}
    \caption{Probabilistic forecast for the first test week.}
    \label{fig:prob_forecast_first_week}
  \end{subfigure}

  \vspace{0.5cm}

  \begin{subfigure}{0.9\textwidth}
    \centering
    \includegraphics[width=\textwidth]{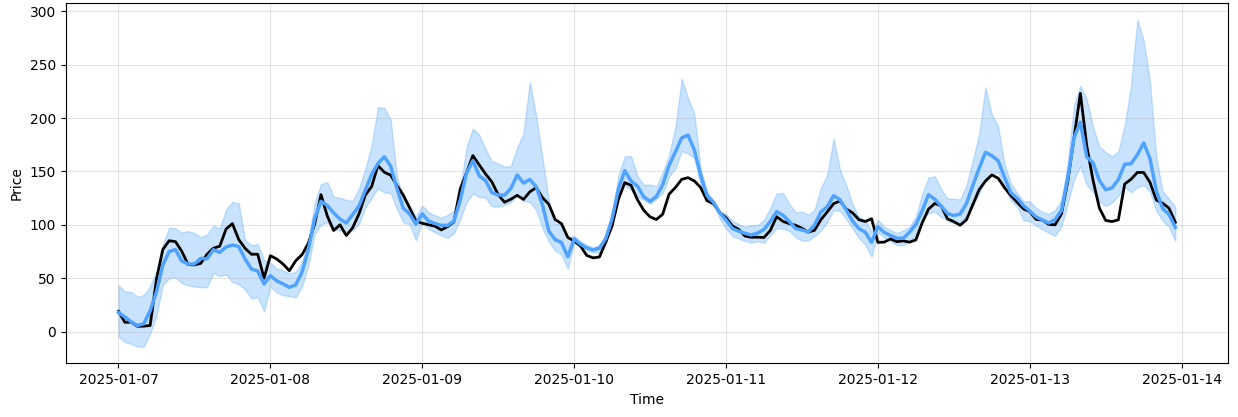}
    \caption{Probabilistic forecast for the last test week.}
    \label{fig:prob_forecast_last_week}
  \end{subfigure}

  \caption{\added{Probabilistic forecasts for the LEM-KF-RNN model on selected test periods.
  The shaded area represents the 80\% prediction interval (P10--P90), while the solid line indicates the median forecast (P50).}}
  \label{fig:probabilistic_forecast}
\end{figure}
\added{Figure \ref{fig:probabilistic_forecast} illustrates representative probabilistic forecasts for the first (Subfigure \ref{fig:prob_forecast_first_week}) and last (Subfigure \ref{fig:prob_forecast_last_week}) weeks of the test sample. Based on the results, the proposed model produces well-calibrated intervals that expand during periods of elevated volatility and contract when price dynamics stabilise, while maintaining accurate point forecasts.\\
From an applied perspective, the probabilistic outputs significantly increase the model's utility for market participants. Moreover, these benefits can be achieved without sacrificing interpretability or adding complexity to the model. It can also be extended to the LEM-KF-RNN components separately, as well as to the other combinations (defined in Section \ref{sec:architectures}) }

\section{Summary and Conclusion}
\label{conclusion_section}
Electricity price forecasting has been a challenging task due to the complex nature of the underlying electricity markets. A wide variety of EPF models have been studied in the past, including statistical models and machine learning techniques, as well as hybrid frameworks that combine both of type models. In this context, our paper introduced a new hybrid framework that combines a linear expert model (LEM) with RNNs and KF using a parallel-branch neural network architecture. Hence, the proposed architecture benefits from the strength of linear and nonlinear methods, making it possible to model complicated market interdependencies in time and structure. Our methodology is empirically tested on the day-ahead German electricity market using hourly data covering the period between 2018 and 2025. we also considered a set of input features that reflet the main drivers of electricity prices. to assess the accuracy of our model we benchmark it with state -of-the-art models from the literature.
Results demonstrate the improvement in forecasting accuracy of the combined architectures compared to stand-alone models. From the stand-alone model results, the RNN model outperformed the linear LEM and KF models, emphasising the importance of nonlinearity dynamics estimation in electricity price. Concerning the hybrid models, our findings demonstrated that the best model performance was realised by the LEM-KF-RNN hybrid model, closely followed by the KF-RNN hybrid model. This emphasizes the importance of incorporating time sequence learning into both nonlinear and linear model components.

Concerning the decomposition analysis, although the RNN components are shown to be dynamically changing with highly volatile market conditions, variable relationships can be explained using the LEM branch by providing interpretability. Moreover, the KF branch enhances linear temporal dependencies, affirming the complementary function of state-space modelling within the hybrid model.

These results align with the findings from \cite{lago2021forecasting} and \cite{amor2024bridging} that support the outperformance of deep learning models over statistical models for electricity price forecasting. Our findings also align with a lot of research which developed hybrid or combined models (e.g., \cite{ben2018forecasting}, \cite{zhang2020adaptive}, etc.), who demonstrated that combining linear and non-linear models improves forecasting accuracy compared to both single components. Another advantage of such hybrid model types is discussed by \cite{rondon2025advancements} in the context of energy consumption forecasting.

\added{Morover, since probabilistic forecasting is important in decision-making related to electricity trading and system operations, the proposed model architecture is be extended to probabilistic forecasting (\cite{ziel2018probabilistic} and \cite{nowotarski2018recent})}.

Our methodology can be further developed in future research. \added{First, by conducting counterfactual "what-if" analyses by perturbing selected primary economic inputs (such as gas prices) and then examining their effects on linear and nonlinear functions. The knowledge gained from such experiments can be used to deepen understanding of the model’s inner workings and to improve its interpretability under more complex market conditions.} Second, investigating more advanced state-space models, such as the Mamba model \cite{gu2024mamba}, as this model allows more expressive and adaptive (dynamic) temporal dependencies. Third, rather than employing a linear summing of the component forecasts in the neural network output layer, the model could dynamically and adaptively ascertain the contribution of each branch, identify the most effective components based on forecast accuracy, and assign appropriate weights (\cite{xu2024novel} and \cite{palou2025novel}). \added{Finally, our methodology could be further tested on other electricity markets with different characteristics (e.g., markets with high renewable penetration or different regulatory frameworks) to assess its generalisability and robustness across diverse market conditions.}

\section*{Declaration of competing interest}
The authors declare that they have no known competing financial interests or personal relationships that could have appeared to influence the work reported in this paper.

\section*{Acknowledgments}
This research was partially funded in the course of TRR 391 Spatio-temporal Statistics for the Transition of Energy and Transport (520388526) by the Deutsche Forschungsgemeinschaft (DFG, German Research Foundation).

\appendix
\section*{Appendix}
\label{appendix:add_results}

\begin{table}[!ht]
  \centering
  \small
  \caption{Descriptive statistics of the data used for the initial calibration window and the out-of-sample test period.}
  \label{tab:descriptive_stats}

  \begin{tabular}{lrrrrrrr}
    \toprule
    \textbf{Series}
    & \textbf{Mean}
    & \textbf{Std}
    & \textbf{Min}
    & \textbf{Q25}
    & \textbf{Median}
    & \textbf{Q75}
    & \textbf{Max} \\
    \midrule

    \multicolumn{8}{l}{\textbf{In-sample data (01.10.2018--14.01.2022)}} \\
    Price (EUR/MWh)              & 56.0  & 53.3  & -90.0  & 30.7  & 42.9  & 60.6  & 620.0 \\
    Load forecast (MWh)          & 55.1  & 9.3   & 32.4   & 47.4  & 55.0  & 63.0  & 77.6 \\
    Wind forecast (MWh)          & 14.1  & 10.1  & 0.2    & 6.1   & 11.5  & 19.9  & 47.2 \\
    PV forecast (MWh)            & 4.8   & 7.6   & 0.0    & 0.0   & 0.1   & 7.4   & 36.4 \\
    EUA (EUR/tCO$_2$)            & 33.8  & 16.1  & 15.2   & 23.8  & 26.3  & 41.8  & 88.9 \\
    API2 Coal (EUR/t)            & 68.7  & 35.4  & 34.7   & 45.9  & 54.2  & 77.4  & 236.7 \\
    TTF Gas (EUR/MWh)            & 24.5  & 24.9  & 3.5    & 11.3  & 15.9  & 24.4  & 180.3 \\
    Brent Oil (EUR/bbl.)         & 52.6  & 12.3  & 17.8   & 42.6  & 55.2  & 60.9  & 75.4 \\
    \midrule

    \multicolumn{8}{l}{\textbf{Out-of-sample data (15.01.2022--13.01.2025)}} \\
    Price (EUR/MWh)              & 135.4 & 115.8 & -500.0 & 73.0  & 102.2 & 161.7 & 936.3 \\
    Load forecast (MWh)          & 53.6  & 9.1   & 30.5   & 46.1  & 53.4  & 61.1  & 75.9 \\
    Wind forecast (MWh)          & 15.4  & 11.5  & 0.2    & 6.0   & 12.3  & 22.7  & 50.3 \\
    PV forecast (MWh)            & 6.7   & 10.2  & 0.0    & 0.0   & 0.2   & 10.7  & 48.2 \\
    EUA (EUR/tCO$_2$)            & 76.5  & 10.3  & 50.5   & 67.9  & 77.5  & 84.9  & 97.6 \\
    API2 Coal (EUR/t)            & 166.1 & 90.6  & 86.6   & 105.8 & 114.7 & 216.8 & 402.7 \\
    TTF Gas (EUR/MWh)            & 68.9  & 53.9  & 23.0   & 34.2  & 43.2  & 92.3  & 310.5 \\
    Brent Oil (EUR/bbl.)         & 81.4  & 12.1  & 62.7   & 72.1  & 77.9  & 88.0  & 117.5 \\
    \bottomrule
  \end{tabular}
\end{table}

\begin{figure}[h!]
  \centering
  \includegraphics[width=0.9\textwidth]{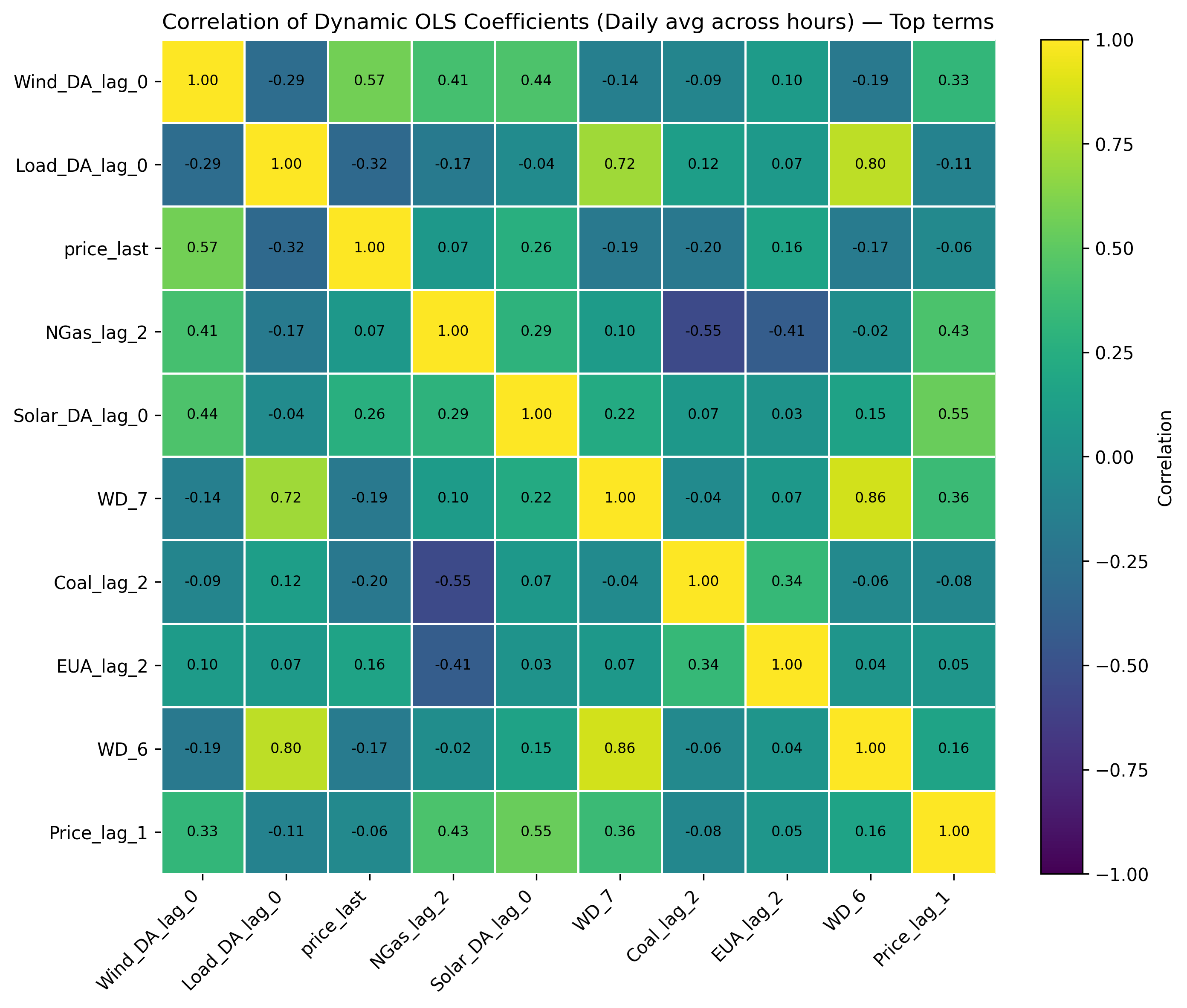}
  \caption{\added{Correlation matrix of LEM coefficients over the test period (averaged daily across hours).}}
  \label{fig:ols_weights_correlation}
\end{figure}

\begin{table}[H]
  \centering
  \caption{\added{Comparison of tuned hyperparameters across all model types (Optuna, TPE).}}
  \label{tab:hyperparams_all_models}
  \resizebox{\textwidth}{!}{%
    \begin{tabular}{lccccccc}
      \toprule
      \textbf{Symbol} & \textbf{RNN} & \textbf{LEM} & \textbf{KF} & \textbf{LEM-RNN} &
      \textbf{KF-RNN} & \textbf{LEM-KF-RNN} \\
      \midrule
      $H$ & 110 & -- & 35 & 76 & 111 & 53 \\[4pt]
      $L$ & 1 & -- & 1 & 1 & 1 & 1 \\[4pt]
      $D_{\mathrm{init}}$ & 540 & 206 & 550 & 114 & 595 & 263 \\[4pt]
      $D_{\mathrm{all}}$ & 67 & 65 & 66 & 68 & 68 & 67 \\[4pt]
      $\eta_{\mathrm{init}}$ & $5.51\times10^{-3}$ & $1.85\times10^{-5}$ &
      $7.56\times10^{-4}$ & $6.72\times10^{-5}$ &
      $6.22\times10^{-5}$ & $5.49\times10^{-3}$ \\[4pt]
      $\eta_{\mathrm{all}}$ & $1.25\times10^{-4}$ & $5.24\times10^{-4}$ &
      $1.18\times10^{-4}$ & $1.10\times10^{-4}$ &
      $1.20\times10^{-4}$ & $1.08\times10^{-4}$ \\[4pt]
      $\lambda_{w,\mathrm{init}}$ & $4.84\times10^{-5}$ & $1.93\times10^{-4}$ &
      $1.28\times10^{-3}$ & $5.81\times10^{-5}$ &
      $6.73\times10^{-4}$ & $1.73\times10^{-5}$ \\[4pt]
      $\lambda_{w,\mathrm{all}}$ & $1.65\times10^{-4}$ & $4.42\times10^{-3}$ &
      $1.32\times10^{-4}$ & $2.72\times10^{-4}$ &
      $3.23\times10^{-4}$ & $6.54\times10^{-4}$ \\[4pt]
      $\lambda_{1,\mathrm{init}}$ & $8.27\times10^{-5}$ & $1.93\times10^{-4}$ &
      $8.64\times10^{-5}$ & $1.46\times10^{-5}$ &
      $1.13\times10^{-5}$ & $8.28\times10^{-5}$ \\[4pt]
      $\lambda_{1,\mathrm{all}}$ & $4.77\times10^{-3}$ & $1.21\times10^{-3}$ &
      $3.27\times10^{-3}$ & $1.09\times10^{-3}$ &
      $1.15\times10^{-3}$ & $8.94\times10^{-4}$ \\[4pt]
      $\alpha$ & -- & 0.356 & -- &  & -- &  \\[4pt]
      \texttt{use\_ols\_weights} & -- & True & -- & False & -- & False \\[4pt]
      \texttt{Trial\_number} & 431 & 500 & 125 & 323 & 255 & 445 \\[4pt]
      \midrule
      \textbf{Best RMSE} & \textbf{12.044} & \textbf{12.824} &
      \textbf{12.254} & \textbf{11.935} &
      \textbf{11.918} & \textbf{11.932} \\[4pt]
      \bottomrule
    \end{tabular}%
  }
\end{table}

\begin{table}[htbp]
  \centering
  \footnotesize
  \renewcommand{\arraystretch}{1.0}
  \setlength{\tabcolsep}{5pt}

  \begin{tabular}{lcccccc}
    \toprule
    \textbf{Model} & \textbf{MAE} & \textbf{RMSE} & \textbf{rMAE} &
    \textbf{rRMSE$_{\text{naive}}$} & \textbf{rRMSE$_{\text{mean}}$} & \textbf{DA} \\
    \midrule

    \multicolumn{7}{c}{\textbf{Test Year\_1: 2022}}\\[-0.6ex]
    \midrule
    Naive        & 85.650 & 113.849 & 1.000 & 1.000 & 0.489 & 0.625 \\
    RNN          & 29.106 & 39.616  & 0.338 & 0.347 & 0.170 & 0.826 \\
    KF           & 29.703 & 40.326  & 0.347 & 0.354 & 0.173 & 0.831 \\
    LEM          & 29.788 & 40.499  & 0.347 & 0.355 & 0.174 & 0.833 \\
    LEM-RNN      & 29.028 & 39.122  & 0.336 & 0.341 & 0.168 & 0.832 \\
    KF-RNN       & 27.974 & 37.987  & 0.323 & 0.331 & 0.163 & 0.844 \\
    LEM-KF-RNN   & \textbf{27.141} & \textbf{36.892} & \textbf{0.317} & \textbf{0.324} & \textbf{0.159} & \textbf{0.845} \\
    Ens--LEM-RNN & 27.327 & 37.325  & 0.319 & 0.328 & 0.160 & \textbf{0.845} \\
    Ens--KF-RNN & 28.010 & 38.216  & 0.326&   0.335  &0.164&  0.838 \\
    DNN    & 29.743 & 41.518  & 0.347 & 0.365 & 0.178 & 0.826 \\
    DNN/-Fuel          & 30.404 & 42.595  & 0.355 & 0.375 & 0.183 & 0.822 \\
    LEAR   & 34.644 & 48.024  & 0.406 & 0.423 & 0.206 & 0.816 \\
    LEAR/-Fuel         & 37.335 & 53.200  & 0.436 & 0.468 & 0.229 & 0.807 \\

    \midrule
    \multicolumn{7}{c}{\textbf{TestYear\_2: 2023}}\\[-0.6ex]
    \midrule
    Naive        & 33.185 & 47.470 & 1.000 & 1.000 & 0.500 & 0.635 \\
    RNN          & 13.321 & 19.758 & 0.401 & 0.416 & 0.208 & 0.827 \\
    KF           & 14.101 & 20.983 & 0.425 & 0.442 & 0.221 & 0.829 \\
    LEM          & 14.264 & 21.244 & 0.430 & 0.448 & 0.224 & 0.818 \\
    LEM-RNN      & 13.091 & 19.428 & 0.394 & 0.409 & 0.205 & 0.838 \\
    KF-RNN       & 13.449 & 20.202 & 0.405 & 0.426 & 0.213 & 0.839 \\
    LEM-KF-RNN   & \textbf{12.978} & \textbf{19.361} & \textbf{0.391} & \textbf{0.408} & \textbf{0.204} & \textbf{0.841} \\
    Ens--LEM-RNN & 13.035 & 19.576 & 0.393 & 0.412 & 0.206 & 0.834 \\
    Ens--KF-RNN  & 13.173 & 19.671 & 0.397 & 0.414 & 0.207 & 0.834 \\
    DNN     & 13.795 & 20.099 & 0.416 & 0.423 & 0.212 & 0.809 \\
    DNN/-Fuel          & 15.317 & 22.237 & 0.462 & 0.468 & 0.234 & 0.819 \\
    LEAR    & 17.661 & 24.574 & 0.532 & 0.518 & 0.259 & 0.827 \\
    LEAR/-Fuel        & 18.285 & 25.067 & 0.551 & 0.528 & 0.264 & 0.821 \\

    \midrule
    \multicolumn{7}{c}{\textbf{TestYear\_3 : 2024}}\\[-0.6ex]
    \midrule
    Naive        & 32.815 & 55.458 & 1.000 & 1.000 & 0.705 & 0.660 \\
    RNN          & 13.220 & 26.576 & 0.403 & 0.479 & 0.338 & 0.852 \\
    KF           & 13.256 & 26.469 & 0.404 & 0.477 & 0.336 & 0.859 \\
    LEM          & 13.550 & 27.530 & 0.413 & 0.496 & 0.350 & 0.854 \\
    LEM-RNN      & 12.966 & 26.475 & 0.395 & 0.477 & 0.336 & 0.855 \\
    KF-RNN       & 12.857 & 26.372 & 0.392 & 0.476 & 0.335 & 0.851 \\
    LEM-KF-RNN   & 12.757 & 26.242 & 0.389 & 0.473 & 0.334 & 0.857 \\
    Ens--LEM-RNN & \textbf{12.535} & \textbf{25.971} & \textbf{0.382} & \textbf{0.468} & \textbf{0.330} & \textbf{0.864} \\
    Ens--KF-RNN  & 12.753 & 26.040 & 0.389 & 0.470 & 0.331 & 0.861 \\
    DNN    & 15.552 & 30.293 & 0.474 & 0.546 & 0.385 & 0.816 \\
    DNN/-Fuel         & 15.233 & 28.319 & 0.464 & 0.511 & 0.360 & 0.834 \\
    LEAR   & 16.830 & 30.896 & 0.513 & 0.557 & 0.393 & 0.837 \\
    LEAR/-Fuel         & 17.060 & 30.767 & 0.520 & 0.555 & 0.391 & 0.837 \\
    \bottomrule
  \end{tabular}

  \vspace{2mm}
  \caption{\added{Forecasting performance by test year. Best values per year are shown in bold.}}
  \label{tab:forecasting_by_year}
\end{table}

\begin{figure}[h!]
  \centering
  \includegraphics[width=\textwidth]{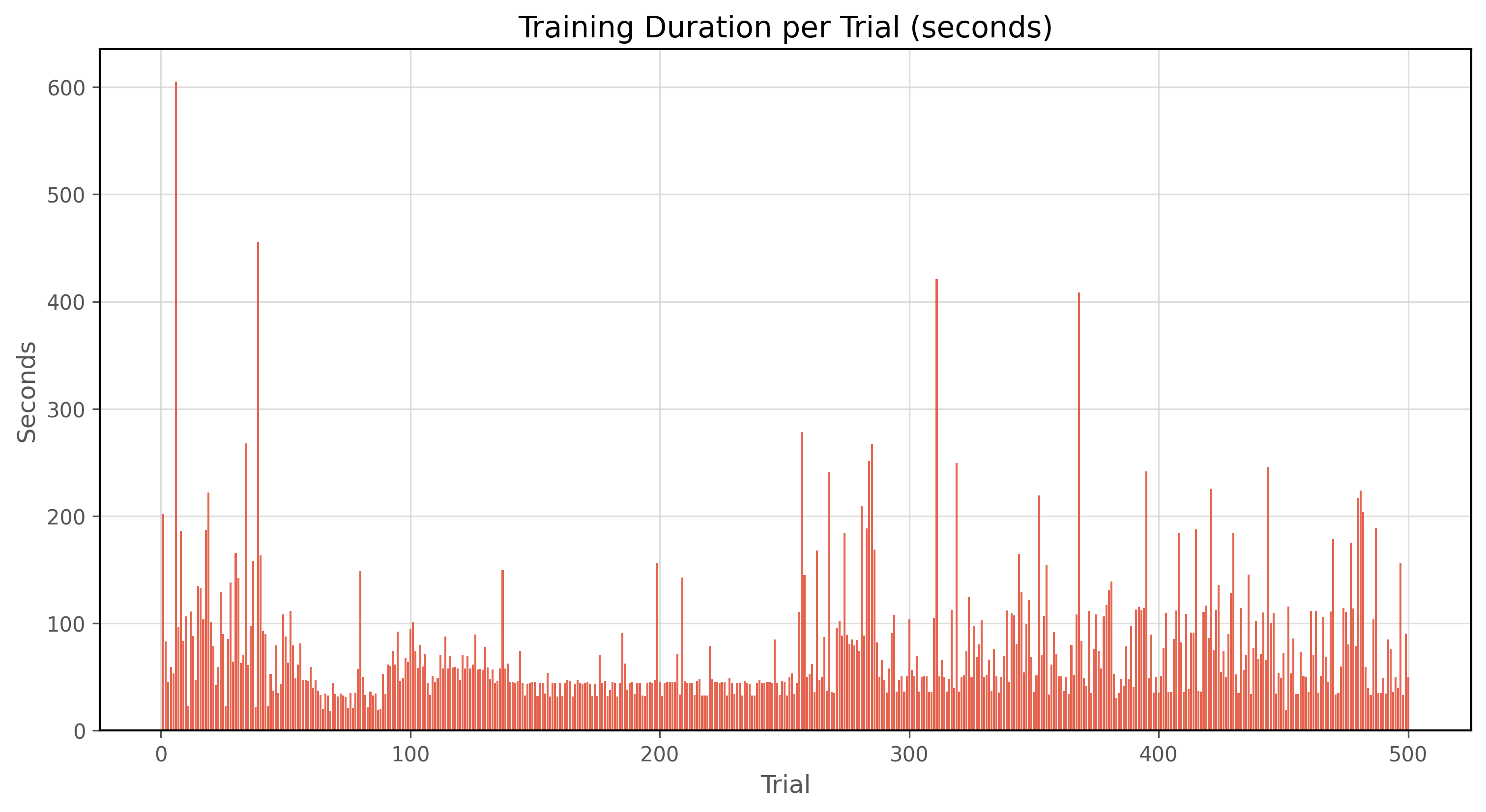}
  \caption{Distribution of Optuna Trial Durations during training for the LEM-KF-RNN Model.}
  \label{fig:trial_durations_bar}
\end{figure}

\bibliographystyle{anc/model5-names}

{\footnotesize
\bibliography{bib}}

\clearpage

\end{document}